\documentclass{article}

\usepackage[main,final]{neurips_2026}

\usepackage[utf8]{inputenc} % allow utf-8 input
\usepackage[T1]{fontenc}    % use 8-bit T1 fonts
\usepackage{url}            % simple URL typesetting
\usepackage{booktabs}       % professional-quality tables
\usepackage{multirow}       % multi-row cells in the real-data results tables
\usepackage{array}          % paragraph-style table columns
\usepackage{tabularx}       % full-width tables with a flexible text column
\usepackage{amsfonts}       % blackboard math symbols
\usepackage{amsmath,amssymb,amsthm}
\usepackage{nicefrac}       % compact symbols for 1/2, etc.
\usepackage{microtype}      % microtypography
\usepackage{xcolor}         % colors
\usepackage{graphicx}
\usepackage{algorithm}
\usepackage{algorithmic}
\usepackage{enumitem}
\usepackage{titletoc}
\usepackage{tikz}
\usepackage{pgfplots}
\pgfplotsset{compat=1.18}

\definecolor{citeblue}{RGB}{28, 110, 164}
\definecolor{refpurple}{RGB}{119, 41, 140}
\definecolor{thmcolor}{RGB}{40, 85, 140}

\usepackage{natbib}

\usepackage[colorlinks=true,
            linkcolor=refpurple,
            citecolor=citeblue,
            urlcolor=citeblue]{hyperref}

\usepackage{thmtools}
\usepackage{placeins}

\declaretheoremstyle[
  headfont=\color{thmcolor}\bfseries,
  notefont=\color{thmcolor}\bfseries,
  bodyfont=\normalfont\itshape,
]{coloredthm}

\declaretheoremstyle[
  headfont=\bfseries,
  notefont=\bfseries,
  bodyfont=\normalfont\itshape,
]{plainthm}

\declaretheoremstyle[
  headfont=\bfseries,
  notefont=\bfseries,
  bodyfont=\normalfont,
]{definitionthm}

\declaretheoremstyle[
  headfont=\itshape,
  notefont=\itshape,
  bodyfont=\normalfont,
]{remarkthm}

\declaretheorem[style=plainthm, name=Theorem]{theorem}
\declaretheorem[style=plainthm, name=Proposition, sibling=theorem]{proposition}
\declaretheorem[style=plainthm, name=Corollary, sibling=theorem]{corollary}
\declaretheorem[style=plainthm, name=Lemma, sibling=theorem]{lemma}

\declaretheorem[style=remarkthm, name=Remark, numbered=no]{remark}

\newcommand{\Prob}{\mathbb{P}}

\newcommand{\DPHCT}{\textsc{DP-HCT}}

\newcommand{\E}{\mathbb{E}}
\newcommand{\R}{\mathbb{R}}

\newcommand{\MCRW}{\mathrm{MCRW}}

\newcommand{\reals}{\mathbb R}

\newcommand{\HC}{\mathrm{HC}}

\providecommand{\dsHeart}{Heart-Circadian}
\providecommand{\dsWorm}{Worm-Bacteria}
\providecommand{\dsBlood}{Blood-Diet}
\providecommand{\dsZebrafish}{Zebrafish-BPF}
\providecommand{\dsBiliary}{Biliary-miRNA}

\makeatletter

\@ifundefined{comment}{%
  \newenvironment{comment}{\color{red}}{}%
}{%
}
\makeatother

\title{Multiclass Classification with Rare Useful Features: Fundamental Limits and the Optimality of Diversity Pursuit Higher Criticism Thresholding}

\author{%
  Gabriel Moryoussef \\
  Department of Statistics and Data Science \\
  Carnegie Mellon University \\
  Pittsburgh, PA 15213, USA \\
  \texttt{gmoryous@andrew.cmu.edu} \\
  \And
  Alon Kipnis \\
  Efi Arazi School of Computer Science \\
  Reichman University \\
  Herzliya, Israel \\
  \texttt{alon.kipnis@runi.ac.il} \\
}

\begin{document}

\maketitle

\begin{abstract}
We consider classification in high-dimensional feature spaces where only a vanishing fraction of features are useful and the number of classes may be large. While this rare-and-weak useful-feature regime is well understood in the binary setting, its multiclass version raises new questions: a feature can be useful for class discrimination without separating any specific pair of classes. We focus on a rare feature diversity model, in which most features are useless, but a small fraction promote discrimination among classes in a way that can be detected by a per-feature diversity test, such as one-way ANOVA or the Kruskal--Wallis test.
For this setting, we propose Diversity Pursuit Higher Criticism Thresholding (DP-HCT), a feature-screening procedure that applies a higher-criticism threshold to the collection of p-values obtained from per-feature diversity tests.
We justify DP-HCT under a Gaussian multiclass Rare-and-Weak Useful Feature model with class-mean contrasts of fixed magnitude in random directions, extending the binary model of Donoho and Jin (2008). When the number of classes is fixed, our model recovers the binary phase transition. When the number of classes grows logarithmically with the number of features, however, we derive a new and strictly harder phase transition, driven by the growing degrees of freedom and their effect on the useful-feature tail rate. We show that whenever classification beyond chance is possible, DP-HCT with ANOVA, followed by a simple classifier based on similarity to class means, achieves asymptotically perfect accuracy.
Experiments on real multiclass classification benchmarks demonstrate substantial improvements, and numerical simulations confirm that the derived phase-transition curve accurately predicts the finite-sample behavior of DP-HCT in the growing-class regime.
\end{abstract}
\section{Introduction} 
\label{sec:intro}

Modern classification problems often involve high-dimensional representations, where the number of measured features $p$ is large but the discriminating information is concentrated in a small subset of coordinates. This sparse-useful-feature structure arises naturally in genomics, text, tabular data, and learned representations, and it motivates feature-screening methods that identify informative coordinates before classification. Such identification is also known to be crucial when data are scarce or the noise level is high \citep{ingster2009classification,donoho2009feature,fan2008high,azar2024semi}. Understanding when classification is possible in this rare-and-weak useful-feature regime is important for assessing sample complexity, the value of feature engineering, and the role of dimensionality reduction or feature screening.

In the binary case, rare-and-weak useful-feature models provide a sharp theoretical framework for understanding this phenomenon \citep{donoho2008higher}. They characterize regimes in which classification is possible despite individual signals being weak \citep{jin2009impossibility,donoho2009feature,loureiro2021learning}. However, multiclass classification introduces qualitatively new issues. A feature may be useful not because it separates one class from another, but because it induces a pattern of variation across many classes \citep{bickel2010hierarchical,abramovich2019classification,thrampoulidis2020theoretical,jin2016influential}. This raises a basic question: how does the number of classes affect the boundary between possible and impossible classification?

We address this question through a Gaussian multiclass Rare-and-Weak Useful Feature model in which class centroids are largely identical, except in a small fraction of the coordinates which individually exhibit statistically detectable diversity signal across classes. This model is designed to isolate the effect of the number of classes on classification in high-dimensions. Our main contribution is a phase-transition analysis for this model that separates classification with accuracy tending to one from classification with accuracy tending to chance level. When the number of classes is fixed, the classification boundary agrees with the binary rare/weak theory \citep{donoho2008higher,donoho2009feature,jin2009impossibility,donoho2015higher}. In contrast, when the number of classes grows logarithmically with the number of features, a new phase transition emerges, capturing the additional difficulty introduced by many classes.

The achievability side of the theory is established by a simple diversity-based screening procedure that computes a diversity score for each feature individually, then applies the higher-criticism threshold for selecting features \citep{donoho2008higher}. We call this screening procedure Diversity Pursuit Higher Criticism Thresholding (\DPHCT{}). Under the rare-and-weak useful feature model, \DPHCT{} followed by a simple centroid-similarity classifier asymptotically achieves perfect classification throughout the region where nontrivial classification is possible. 
Complementing this theoretical characterization of the classification boundary, experiments on several real-world multiclass benchmarks show performance gains from \DPHCT{}.

\begin{figure}[t]
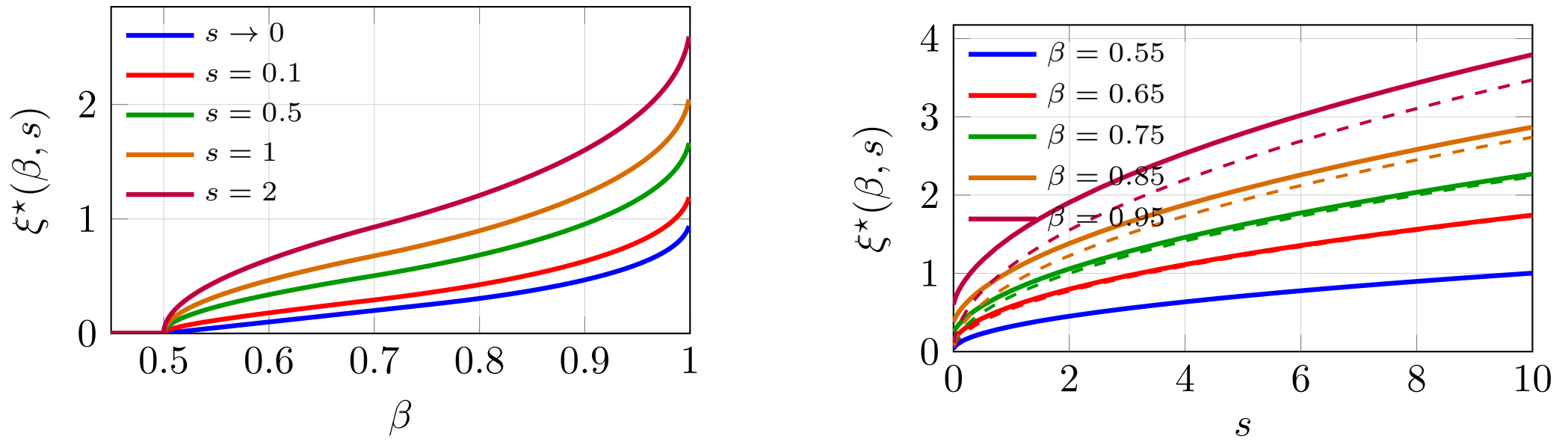

    \centering
    \input{tkz_xi_vs_beta}
    \hfill
    \input{tkz_xi_vs_s}
    \caption{Multiclass classification boundary \(\xi^\star(\beta,s)\) of~\eqref{eq:xi-boundary}. Left:
    boundary as a function of the feature-rarity exponent \(\beta\), for
    several values of the class-growth parameter \(s\). Right: boundary as a
    function of \(s\), for several values of \(\beta\). Dashed lines represent the large $s$ approximation $\sqrt{2 s r}$. The case $s\to 0$ corresponds to the binary Rare-and-Weak Useful Feature boundary \citep{donoho2008higher,jin2009impossibility}. Larger \(s\) shifts the
    boundary upward, indicating the larger signal strength \(r\) required for
    nontrivial classification. }
    \label{fig:xi-vs-beta}
\end{figure}

\paragraph{Contributions}
\begin{enumerate}
\item \textbf{A fundamental limit for multiclass classification with rare and weak useful features.}
We introduce a multiclass Rare-and-Weak Useful Feature model in which the number of classes grows logarithmically with the number of features \(p\). The model is governed by three calibration parameters: \(\beta\), which controls feature rarity; \(s\), which controls the growth rate of the number of classes; and \(r\), which controls signal strength. Our main result establishes a sharp phase transition as \(p \to \infty\): there exists a function \(\xi^\star(\beta,s)\) such that if \(r < \xi^\star(\beta,s)\), no classifier achieves accuracy beyond chance asymptotically, whereas if \(r > \xi^\star(\beta,s)\), classification with accuracy tending to one is possible. The function \(\xi^\star(\beta,s)\) is illustrated in Fig.~\ref{fig:xi-vs-beta}.

\item \textbf{Constructive achievability via Diversity Pursuit Higher Criticism Thresholding.}
We prove that this fundamental limit is attained by a simple diversity-based screening procedure. Specifically, when \(r>\xi^\star(\beta,s)\), applying higher-criticism thresholding to ANOVA \(p\)-values, followed by a centroid-similarity classifier on the selected features, achieves classification accuracy tending to one. 

\item \textbf{Numerical validation and empirical evidence on multiclass benchmarks.}
We run large-scale Monte Carlo experiments across the model's defining parameters. The simulations show that HC thresholding, despite not using the model parameters, closely tracks the oracle threshold and yields empirical accuracy consistent with the predicted phase-transition curve \(\xi^\star(\beta,s)\). We also compare classification post-\DPHCT{} to alternative screening rules, including Bonferroni and Benjamini--Hochberg FDR \citep{benjamini1995controlling}. Finally, we evaluate classification post-\DPHCT{} on five real multiclass benchmarks: gene expression (\dsHeart{}, \dsWorm{}), metabolomics (\dsBlood{}, \dsZebrafish{}) and miRNA expression (\dsBiliary{}). Across these datasets, \DPHCT{} substantially improves the kernel SVM and logistic-regression baselines, providing proof-of-concept evidence for diversity-based HC screening in multiclass problems.
\end{enumerate}

\paragraph{Significance of the contributions compared to previous work.}
The phase transition expression we derive provides an accurate description of the effect of the number of classes on the possibility of successful classification. By using this expression as a pivot, one can assess the sample complexity of classification problems or the amount of feature engineering needed to achieve successful classification. Similar attempts in related models are provided in \citet{abramovich2021multiclass,thrampoulidis2020theoretical}. To the best of our knowledge, our setting is the first to incorporate the number of classes in an asymptotic phase transition analysis.

Classification under a rare-and-weak useful-feature setting has been studied by \citet{donoho2008higher,donoho2009feature}, who also proposed using higher criticism for feature screening. The impossibility of classification in such a model was established in \citet{jin2009impossibility}. Our work is best seen as an extension of this result to the multiclass setting with an increasing number of classes. Our phase transition curve accounts for the change in the full useful-feature tail rate caused by the added degrees of freedom, an effect absent from the binary and fixed-class settings. Other rare-and-weak useful-feature settings can be found in \citet{azar2024semi,verzelen2017detection} and references therein. More closely related to the multiclass setting, the works \citet{jin2016influential,jin2017phase,ma2026if} proposed using higher criticism to identify useful features for clustering. Their setting can also be seen as a diversity-seeking situation among several groups, for which higher criticism has provided promising empirical results. The associated theoretical analysis, however, does not incorporate growth in the number of groups. We believe that ideas from our work can refine the theoretical understanding of clustering as well.

\paragraph{Analysis Overview}

%\AK{See if we use \cite{kipnis2025unification} or directly from the cho-saured moment-generating function and fix this paragraph accordingly.}
The contributions combine several techniques. We derive the useful-feature tail exponent directly from the noncentral chi-square moment-generating function, using a Chernoff bound and an exponential tilting adjusted to the increase in the degrees of freedom. This analysis extends the fixed-degree chi-square approximations in \citet{kipnis2025unification,donoho2004higher}. Concentration of the within-class variance transfers this exponent to the exact ANOVA F-test p-values. We distinguish two regimes: $C=o(\log p)$, including fixed $C$, which recovers the binary boundary; and $C-1\sim2s\log p$ with fixed $s>0$, which yields the growing-class boundary.

% the broader tail-exponent viewpoint is related to \citet{kipnis2025unification}.

When $C$ is fixed (or grows as $o(\log p)$), the analysis recovers the classical Gaussian boundary \citep{donoho2004higher,donoho2008higher}; when $C \sim 2s \log p$, it produces a strictly different boundary. The new boundary represents an increase in the difficulty of classification due to the increased number of classes.

The impossibility region follows from a truncated likelihood-ratio second-moment argument for the joint training and test experiment. Above the boundary, HC attains a polynomially large score, which guarantees polynomially many useful selections and a favorable ratio of squared useful selections to total selections. Uniform threshold bounds transfer these properties to the data-chosen threshold.

\paragraph{Organization}
Section~\ref{sec:method} introduces \DPHCT{} for feature screening. In Section~\ref{sec:model}, we specify the multiclass Rare-and-Weak Useful Feature framework. In Section~\ref{sec:theory}, we derive the asymptotic phase transition for successful classification and we show that \DPHCT{} based on ANOVA and a centroid-similarity classifier achieves the theoretical detectability region. Section~\ref{sec:experiments} reports simulation and real-data results. Additional simulation results and full proofs are provided in the Supplementary Material.

\section{Diversity Pursuit with HC Thresholding}
\label{sec:method}

We now introduce the \DPHCT{} feature-screening procedure. Subsequent sections justify this procedure theoretically and empirically.
%The framework is deliberately model-free at this stage: it takes labelled data as input and returns a feature subset together with a classifier, without committing to any particular distributional model. The statistical analysis, which \emph{does} require a model, is deferred to Sections~\ref{sec:model}--\ref{sec:theory}.

\subsection{Feature screening}
\label{sec:method-setup}

We have a training dataset $\{(X_i, Y_i)\}_{i=1}^{n}$ with $X_i  \in \reals^p$ and $Y_i \in [C] := \{1, \ldots, C\}$. For $j \in [p]$ and $c \in [C]$, let $P_{c,j}$ denote the class-conditional marginal distribution of feature $j$. A feature $j$ is \emph{useless} (a.k.a. \emph{null}) if 
\begin{align}
    \label{eq:feature-hypothesis}
    H_0^{(j)} : P_{1,j} = P_{2,j} = \cdots = P_{C,j},
\end{align}
and \emph{useful} otherwise. Our goal is to improve classification performance by removing useless features. We propose the following \DPHCT screening procedure, which decomposes into three steps.

\textit{(i) Per-feature diversity testing.} For each $j$, compute a test statistic $D_j$ of~\eqref{eq:feature-hypothesis} on $\{(X_{i,j}, Y_i)\}_{i=1}^n$ and convert it to a p-value $\pi_j \in [0,1]$. 
%, such that $\pi_j \sim \mathrm{Uniform}[0,1]$ under $H_0^{X_{i,j}}$ (exactly, or up to a vanishing error).

\textit{(ii) Higher Criticism Thresholding.} Let $\pi_{(1)} \leq \cdots \leq \pi_{(p)}$ denote the order statistics of $\{\pi_j\}_{j=1}^p$. The Higher Criticism Thresholding (HCT) procedure selects a set of indices $\hat{\mathcal{J}}  \subseteq [p]$ whose p-values are below a threshold $\pi_{(j^\star)}$, where $j^\star$ is the index that maximizes the Higher Criticism statistic. Specifically, with
\begin{align*}
    \HC_{p,j} \triangleq \sqrt{p} \cdot \frac{j/p - \pi_{(j)} }{\sqrt{t_j(1-t_j)}},
    \qquad j = 1, \ldots, p,
\end{align*}
where %either $t_j := p_{X_{i,j}}$ \citep{donoho2004higher} or 
$t_j := j/p$. The HC statistic and the HCT are defined as in \citet{donoho2008higher,donoho2009feature,donoho2015higher}:
\begin{align}
    \label{eq:HC-statistic}
    \HC_p := \HC_{p,j^\star},
    \qquad
    j^\star := \arg\max_{ 1 \leq j \leq \lfloor p\gamma_0 \rfloor} \HC_{p,j},
\end{align}
where $\gamma_0$ is a tunable parameter that reflects prior knowledge about the proportion of useful features; we used $\gamma_0=1/2$ in all evaluations. 

\textit{(iii) Selection.} The higher criticism-selected set is 
\begin{align*}
    \hat{\mathcal{J}} := \{j \in [p] : \pi_j \leq \pi_{(j^\star)}\}. 
\end{align*}

The selection procedure (i)-(iii) is summarized in Algorithm~\ref{alg:dphct}. 

\begin{algorithm}[t]
\caption{\DPHCT{}: Diversity Pursuit with Higher Criticism Thresholding.}
\label{alg:dphct}
\begin{algorithmic}[1]
\REQUIRE training data $\{(X_i, Y_i)\}_{i=1}^n$; diversity test $D$; 
\FOR{$j = 1, \ldots, p$}
  \STATE compute $D_j$ on $\{(X_{i,j}, Y_i)\}_{i=1}^n$; obtain p-value $\pi_j$.
\ENDFOR
\STATE sort: $\pi_{(1)} \leq \pi_{(2)} \leq \cdots \leq \pi_{(p)}$.
\STATE for $j = 1, \ldots, \lfloor \gamma_0 p \rfloor$, set $\HC_{p,j} \leftarrow \sqrt{p}\,(j/p - \pi_{(j)})/\sqrt{j/p(1-j/p)}$.
\STATE $j^\star \leftarrow \arg\max_{ 1 \leq j \leq \lfloor \gamma_0 p \rfloor} \HC_{p,j}$;
\STATE $\hat{\mathcal{J}} \leftarrow \{j \in [p] : \pi_j \leq \pi_{(j^\star)}\}$.
\ENSURE selected set $\hat{\mathcal{J}}$
\end{algorithmic}
\end{algorithm}

Given the selected set $\hat{\mathcal{J}}$, we define the feature-screened data as 
\begin{align}
    \label{eq:screened-data}
    X_i[\hat{\mathcal{J}}] := (X_{i,j})_{j \in \hat{\mathcal{J}}} \in \reals^{|\hat{\mathcal{J}}|}.
\end{align}
We train a multiclass classifier $\hat h$ on $\{(X_i[\hat{\mathcal{J}}], Y_i)\}_{i=1}^n$. For a test point $X_{n+1}$, predict $\hat Y_{n+1} := \hat h(X_{n+1}[\hat{\mathcal{J}}])$.

\subsection{Diversity Statistics} 
\label{sec:method-menu}

Any per-feature test of~\eqref{eq:feature-hypothesis} that produces a null-calibrated p-value fits the template, and different tests can even be applied to different features. Two canonical tests that we explain in detail and apply in practice are the classical Analysis of Variance (ANOVA) F-test and the Kruskal--Wallis test \citep{kruskal1952use}. Aside from these tests, other procedures can be used as diversity statistics, such as the Friedman test and permutation-based tests \citep[Ch. 7]{lehmann2006testing}. Post-hoc tests such as Tukey's method \citep[Ch. 9]{lehmann2006testing} may be useful in extensions of the framework to hierarchical features. 

\paragraph{ANOVA F-test.}
Let $n_c := |\{i : Y_i = c\}|$ be the number of observations in class $c$ and assume $n_c \geq 2$. At feature $j$, denote the class sample mean by
\begin{align}
    \label{eq:mu_hat}
    \hat{\mu}_{c,j} := \frac{1}{n_c} \sum_{i \,:\, Y_i = c} X_{i,j},\quad c=1,\ldots,C,
\end{align}
Define the within- and between-class sums of squares as 
\begin{align}
    \label{eq:SS}
\mathrm{SS}_{\mathrm{between}}(j) \;=\; \sum_{c=1}^C n_c \bigl( \hat{\mu}_{c,j} - n^{-1}\sum_{a=1}^C n_a \hat{\mu}_{a,j} \bigr)^2,\quad \mathrm{SS}_{\mathrm{within}}(j) \;=\; \sum_{c=1}^C \sum_{i \,:\, Y_i = c} \bigl(X_{i,j} - \hat{\mu}_{c,j} \bigr)^2,
\end{align}
and
\begin{align}
    \label{eq:F-statistic}
f_j \;:=\; \frac{\mathrm{SS}_{\mathrm{between}}(j)/(C-1)}{\mathrm{SS}_{\mathrm{within}}(j)/(n-C)}.
\end{align}
With $F_{C-1,n-C}$ denoting the $F$ distribution with $(C-1, n-C)$ degrees of freedom, p-values are provided by 
\begin{align}
    \label{eq:F-p-value}
    \pi_j := \Prob\{F_{C-1,n-C}\ge f_j\}, \qquad j=1,\ldots,p.
\end{align}
This test is appropriate under $H_0^{(j)}$ of~\eqref{eq:feature-hypothesis} when the class-conditional marginals are Gaussian, i.e., $P_{c,j} = \mathcal{N}(\mu_{c}^{(j)}, \sigma^2)$, with an unknown common variance $\sigma^2$.

\paragraph{Kruskal--Wallis Test.}
This test is often seen as the rank-based analogue of the ANOVA F-test. Rank-based diversity is invariant to monotone transformations and requires no parametric assumption. Therefore, the Kruskal--Wallis (KW) test is particularly useful when class-conditional marginals are heavy-tailed, skewed, or contaminated \citep[Ch. 7]{lehmann2006testing}. 

Let $R_{i,j}$ be the rank of $X_{i,j}$ in $\{X_{1,j}, \ldots, X_{n,j}\}$ and $\bar R_{c,j} := n_c^{-1}\sum_{i \,:\, Y_i = c} R_{i,j}$ be the class mean rank at feature $j$. The KW statistic of the $j$-th feature is defined as
\begin{align}
\mathrm{KW}_j := \frac{12}{n(n+1)} \sum_{c=1}^C n_c \Bigl(\bar R_{c,j} - \tfrac{n+1}{2}\Bigr)^2. 
\label{eq:KW-test}
\end{align}
As $n \to \infty$ under $H_0^{(j)}$ with continuous marginals, $\mathrm{KW}_j$ converges to a $\chi^2_{C-1}$ distribution. Thus, (asymptotic) p-values are provided by 
\begin{align}
    \label{eq:KW-p-value}
    \pi_j := \Prob\{\chi^2_{C-1}\ge \mathrm{KW}_j\}, \qquad j=1,\ldots,p.
\end{align}

\section{Multiclass Rare-and-Weak Useful Feature Model}
\label{sec:model}

\subsection{Overview}
We consider a model with $C$ classes, where only a vanishing fraction of features are useful for inter-class discrimination. We call a feature \emph{useful} if it promotes discrimination among classes in a way that can be detected by a per-feature diversity test, such as \eqref{eq:F-p-value} or \eqref{eq:KW-p-value}. We introduce a single parameter \(\tau\) that controls the strength of the useful class-mean contrasts. To enforce rarity, we suppose that each feature belongs to the set of useful features $\mathcal{J} \subset [p]$ with probability $\epsilon$, for some small number $\epsilon \in (0,1)$. The resulting Multiclass Rare-and-Weak Useful Feature model is denoted by $\MCRW(p,C,\epsilon,\tau)$. 

\subsection{Multiclass Gaussian data}
\label{sec:model-setup}

Suppose that, conditional on $Y_i = c$,
\begin{equation}
\label{eq:gaussian-model}
X_i \mid Y_i = c \;\sim\; \mathcal{N}\bigl(\mu_c,\ \sigma^2 I_p\bigr),
\qquad
\mu_c = (\mu_c^{(1)}, \ldots, \mu_c^{(p)}) \in \reals^p.
\end{equation}
Here \(\mu_c\) is the class-\(c\) centroid and \(\sigma^2>0\) is a shared noise-scaling parameter. For simplicity, we assume that $n_c = n/C$ is an integer, and that the design is balanced with $n_c$ observations per class. Unbalanced designs are interesting in some contexts \citep{krawczyk2016learning}, but we leave their analysis as future work.

\subsection{Rare useful features}
\label{sec:rare}
Inter-class diversity in the generation process \eqref{eq:gaussian-model} is encoded by the class means $\mu_1,\ldots,\mu_C$. Let $\mathcal{J} \subset [p]$ denote the coordinates of features that promote inter-class separation in the following sense. For $j \notin \mathcal{J}$, all class means have the same $j$-th coordinate value, whereas for $j \in \mathcal{J}$, some class means may have $j$-th coordinates that differ from the others. The set of useful features $\mathcal{J}$ is assumed to be random, with each coordinate belonging to $\mathcal{J}$ with probability $\epsilon$ independently of the others. 
%$\beta$ is called the \emph{rarity} parameter. This calibration implies that the size $\mathcal{J}$ is roughly $n^{1-\beta}$. 

Deterministic constructions of \(\mathcal J\) that directly control \(|\mathcal J|\) are also possible \citep{stoepker2024anomaly}; they are slightly more difficult to analyze, yet behave similarly under the asymptotic regime of interest introduced below.

\subsection{Spherically symmetric diversity model of useful features}
Let
\begin{align*}
    \mathcal{V} :=\mathbf 1^\perp =
    \left\{
    u\in \reals^C:\mathbf 1^\top u=0
    \right\},
    \qquad
    P=I_C-\frac1C\mathbf 1\mathbf 1^\top .
\end{align*}
For each $j \in \mathcal{J}$, draw independently
\[
    U^{(j)} \sim \operatorname{Unif}\{u \in \mathcal{V} : \|u\|_2 = 1\},
\]
and set
\begin{equation}
\label{eq:spherical-contrast}
    v^{(j)} := \sqrt{C}\, U^{(j)},
    \qquad
    \mu^{(j)} := \tau v^{(j)} = \bigl(\mu_1^{(j)},\ldots,\mu_C^{(j)}\bigr)^\top.
\end{equation}
For $j \notin \mathcal{J}$, all class means are equal in coordinate $j$, equivalently $P\mu^{(j)} = 0$. We say that this construction is spherically symmetric because, for each
\(j\in\mathcal J\),
\[
    \mathbf 1^\top \mu^{(j)}=0,
    \qquad
    \|\mu^{(j)}\|_2^2=\tau^2 C.
\]

The spherical contrast prior is a symmetry-preserving way to extend the
binary Rare-and-Weak Useful Feature model of Donoho and Jin to multiple classes. When \(C=2\), the
centered contrast space \(\mathbf 1^\perp\) is one-dimensional, and the
uniform distribution on its unit sphere reduces to a random sign. For
\(C>2\), the same construction gives the rotationally invariant prior on the centered class-contrast space. Such spherical priors are common in high-dimensional inference and random matrix theory because they preserve the symmetries of the null model and often lead to analytically tractable, least-favorable alternatives; see, for example,
\citet{baik2005phase}, \citet{perry2018optimality}, and
\citet{elalaoui2020fundamental}. Other multiclass sparse-feature
constructions are possible, including the high-dimensional
normal model of \citet{abramovich2019classification}, but it is unclear whether these can yield an interesting asymptotic behavior.

We denote the data-generation model defined so far as $\MCRW(p,C,\epsilon,\tau)$. We keep the dependence of the model on $\sigma^2$ and $n$ implicit, since these parameters do not directly affect the asymptotic behavior. 

\subsection{The ideal threshold}
\label{sec:ideal-threshold}

For simplicity of the analysis, but also due to asymptotic optimality under $\MCRW(p,C,\epsilon,\tau)$, we consider a maximum-inner-product classification rule over the selected features:
\begin{align}
    \label{eq:ncm}
        \hat h(x; \mathcal J):=
      \arg\max_{c\in[C]}
    \bigl(x[\mathcal J]\bigr)^\top \hat\mu_c[\mathcal J],
    \end{align}
where $\hat\mu_c=(\hat{\mu}_{c,1},\ldots,\hat{\mu}_{c,p})$ collects the class sample means \eqref{eq:mu_hat} and $x[\mathcal J]$ is coordinate selection operation \eqref{eq:screened-data}. For a general threshold $t \in [0,1]$, define the selected set of features based on ANOVA p-values as
\begin{align}
    \label{eq:p-value-threshold}
    \hat{\mathcal J}(t) := \{j \in [p] : \pi_j \leq t\}.
\end{align}

Notice that under this construction, the expected misclassification error under $\MCRW(p,C,\epsilon,\tau)$ is a function of the threshold $t$. We denote this function by 
\[
e(t) := \E_{\MCRW(p,C,\epsilon,\tau)} \left[ {\mathbf 1}\{ \hat h(X; \hat{\mathcal J}(t)) \neq Y \} \right].
\]
The ideal threshold, denoted by $t^\star$, is the one that minimizes $e(t)$:
\begin{align}
    \label{eq:ideal-threshold}
    t^\star := \arg\min_{t\in[0,1]} e(t).
\end{align}
It is not clear how to compute $e(t)$, as even the case of $C=2$ is challenging and was not fully solved in \cite{donoho2009feature}. However, we can approximate $e(t)$, and thus $t^\star$, relatively straightforwardly using Monte Carlo simulations, as demonstrated in Fig.~\ref{fig:compare_line_scans} below. 

The ideal threshold $t^\star$ depends on the latent model parameters $\epsilon$ and $\tau$, hence it is not accessible to the data analyst. We hope that HCT approximates the ideal threshold, at least in the parts of the parameter space above and nearby the detection threshold, where non-trivial classification is challenging but possible.
\subsection{Model calibration for asymptotic analysis}
\label{sec:model-calibration}

We consider the asymptotic limit as \(p\to\infty\), where the model
parameters are calibrated to \(p\) through fixed parameters
\((\beta,r,s)\). The rarity level is
\begin{equation}
\label{eq:rare_calibration}
    \epsilon = p^{-\beta},
    \qquad
    \beta\in(1/2,1).
\end{equation}
The signal amplitude is calibrated by
\begin{equation}
\label{eq:tau-calibration}
    \tau^2
    =
    \frac{\sigma^2}{n}\,2r\log p,
    \qquad
    r>0.
\end{equation}
Notice that under the spherical contrast construction, every useful feature has ANOVA non-centrality
\begin{equation}
\label{eq:lambda-calibration}
\lambda_j
    =
    \frac{n_c}{\sigma^2}\mu^{(j)\top}P\mu^{(j)}
    = \frac{n/C}{\sigma^2}\,C\tau^2 = 
    2r\log p,
    \qquad j\in\mathcal J.
\end{equation}
Thus, \(r\) is the standardized ANOVA-visible rare and weak signal strength \citep{donoho2004higher}.

For the number of classes, we focus on the logarithmic multiclass regime
\begin{equation}
\label{eq:class_count}
    C \sim 2s\log p,
    \qquad
    s>0.
\end{equation}
We also discuss the fixed- or slowly-growing-class regime \(C=o(\log p)\)
as a comparison case. For the ANOVA F-test analysis, we assume
\begin{equation}
\label{eq:f_to_chi}
    \log p=o(n),
\end{equation}
so that the normalized within-class variance denominator of the F statistic concentrates at $\sigma^2$. Consequently, $(C-1)f_j$ has the same polynomial-tail exponent as a non-central chi-square random variable with $C-1$ degrees of freedom and non-centrality parameter $\lambda_j$.

\section{Main Results: Classification Phase Transition and Optimality of \DPHCT{}}
\label{sec:theory}

We now state the main theoretical consequences for \DPHCT{} under the
\(\MCRW(p,C,\epsilon,\tau)\) framework of Section~\ref{sec:model}. Throughout, \(\pi_j\) denotes the p-values obtained from the one-way ANOVA F-test \eqref{eq:F-p-value}.

\subsection{The classification boundary}
\label{sec:theory-boundary}

%For the ANOVA statistic with growing degrees of freedom, the exact useful-feature tail exponent must instead be computed directly, as in Section~\ref{app:md-proof} of the Supplementary Material. 

% The useful-feature exponent $\Psi_{r,s}(q)$ is characterized by
% \[
%  \Prob\{\pi_j\le p^{-q}\mid j\in\mathcal J\}
%  =p^{-\Psi_{r,s}(q)+o(1)},\qquad q>0,
% \]
% as established in Section~\ref{app:md-proof} of the Supplementary Material.
For $s>0$, let $x_s(q)\ge s$ solve
\[
 g_s(x_s(q))=q,\qquad g_s(x)=x-s-s\log(x/s),
\]
and set $x_0(q)=q$. Define
\[
 \Psi_{r,s}(q)=J_{r,s}(x_s(q)),\qquad
 J_{r,s}(x)=\sup_{0\le\theta<1}
 \left\{\theta x+s\log(1-\theta)-\frac{r\theta}{1-\theta}\right\}.
\]
For fixed $(\beta,s)$, the signal-strength boundary function is defined by 
\begin{equation}
\label{eq:xi-boundary}
 \xi^\star(\beta,s):=\inf\left\{r>0:
 \max_{0\le q\le1}\left(\frac12-\beta+\frac q2-\Psi_{r,s}(q)\right)>0\right\}.
\end{equation}
The maximum in \eqref{eq:xi-boundary} has an explicit optimizer, given in Section~\ref{app:pivot} of the Supplementary Material. 

The boundary function $\xi^\star(\beta,s)$ is strictly increasing in $s$ and has the following limiting forms. 
\begin{itemize}[leftmargin=*,noitemsep]
\item As $s \to 0$, $\xi^\star(\beta,s) \to \rho^\star(\beta)$, where $\rho^\star(\beta)$ is the Ingster-Donoho-Jin rare signal detection boundary function
\begin{align*}
    \rho^\star(\beta) := \begin{cases}
\beta-1/2, & 1/2 < \beta < 3/4, \\
(1-\sqrt{1-\beta})^2, & 3/4 \le \beta < 1.
\end{cases}
\end{align*}
This is also the boundary for $C=o(\log p)$, fixed $C$ in particular.
\item As $s\to\infty$, $\xi^\star(\beta,s)\sim\sqrt{2s\,\rho^\star(\beta)}$. This is an iterated limit of the fixed-$s$ boundary. The ANOVA numerator has null standard deviation $\sqrt{2(C-1)}$, so its standardized non-centrality is $\lambda/\sqrt{2(C-1)}\sim r\sqrt{\log p/s}$. Thus additional degrees of freedom reduce per-feature separation. The large-$s$ approximation is shown as dashed reference lines in Fig.~\ref{fig:xi-vs-beta} (right).
\end{itemize}

%%
% Then $\Prob(\pi_j\le p^{-q}\mid j\in\mathcal J)=p^{-\Psi_{r,s}(q)+o(1)}$. For $x\le s+r$, $J_{r,s}(x)=0$; for $x>s+r$, it has the explicit expression~\eqref{eq:J-explicit}. In particular, $\Psi_{r,0}(q)=(\sqrt q-\sqrt r)_+^2$.
% %%

\subsection{Impossibility below the boundary}
\label{sec:theory-impossibility}

\begin{theorem}[Impossibility below the boundary]
\label{thm:impossibility}
Under \(\MCRW(p,C,\epsilon,\tau)\)
with calibrations~\eqref{eq:rare_calibration}--\eqref{eq:f_to_chi} and a uniformly distributed independent test label, if \(r<\xi^\star(\beta,s)\), then every
measurable classifier $\hat{h}$ satisfies
\[
    \Prob\{\hat h(X_{n+1})=Y_{n+1}\}
    \le
    \frac{1}{C} + o(1).
\]
Consequently,
\[
    \sup_{\hat h}\Prob\{\hat h(X_{n+1})=Y_{n+1}\}
    =
    \frac{1}{C} + o(1).
\]
\end{theorem}

The proof is in Section~\ref{app:impossibility-proof} of the Supplementary Material. It uses a truncated likelihood-ratio second-moment argument to establish vanishing total variation for the joint training and test experiment; see \citet{butucea2013detection} for a related use of this method.

\subsection{\DPHCT{} succeeds above the boundary}
\label{sec:theory-classification}

%\AK{does this maps cleanly to the standard HC? if so, can we introduce this new HC version only in the proof and re-use our HCT definition from Section 2.2?}

\begin{theorem}[\DPHCT{} succeeds above the boundary]
\label{thm:classification}
Under \(\MCRW(p,C,\epsilon,\tau)\) with
calibrations~\eqref{eq:rare_calibration}--\eqref{eq:f_to_chi} and
\(n=p^{o(1)}\), let $\hat{\mathcal J}$ be the selected set of features based on ANOVA p-values \eqref{eq:F-p-value} and threshold \eqref{eq:p-value-threshold}, and let $\hat h$ be the maximum-inner-product classifier~\eqref{eq:ncm} with $\tilde{\mathcal J}=\hat{\mathcal J}$. If \(r>\xi^\star(\beta,s)\), then, for an independent test point
$(X_{n+1},Y_{n+1}) \sim \MCRW(p,C,\epsilon,\tau)$, 
\[
    \Prob\{\hat h(X_{n+1}; \hat{\mathcal J}) \neq Y_{n+1}\}=o(1).
\]
\end{theorem}

We leave for future work the study of \DPHCT{} optimality under other sample-size regimes, including polynomial growth $n=p^{\theta+o(1)}$ with fixed $\theta\in(0,1)$.

\section{Experiments}
\label{sec:experiments}

\subsection{Verification of successful classification region}
\label{sec:experiments-verification}

\paragraph{Setup.}
We simulate from the $\MCRW(p,C,\epsilon,\tau)$ model with $p = 2{,}000$ and 
$\epsilon$, $\tau$, and $C$ as in, respectively,~\eqref{eq:rare_calibration}, \eqref{eq:tau-calibration}, and \eqref{eq:class_count}. We use $n_{\mathrm{Monte}} = 150$ independent problem instances per $(\beta, r)$ grid point. After feature screening, classification is performed with the centroid-similarity classifier of~\eqref{eq:ncm}.

\begin{figure}[t]
  \centering
  \includegraphics[width=\linewidth]{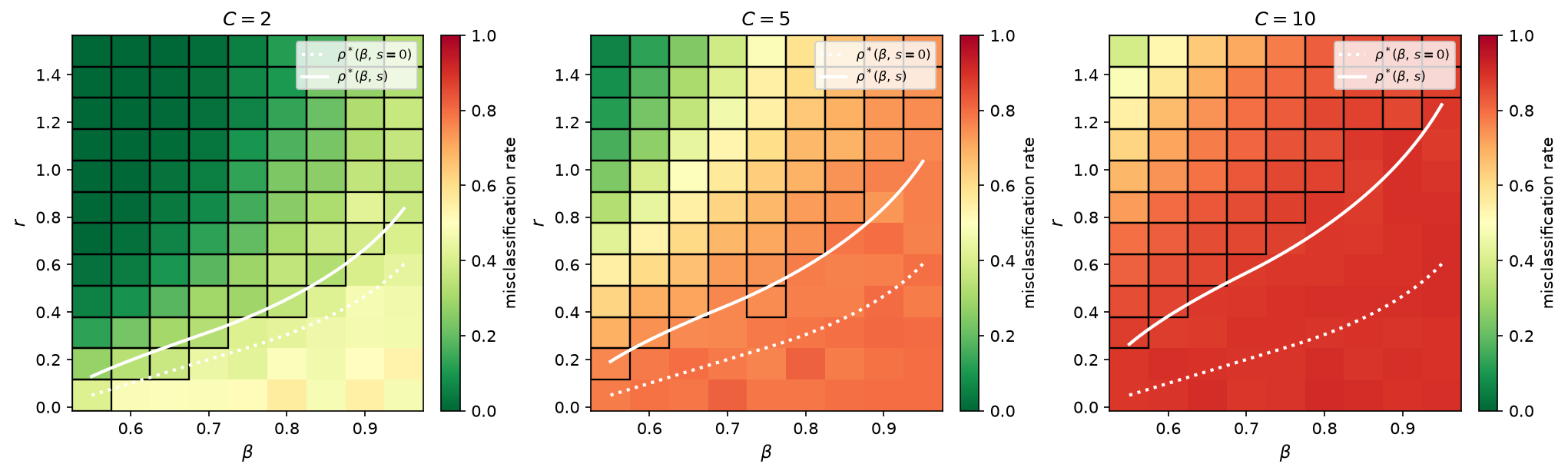}
  \caption{%
    \textbf{\DPHCT{} misclassification phase transition}. Misclassification rate (color) over a grid of $(\beta, r)$ values for $C \in \{2,5,10\}$, under the Gaussian multiclass Rare-and-Weak Useful Feature model with $p = 2{,}000$, $n_{\mathrm{Monte}} = 150$.
    The x-axis is the sparsity exponent $\beta$ and the y-axis is the calibrated signal strength $r$. Configurations in which the misclassification rate is above chance are marked with black frames. Solid white: the theoretical phase transition curve (Theorem~\ref{thm:classification}). Dotted white: the phase transition curve corresponding to a fixed number of classes, derived in \citet{donoho2008higher}.
  }
  \label{fig:phase_transition_hct}
\end{figure}

\paragraph{Misclassification phase transition.}
Fig.~\ref{fig:phase_transition_hct} shows the misclassification rate over a grid of $(\beta, r)$ values for $C \in \{2, 5, 10\}$; configurations in which the misclassification rate is significantly above chance are marked with black frames. As $C$ increases, the boundary rises, reflecting the increased difficulty of multiclass discrimination; two theoretical phase-transition boundaries are overlaid on every heatmap: (1)
the boundary $r = \xi^\star(\beta,s)$ of~\eqref{eq:xi-boundary} (solid) and (2) 
the boundary $r = \rho^\star(\beta)$ (dotted). The boundary (1), established in  Theorems~\ref{thm:impossibility} and \ref{thm:classification}, assumes that $C$ increaes with $p$. The boundary (2), derived in \citet{donoho2008higher}, corresponds to a fixed $C$. The figure shows that the above-chance region better follows boundary (1), even for small $C$. Additional figures characterizing the ideal threshold and per-$C$ side-by-side comparisons are collected in the Supplementary Material, Section~\ref{app:rw-simulations}.

\paragraph{Comparison with other thresholds.}

\begin{figure}[t]
  \centering
  \includegraphics[width=\linewidth]{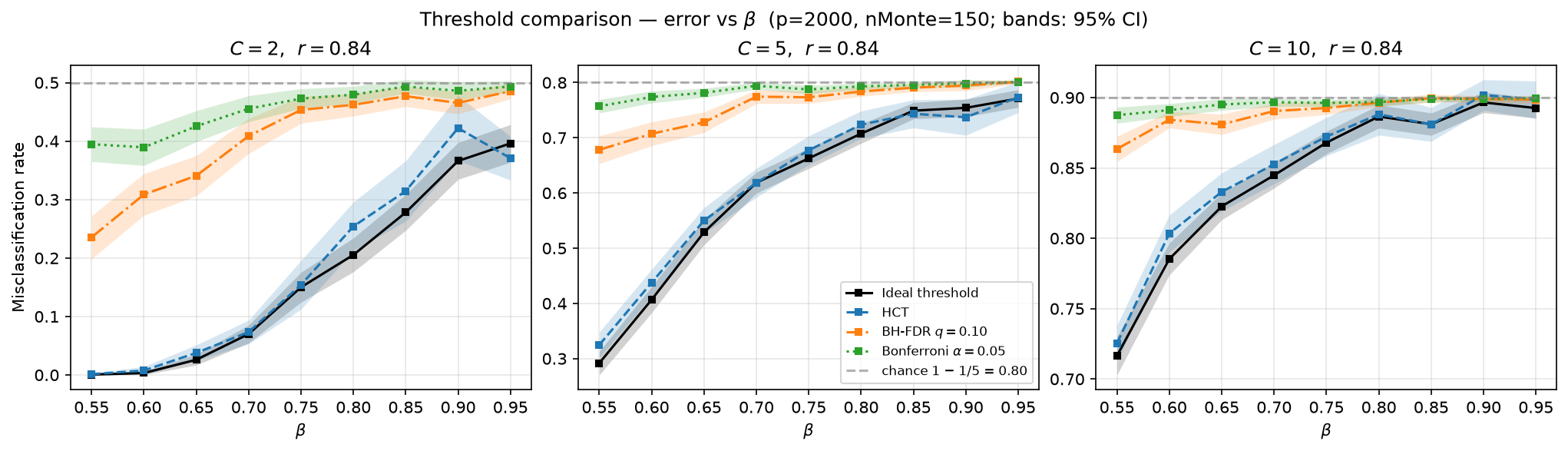}
  \caption{%
    \textbf{Threshold comparison:} classification error vs.\ $\beta$ for a fixed $r$ value and several value of $C$ ($p = 2{,}000$, $n_{\mathrm{Monte}} = 150$).
    Solid black: ideal threshold of~\eqref{eq:ideal-threshold}; dashed blue: \DPHCT{}; dash-dot orange: Benjamini-Hochberg's False-Discovery Rate (BH-FDR) at level $0.10$; dotted green: Bonferroni ($\alpha=0.05$).
    Dashed grey: chance-level error $1-1/C$. It can be seen that \DPHCT{} classification performance closely follows the ideal threshold's performance.
  }
  \label{fig:compare_line_scans}
\end{figure}

Fig.~\ref{fig:compare_line_scans} compares four feature-selection rules for data sampled from the $\MCRW(p,C,\epsilon,\tau)$. The \emph{ideal threshold} of~\eqref{eq:ideal-threshold}, which achieves the best performance achievable by thresholding the ANOVA \(p\)-values, but is not a function of the data. We compare it with \emph{\DPHCT{}}, Benjamini-Hochberg's False-Discovery Rate at level \(0.10\) \citep{benjamini1995controlling}, and selection after Bonferroni's correction with \(\alpha=0.05\).

\subsection{Real data benchmarks}
\label{sec:experiments-real}

The main theoretical justification for \DPHCT{} is its role in a classification pipeline that attains the multiclass Rare-and-Weak Useful Feature phase transition. To complement this theoretical characterization, we evaluate classification following \DPHCT{} against baseline methods on five public biological multiclass benchmarks with $p>n$, spanning gene expression, metabolomics, and miRNA expression; see Section~\ref{app:real-data-datasets} of the Supplementary Material for preprocessing details.

\begin{itemize}[nosep, left=0pt]
  \item \dsHeart{} (GSE10045; \citealp{bray2008circadian}): $n=130$ mouse-heart expression profiles,
    $C=8$ times of day, $p=46{,}120$ probes.
  \item \dsWorm{} (GSE15923; \citealp{coolon2009elegans}): $n=72$ \emph{C.\ elegans} expression profiles,
    $C=4$ bacterial diets, $p=672$ features.
  \item \dsBlood{} (ST002019; \citealp{ferguson2022tips}, Metabolomics Workbench): $n=70$ blood LC--MS profiles from $35$ participants,
    $C=4$ diet $\times$ study-phase groups, $p=736$ metabolite features.
  \item \dsZebrafish{} (ST003804; \citealp{guo2025bpf}, Metabolomics Workbench): $n=62$ zebrafish-liver LC--MS profiles,
    $C=4$ exposure levels of bisphenol~F, $p=117$ metabolite features.
  \item \dsBiliary{} (GSE140001; \citealp{loeffler2020mirna}): $n=36$ microdissected tissue samples from $12$ patients,
    $C=3$ tissue types, $p=803$ miRNAs.
\end{itemize}
 
For the diversity test we use either ANOVA F \eqref{eq:F-p-value} or KW \eqref{eq:KW-p-value}, depending on which one obtained higher classification accuracy. We evaluate two classifiers before and after feature screening: a radial-basis-function SVM (\DPHCT{}-SVM) and $\ell_2$-regularized multinomial logistic regression (\DPHCT{}-LoR). All baselines apply the same classifiers to the full, unselected feature matrix; see Section~\ref{app:real-data} of the Supplementary Material for full protocol details. Table~\ref{tab:clean-results} reports mean cross-validated accuracy for both classifiers pre- and post-\DPHCT{}, and the number of features retained ("\# Sel."). 

\begin{table}[t]
    \centering
    \caption{%
    CV accuracy (mean $\pm$ SEM, \%) for classification post-\DPHCT{} feature screening with ANOVA F ("F") or KW ("KW") tests; 
    results are averaged over 10-fold, unit stratified cross-validation (7-fold for \dsHeart{}). ``\# Sel.'' is the mean number of features retained by \DPHCT{}. 
    }
    \label{tab:clean-results}
    \footnotesize
    \setlength{\tabcolsep}{2.6pt}
    \begin{tabular}{lc cc cc c}
    \toprule
     & & \multicolumn{2}{c}{\textbf{Baseline (all features)}}
       & \multicolumn{2}{c}{\textbf{\DPHCT{}-selected features}} \\
    \cmidrule(lr){3-4}\cmidrule(lr){5-6}
    Dataset & $n/p/C$
      & SVM & LoR
      & SVM & LoR & \# Sel. \\
    \midrule
    \dsHeart{} (F)
      & $130/46120/8$
      & $14.7{\pm}2.4$ & $32.4{\pm}4.5$
      & $\mathbf{74.8{\pm}6.1}$ & $\mathbf{76.3{\pm}4.9}$
      & $460{\pm}47$ \\
    \dsWorm{} (KW)
      & $72/672/4$
      & $24.2{\pm}4.2$ & $71.5{\pm}5.1$
      & $\mathbf{53.4{\pm}7.7}$ & $\mathbf{80.5{\pm}5.4}$
      & $56{\pm}5$ \\
    \dsBlood{} (F)
      & $70/736/4$
      & $47.5{\pm}5.0$ & $60.8{\pm}7.1$
      & $\mathbf{78.8{\pm}4.1}$ & $\mathbf{82.9{\pm}3.9}$
      & $108{\pm}7$ \\
    \dsZebrafish{} (KW)
      & $62/117/4$
      & $64.8{\pm}5.9$ & $59.8{\pm}5.7$
      & $\mathbf{72.9{\pm}7.7}$ & $\mathbf{73.9{\pm}7.0}$
      & $38{\pm}3$ \\
    \dsBiliary{} (KW)
      & $36/803/3$
      & $71.7{\pm}6.1$ & $83.3{\pm}5.0$
      & $\mathbf{90.0{\pm}4.4}$ & $\mathbf{96.7{\pm}2.2}$
      & $62{\pm}38$ \\
    \bottomrule
    \end{tabular}
    \end{table}

We also stress-test the pipeline, either by adding independent noise to every entry, scaled to each feature's own standard deviation, or by appending many pure-noise (useless) features. With added noise, \DPHCT{}-LoR substantially improves over the baseline at small noise levels in most cases. The improvement also holds up when a substantial number of pure-noise features is appended; the details and the results are in Section~\ref{app:real-data-stress} of the Supplementary Material.

\section{Conclusion}
\DPHCT{} extends higher-criticism feature screening to multiclass problems in
which useful features are rare and individually weak. The theory identifies a
multiclass phase transition that depends explicitly on the growth of the number
of classes, and shows that diversity-based HC screening attains this boundary
with a simple centroid-similarity classifier. Simulations confirm the predicted
transition, while real-data benchmarks and stress tests show that the same
screening principle can substantially improve classification when irrelevant or
noisy coordinates dominate the feature space.

\bibliographystyle{plainnat}
\bibliography{refs}

\begin{thebibliography}{33}
\providecommand{\natexlab}[1]{#1}
\providecommand{\url}[1]{\texttt{#1}}
\expandafter\ifx\csname urlstyle\endcsname\relax
  \providecommand{\doi}[1]{doi: #1}\else
  \providecommand{\doi}{doi: \begingroup \urlstyle{rm}\Url}\fi

\bibitem[Abramovich and Pensky(2019)]{abramovich2019classification}
Felix Abramovich and Marianna Pensky.
\newblock Classification with many classes: Challenges and pluses.
\newblock \emph{Journal of Multivariate Analysis}, 174:\penalty0 104536, 2019.
\newblock \doi{10.1016/j.jmva.2019.104536}.

\bibitem[Abramovich et~al.(2021)Abramovich, Grinshtein, and
  Levy]{abramovich2021multiclass}
Felix Abramovich, Vadim Grinshtein, and Tomer Levy.
\newblock Multiclass classification by sparse multinomial logistic regression.
\newblock \emph{IEEE Transactions on Information Theory}, 67\penalty0
  (7):\penalty0 4637--4646, 2021.

\bibitem[Azar and Nadler(2024)]{azar2024semi}
Eyar Azar and Boaz Nadler.
\newblock Semi-supervised sparse gaussian classification: Provable benefits of
  unlabeled data.
\newblock \emph{Advances in Neural Information Processing Systems},
  37:\penalty0 20132--20169, 2024.

\bibitem[Baik et~al.(2005)Baik, Ben~Arous, and P{\'e}ch{\'e}]{baik2005phase}
Jinho Baik, G{\'e}rard Ben~Arous, and Sandrine P{\'e}ch{\'e}.
\newblock Phase transition of the largest eigenvalue for nonnull complex sample
  covariance matrices.
\newblock \emph{The Annals of Probability}, 33\penalty0 (5):\penalty0
  1643--1697, 2005.
\newblock \doi{10.1214/009117905000000233}.

\bibitem[Benjamini and Hochberg(1995)]{benjamini1995controlling}
Yoav Benjamini and Yosef Hochberg.
\newblock Controlling the false discovery rate: A practical and powerful
  approach to multiple testing.
\newblock \emph{Journal of the Royal Statistical Society, Series B},
  57\penalty0 (1):\penalty0 289--300, 1995.

\bibitem[Bickel et~al.(2010)Bickel, Ritov, and
  Tsybakov]{bickel2010hierarchical}
Peter~J Bickel, Ya’acov Ritov, and Alexandre~B Tsybakov.
\newblock Hierarchical selection of variables in sparse high-dimensional
  regression.
\newblock In \emph{Borrowing strength: theory powering applications--a
  Festschrift for Lawrence D. Brown}, volume~6, pages 56--70. Institute of
  Mathematical Statistics, 2010.

\bibitem[Bray et~al.(2008)Bray, Shaw, Moore, Garcia, Zanquetta, Durgan, Jeong,
  Tsai, Bugger, Zhang, et~al.]{bray2008circadian}
Molly~S. Bray, Chad~A. Shaw, Michael W.~S. Moore, Rebeca A.~P. Garcia,
  Marcelo~M. Zanquetta, David~J. Durgan, Woo~Jung Jeong, Ju-Yun Tsai, Heiko
  Bugger, Dongfang Zhang, et~al.
\newblock Disruption of the circadian clock within the cardiomyocyte influences
  myocardial contractile function, metabolism, and gene expression.
\newblock \emph{American Journal of Physiology-Heart and Circulatory
  Physiology}, 294\penalty0 (2):\penalty0 H1036--H1047, 2008.

\bibitem[Butucea and Ingster(2013)]{butucea2013detection}
Cristina Butucea and Yuri~I. Ingster.
\newblock Detection of a sparse submatrix of a high-dimensional noisy matrix.
\newblock \emph{Bernoulli}, 19\penalty0 (5B):\penalty0 2652--2688, 2013.

\bibitem[Coolon et~al.(2009)Coolon, Jones, Todd, Carr, and
  Herman]{coolon2009elegans}
Joseph~D. Coolon, Kenneth~L. Jones, Timothy~C. Todd, Bryan~C. Carr, and
  Michael~A. Herman.
\newblock {Caenorhabditis elegans} genomic response to soil bacteria predicts
  environment-specific genetic effects on life history traits.
\newblock \emph{PLoS Genetics}, 5\penalty0 (6):\penalty0 e1000503, 2009.

\bibitem[Donoho and Jin(2004)]{donoho2004higher}
David Donoho and Jiashun Jin.
\newblock Higher criticism for detecting sparse heterogeneous mixtures.
\newblock \emph{Annals of Statistics}, 32\penalty0 (3):\penalty0 962--994,
  2004.

\bibitem[Donoho and Jin(2008)]{donoho2008higher}
David Donoho and Jiashun Jin.
\newblock Higher criticism thresholding: Optimal feature selection when useful
  features are rare and weak.
\newblock \emph{Proceedings of the National Academy of Sciences}, 105\penalty0
  (39):\penalty0 14790--14795, 2008.

\bibitem[Donoho and Jin(2015)]{donoho2015higher}
David Donoho and Jiashun Jin.
\newblock Higher criticism for large-scale inference, especially for rare and
  weak effects.
\newblock \emph{Statistical Science}, 30\penalty0 (1):\penalty0 1--25, 2015.

\bibitem[Donoho and Jin(2009)]{donoho2009feature}
David~L. Donoho and Jiashun Jin.
\newblock Feature selection by higher criticism thresholding achieves the
  optimal phase diagram.
\newblock \emph{Philosophical Transactions of the Royal Society A:
  Mathematical, Physical and Engineering Sciences}, 367\penalty0
  (1906):\penalty0 4449--4470, 2009.

\bibitem[El~Alaoui et~al.(2020)El~Alaoui, Krzakala, and
  Jordan]{elalaoui2020fundamental}
Ahmed El~Alaoui, Florent Krzakala, and Michael~I. Jordan.
\newblock Fundamental limits of detection in the spiked {Wigner} model.
\newblock \emph{The Annals of Statistics}, 48\penalty0 (2):\penalty0 863--885,
  2020.
\newblock \doi{10.1214/19-AOS1826}.

\bibitem[Fan and Fan(2008)]{fan2008high}
Jianqing Fan and Yingying Fan.
\newblock High dimensional classification using features annealed independence
  rules.
\newblock \emph{Annals of statistics}, 36\penalty0 (6):\penalty0 2605, 2008.

\bibitem[Ferguson(2022)]{ferguson2022tips}
Jane Ferguson.
\newblock {TIPs} metabolomics ({Metabolomics Workbench} study {ST002019},
  project {PR001282}).
\newblock NIH Common Fund National Metabolomics Data Repository, Metabolomics
  Workbench, 2022.
\newblock Vanderbilt University Medical Center.
  \url{https://doi.org/10.21228/M80Q51}.

\bibitem[Guo(2025)]{guo2025bpf}
Qiuyuan Guo.
\newblock Metabolomics analysis of zebrafish liver under long-term {BPF}
  exposure ({Metabolomics Workbench} study {ST003804}, project {PR002378}).
\newblock NIH Common Fund National Metabolomics Data Repository, Metabolomics
  Workbench, 2025.
\newblock Washington University in St. Louis.
  \url{https://doi.org/10.21228/M87Z5M}.

\bibitem[Ingster et~al.(2009)Ingster, Pouet, and
  Tsybakov]{ingster2009classification}
Yuri~I Ingster, Christophe Pouet, and Alexandre~B Tsybakov.
\newblock Classification of sparse high-dimensional vectors.
\newblock \emph{Philosophical Transactions of the Royal Society A:
  Mathematical, Physical and Engineering Sciences}, 367\penalty0
  (1906):\penalty0 4427--4448, 2009.

\bibitem[Jin(2009)]{jin2009impossibility}
Jiashun Jin.
\newblock Impossibility of successful classification when useful features are
  rare and weak.
\newblock \emph{Proceedings of the National Academy of Sciences}, 106\penalty0
  (22):\penalty0 8859--8864, 2009.

\bibitem[Jin and Wang(2016)]{jin2016influential}
Jiashun Jin and Wanjie Wang.
\newblock Influential features pca for high dimensional clustering.
\newblock \emph{The Annals of Statistics}, pages 2323--2359, 2016.

\bibitem[Jin et~al.(2017)Jin, Ke, and Wang]{jin2017phase}
Jiashun Jin, Zheng~Tracy Ke, and Wanjie Wang.
\newblock Phase transitions for high dimensional clustering and related
  problems.
\newblock \emph{The Annals of Statistics}, 45\penalty0 (5):\penalty0
  2151--2189, 2017.

\bibitem[Kipnis(2025)]{kipnis2025unification}
Alon Kipnis.
\newblock Unification of rare and weak multiple testing models using moderate
  deviations analysis and log-chisquared p-values.
\newblock \emph{Statistica Sinica}, 35:\penalty0 1155--1181, 2025.

\bibitem[Krawczyk(2016)]{krawczyk2016learning}
Bartosz Krawczyk.
\newblock Learning from imbalanced data: open challenges and future directions.
\newblock \emph{Progress in artificial intelligence}, 5\penalty0 (4):\penalty0
  221--232, 2016.

\bibitem[Kruskal and Wallis(1952)]{kruskal1952use}
William~H. Kruskal and W.~Allen Wallis.
\newblock Use of ranks in one-criterion variance analysis.
\newblock \emph{Journal of the American Statistical Association}, 47\penalty0
  (260):\penalty0 583--621, 1952.

\bibitem[Lehmann and Romano(2006)]{lehmann2006testing}
Erich~L Lehmann and Joseph~P Romano.
\newblock \emph{Testing statistical hypotheses}.
\newblock Springer Science \& Business Media, 2006.

\bibitem[Loeffler et~al.(2020)Loeffler, Hu, Kirchner, Wei, Xiao, Albrecht,
  Torre, Sticht, et~al.]{loeffler2020mirna}
Markus~A. Loeffler, Jing Hu, Marcus Kirchner, Xiaoyu Wei, Yue Xiao, Thomas
  Albrecht, Carolina De~La Torre, Carsten Sticht, et~al.
\newblock {miRNA} profiling of biliary intraepithelial neoplasia reveals
  stepwise tumorigenesis in distal cholangiocarcinoma via the {miR-451a/ATF2}
  axis.
\newblock \emph{The Journal of Pathology}, 252\penalty0 (3):\penalty0 239--251,
  2020.

\bibitem[Loureiro et~al.(2021)Loureiro, Sicuro, Gerbelot, Pacco, Krzakala, and
  Zdeborov{\'a}]{loureiro2021learning}
Bruno Loureiro, Gabriele Sicuro, C{\'e}dric Gerbelot, Alessandro Pacco, Florent
  Krzakala, and Lenka Zdeborov{\'a}.
\newblock Learning gaussian mixtures with generalized linear models: Precise
  asymptotics in high-dimensions.
\newblock \emph{Advances in Neural Information Processing Systems},
  34:\penalty0 10144--10157, 2021.

\bibitem[Ma et~al.(2026)Ma, Wang, and Fan]{ma2026if}
Chen Ma, Wanjie Wang, and Shuhao Fan.
\newblock i-if-learn: Iterative feature selection and unsupervised learning for
  high-dimensional complex data.
\newblock \emph{arXiv preprint arXiv:2603.24025}, 2026.

\bibitem[Perry et~al.(2018)Perry, Wein, Bandeira, and
  Moitra]{perry2018optimality}
Amelia Perry, Alexander~S. Wein, Afonso~S. Bandeira, and Ankur Moitra.
\newblock Optimality and sub-optimality of {PCA I}: Spiked random matrix
  models.
\newblock \emph{The Annals of Statistics}, 46\penalty0 (5):\penalty0
  2416--2451, 2018.
\newblock \doi{10.1214/17-AOS1625}.

\bibitem[Stoepker et~al.(2024)Stoepker, Castro, Arias-Castro, and van~den
  Heuvel]{stoepker2024anomaly}
Ivo~V Stoepker, Rui~M Castro, Ery Arias-Castro, and Edwin van~den Heuvel.
\newblock Anomaly detection for a large number of streams: A permutation-based
  higher criticism approach.
\newblock \emph{Journal of the American Statistical Association}, 119\penalty0
  (545):\penalty0 461--474, 2024.

\bibitem[Sud et~al.(2016)Sud, Fahy, Cotter, Azam, Vadivelu, Burant, Edison,
  Fiehn, Higashi, Nair, Sumner, and Subramaniam]{sud2016metabolomics}
Manish Sud, Eoin Fahy, Dawn Cotter, Kenan Azam, Ilango Vadivelu, Charles
  Burant, Arthur Edison, Oliver Fiehn, Richard Higashi, K.~Sreekumaran Nair,
  Susan Sumner, and Shankar Subramaniam.
\newblock {Metabolomics Workbench}: An international repository for
  metabolomics data and metadata, metabolite standards, protocols, tutorials
  and training, and analysis tools.
\newblock \emph{Nucleic Acids Research}, 44\penalty0 (D1):\penalty0 D463--D470,
  2016.
\newblock \doi{10.1093/nar/gkv1042}.

\bibitem[Thrampoulidis et~al.(2020)Thrampoulidis, Oymak, and
  Soltanolkotabi]{thrampoulidis2020theoretical}
Christos Thrampoulidis, Samet Oymak, and Mahdi Soltanolkotabi.
\newblock Theoretical insights into multiclass classification: A
  high-dimensional asymptotic view.
\newblock \emph{Advances in Neural Information Processing Systems},
  33:\penalty0 8907--8920, 2020.

\bibitem[Verzelen and Arias-Castro(2017)]{verzelen2017detection}
Nicolas Verzelen and Ery Arias-Castro.
\newblock Detection and feature selection in sparse mixture models.
\newblock \emph{Annals of Statistics}, 45\penalty0 (5):\penalty0 1920--1950,
  2017.
\newblock \doi{10.1214/16-AOS1513}.
\newblock URL \url{https://hal.inrae.fr/hal-02624460}.

\end{thebibliography}

\clearpage

\appendix
% -------- Local ToC for the appendix --------
\renewcommand{\contentsname}{Supplementary Material: Contents}
\setcounter{tocdepth}{2}
\startcontents[appendix]
\printcontents[appendix]{}{1}{\section*{Supplementary Material: Contents}\vspace{-0.5em}}
\vspace{1em}

\noindent

% ================================================

%% ---------------------------------------------------------------
\section{Synthetic-Data Simulations: Additional Figures}
\label{app:rw-simulations}
%% ---------------------------------------------------------------

All figures in this section use the Gaussian multiclass Rare-and-Weak Useful Feature model $\MCRW(p,C,\epsilon,\tau)$ with $n_{\mathrm{Monte}} = 150$ Monte Carlo draws per
$(\beta, r)$ grid point, unless stated otherwise.
The ideal threshold for $\MCRW(p,C,\epsilon,\tau)$, HCT, BH-FDR ($q=0.10$), and Bonferroni ($\alpha=0.05$) selectors are all followed by a centroid similarity classifier. Phase boundaries overlaid on heatmaps are the multiclass boundary
\(r=\xi^\star(\beta,s)\) (solid) and the fixed-number-of-classes boundary of \cite{donoho2008higher} (dotted).

%% --- (1) Ideal-threshold misclassification rate heatmaps ---
\begin{figure}[!htbp]
  \centering
  \includegraphics[width=\linewidth]{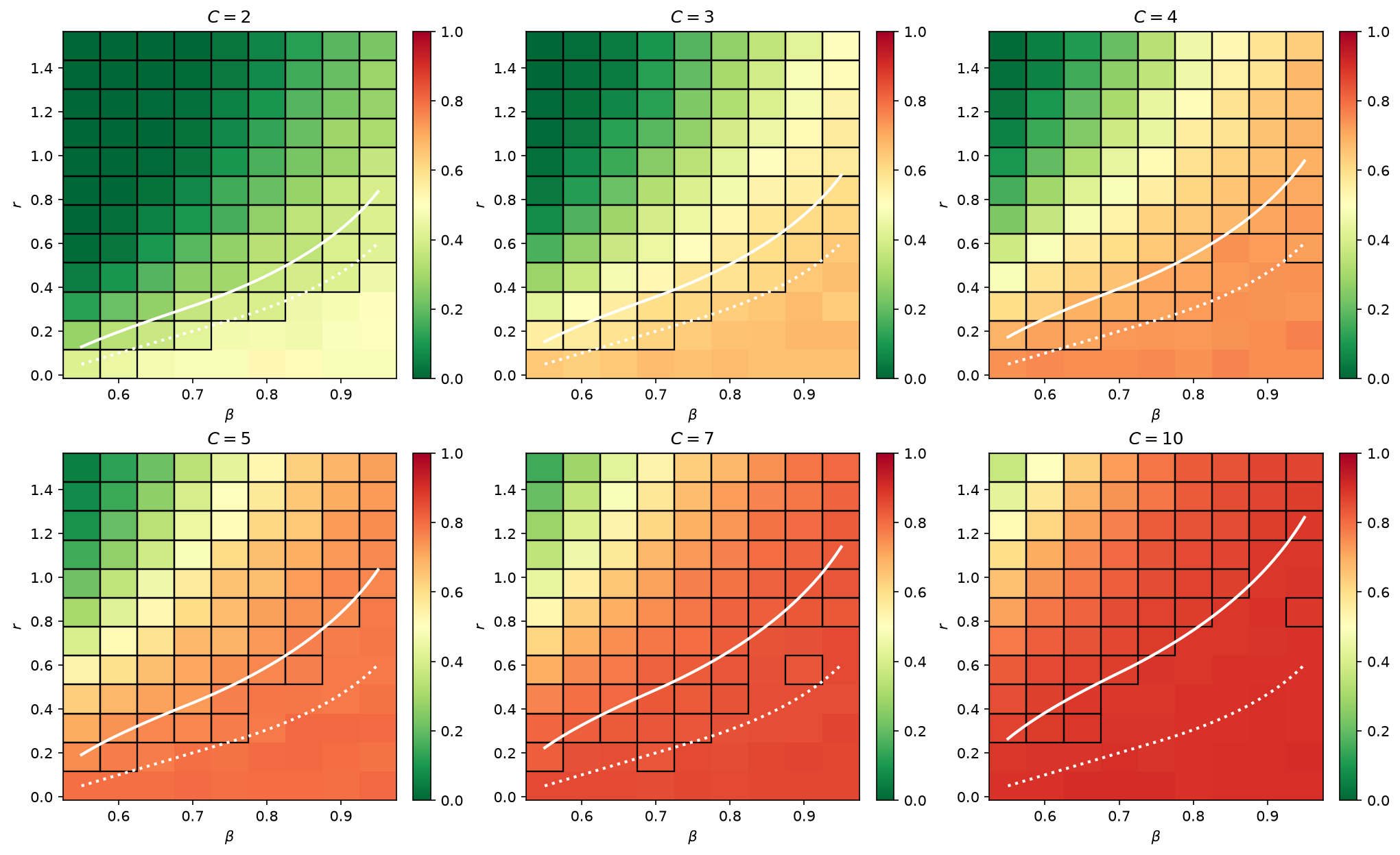}
  \caption{%
    \textbf{Ideal-threshold misclassification rate heatmaps} for
    \(C\in\{2,3,4,5,7,10\}\). Color encodes misclassification rate. Solid white: the multiclass boundary
    \(r=\xi^\star(\beta,s)\) of~\eqref{eq:xi-boundary}. Dotted white: the fixed-class Donoho--Jin
    boundary \(r=\rho^\star(\beta)\). The transition from high to low error closely tracks the multiclass boundary; as \(C\) increases, the boundary rises, reflecting the harder multiclass discrimination problem.
  }
  \label{fig:ideal_error_heatmap}
\end{figure}

%% --- (5) Empirical FDR of the threshold rules (replaces the HCT-minus-ideal heatmaps) ---
\begin{figure}[!htbp]
  \centering
  \includegraphics[width=\linewidth]{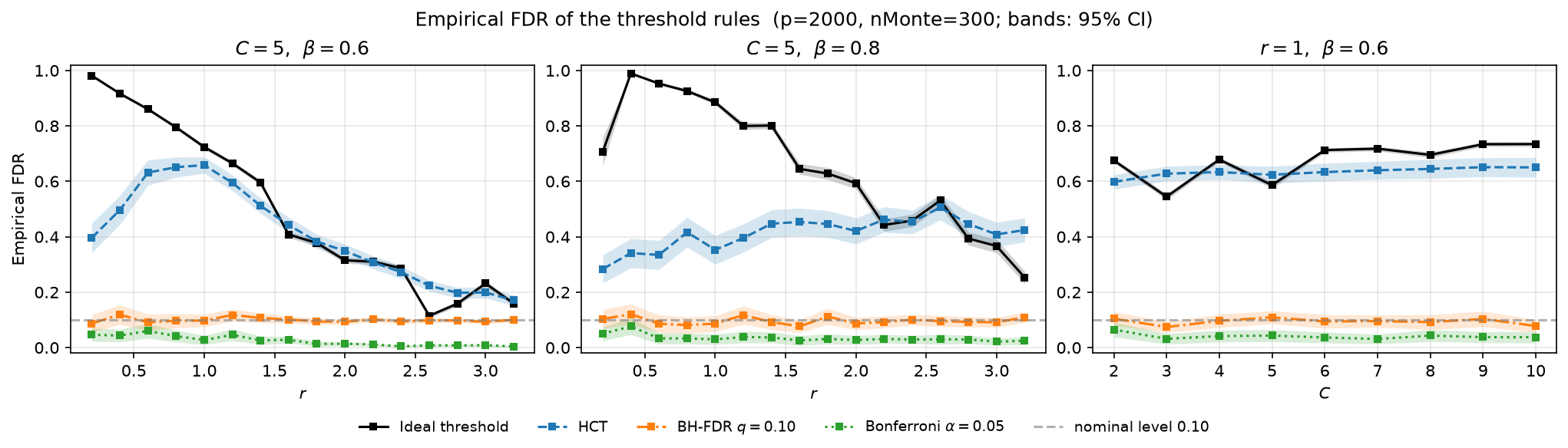}
  \caption{%
    \textbf{Empirical false-discovery rate of the threshold rules.}
    The empirical FDR is the average, over Monte Carlo draws, of the proportion of selected features that are null.
    \emph{Left}: FDR vs.\ $r$ for $C=5$, $\beta=0.6$.
    \emph{Middle}: FDR vs.\ $r$ for $C=5$, $\beta=0.8$.
    \emph{Right}: FDR vs.\ the number of classes $C\in\{2,\dots,10\}$ for $r=1$, $\beta=0.6$.
    Solid black: ideal threshold of~\eqref{eq:ideal-threshold}; dashed blue: \DPHCT{}; dash-dot orange: Benjamini--Hochberg's False-Discovery Rate ($q=0.10$); dotted green: Bonferroni ($\alpha=0.05$); dashed grey: the nominal level $0.10$.
    Shaded bands are $95\%$ confidence intervals of the Monte Carlo mean
    ($p = 2{,}000$, $n_{\mathrm{Monte}} = 300$).
    Draws in which \DPHCT{} does not reject the global null (HC statistic below $1.9$) select no features and count as an FDR of zero. 
    The figure shows that \DPHCT{} closely tracks the ideal threshold in the regime where detection is possible. These two rules are tuned for classification rather than for error control, and therefore admit many false discoveries. BH attains its nominal level and Bonferroni is conservative.
  }
  \label{fig:compare_diff}
\end{figure}

%% ---- Previous version of this figure (HCT minus ideal misclassification rate), kept for comparison ----
% %% --- (5) HCT minus Ideal misclassification rate ---
% \begin{figure}[!htbp]
%   \centering
%   \includegraphics[width=0.75\linewidth]{figures/compare_error_diff_heatmaps_p2000_rademacher_nmc150}
%   \caption{%
%     \textbf{HCT minus ideal misclassification rate} for
%     $C \in \{2,3,5,10\}$.
%     The diverging color scale is centered at zero (white); red (positive)
%     cells show where HCT incurs excess error relative to the oracle.
%     The excess is concentrated near and below the theoretical multiclass boundary of Theorem~\ref{thm:classification} (dashed black), confirming that HCT is near-optimal when the signal is clearly detectable. Dotted black: The phase transition curve corresponding to a fixed number of classes, derived in \cite{donoho2008higher}. 
%   }
%   \label{fig:compare_diff}
% \end{figure}
%

\FloatBarrier
\section{Real-Data Experimental Details}
\label{app:real-data}

\subsection{Datasets}
\label{app:real-data-datasets}

The 5 datasets are listed in 
Table~\ref{tab:class-balance}. 
\begin{table}[h]
\centering
\caption{Five real-data benchmarks, their sizes, and additional relevant information. ``Units'' is the number of
independent patients, participants, animals or arrays (equal to $n$ when samples are independent).}
\label{tab:class-balance}
\scriptsize
\setlength{\tabcolsep}{2pt}
\begin{tabularx}{\textwidth}{@{}>{\raggedright\arraybackslash}p{2.0cm} >{\raggedright\arraybackslash}X ccc >{\centering\arraybackslash}p{1.5cm} c@{}}
\toprule
Dataset & Description & $n$ & $p$ & $C$ & Class sizes & Units \\
\midrule
\dsHeart{} (GSE10045) \citep{bray2008circadian} & Cardiac expression in wild-type and transgenic mice sampled every three hours. Classes are Zeitgeber times; genotypes and cardiac chambers are pooled. Units are animals. & $130$ & $46{,}120$ & $8$ & $19/\allowbreak17/\allowbreak16/\allowbreak16/\allowbreak16/\allowbreak16/\allowbreak16/\allowbreak14$ & $66$ \\
\dsWorm{} (GSE15923) \citep{coolon2009elegans} & Two-color expression arrays from \emph{C.\ elegans} grown on four bacterial environments, which define the classes. Units are arrays, each contributing two channel samples. & $72$ & $672$ & $4$ & $18/\allowbreak18/\allowbreak18/\allowbreak18$ & $38$ \\
\dsBlood{} (ST002019) \citep{ferguson2022tips} & Blood LC--MS metabolomics. Classes combine diet (pre/post) and study phase (1/2). Units are participants, each sampled at two visits. & $70$ & $736$ & $4$ & $18/\allowbreak18/\allowbreak17/\allowbreak17$ & $35$ \\
\dsZebrafish{} (ST003804) \citep{guo2025bpf} & Zebrafish liver LC--MS metabolomics after long-term bisphenol~F exposure. Classes are four exposure groups; samples are independent fish. & $62$ & $117$ & $4$ & $16/\allowbreak16/\allowbreak15/\allowbreak15$ & $62$ \\
\dsBiliary{} (GSE140001) \citep{loeffler2020mirna} & miRNA expression in distal cholangiocarcinoma. Classes are three tissue types; each patient contributes one sample of each type and is a unit. & $36$ & $803$ & $3$ & $12/\allowbreak12/\allowbreak12$ & $12$ \\
\bottomrule
\end{tabularx}
\end{table}

\paragraph{Preprocessing.}
We use the matrices provided by the repositories (GEO series matrices, Metabolomics Workbench data tables \citep{sud2016metabolomics}). Features with fewer than $95\%$ finite values are removed, remaining missing values are imputed by the feature median, $\log_2(x+1)$ is applied when the matrix is non-negative with maximum above $100$, and constant features are dropped. The values of $p$ reported below are counted after these steps. No other preprocessing is applied beyond per-fold standardization inside the classifiers.

\subsection{Cross-validation protocol}
\label{app:real-data-cv}

Classification accuracy results use stratified CV with $K=10$ folds ($K=7$ for \dsHeart{}, the largest number of folds that leaves at least one unit of every class in each fold). Table~\ref{tab:clean-full} reports averages over five CV runs with different seeds. Stress-test results use $K=10$ folds with three repeats ($K=7$ with a single repeat for \dsHeart{}, whose stress runs are the slowest).

We use stratified group folds when a patient, participant, animal or array contributes several samples, so that all samples of a unit are in the same fold and no unit appears in both a training and a test fold. Otherwise the repeated samples of a unit leak across the split and inflate the accuracy of both the baseline and the screened classifier. The units are listed in Table~\ref{tab:class-balance}.

\subsection{\DPHCT{} pipeline}
\label{app:real-data-pipeline}

We executes the following steps within each fold on training data only.

\begin{enumerate}[nosep, left=0pt]
  \item \textbf{Diversity p-values.}
    Compute the ANOVA F or Kruskal--Wallis statistic per feature on the training split, and convert it to a p-value $\pi_j$
    from the $F_{C-1,n-C}$ or $\chi^2_{C-1}$ reference distribution.
    p-values are clipped to $[10^{-300},\,1-10^{-15}]$.
    The lower clip exists only to keep exact zeros out of the HC denominator and must lie below the smallest p-value the data produce.
  \item \textbf{HC thresholding (HCT).}
    Obtain $j^\star$ of~\eqref{eq:HC-statistic} and 
    set $\hat{\mathcal{J}} = \{j:\pi_j\le \pi_{(j^\star)} \}$.
    
  \item \textbf{Classifiers.}
    Two plug-in classifiers are evaluated on $\hat{\mathcal{J}}$:
    \begin{itemize}[nosep]
      \item \emph{SVM}: RBF-kernel SVC (\texttt{sklearn}, \texttt{random\_state=0}),
        features standardized with \texttt{StandardScaler}.
      \item \emph{Logistic}: $\ell_2$-penalized multinomial logistic regression
        (\texttt{saga} solver).
    \end{itemize}
\end{enumerate}

\begin{table}[h]
    \centering
    \caption{%
      CV accuracy (mean $\pm$ SEM, \%) for the five datasets, averaged over five CV runs with different seeds
      (the SEM is the mean within-run SEM across folds).
      The diversity test is given next to each dataset (F = ANOVA F, KW = Kruskal--Wallis); raw p-values; HC with $\gamma_0=1/2$.
      ``Frac.\ sel.'' = mean fraction of features selected by HC.
    }
    \label{tab:clean-full}
    \footnotesize
    \setlength{\tabcolsep}{4pt}
    \begin{tabular}{ll cc c}
    \toprule
    Dataset & Method & SVM & LoR & Frac.\ sel. \\
    \midrule
    \multirow{2}{*}{\dsHeart{} (F)}
      & Baseline & $13.6{\pm}1.9$ & $29.9{\pm}3.5$ & --- \\
      & \DPHCT{} & $\mathbf{69.3{\pm}5.5}$ & $\mathbf{75.2{\pm}4.7}$ & $1.0\%$ \\
    \midrule
    \multirow{2}{*}{\dsWorm{} (KW)}
      & Baseline & $22.9{\pm}4.2$ & $71.5{\pm}5.5$ & --- \\
      & \DPHCT{} & $\mathbf{55.2{\pm}6.1}$ & $\mathbf{80.3{\pm}5.6}$ & $8.9\%$ \\
    \midrule
    \multirow{2}{*}{\dsBlood{} (F)}
      & Baseline & $41.0{\pm}5.0$ & $58.2{\pm}5.3$ & --- \\
      & \DPHCT{} & $\mathbf{76.9{\pm}5.4}$ & $\mathbf{80.1{\pm}4.0}$ & $14.6\%$ \\
    \midrule
    \multirow{2}{*}{\dsZebrafish{} (KW)}
      & Baseline & $62.0{\pm}5.9$ & $60.8{\pm}6.2$ & --- \\
      & \DPHCT{} & $\mathbf{70.1{\pm}5.0}$ & $\mathbf{71.3{\pm}5.5}$ & $33.7\%$ \\
    \midrule
    \multirow{2}{*}{\dsBiliary{} (KW)}
      & Baseline & $65.0{\pm}5.1$ & $83.0{\pm}5.0$ & --- \\
      & \DPHCT{} & $\mathbf{91.0{\pm}4.5}$ & $\mathbf{98.7{\pm}1.1}$ & $8.5\%$ \\
    \bottomrule
    \end{tabular}
\end{table}

\subsection{Stress-test protocol}
\label{app:real-data-stress}

\begin{figure}[p]
  \centering
  \includegraphics[width=0.86\linewidth]{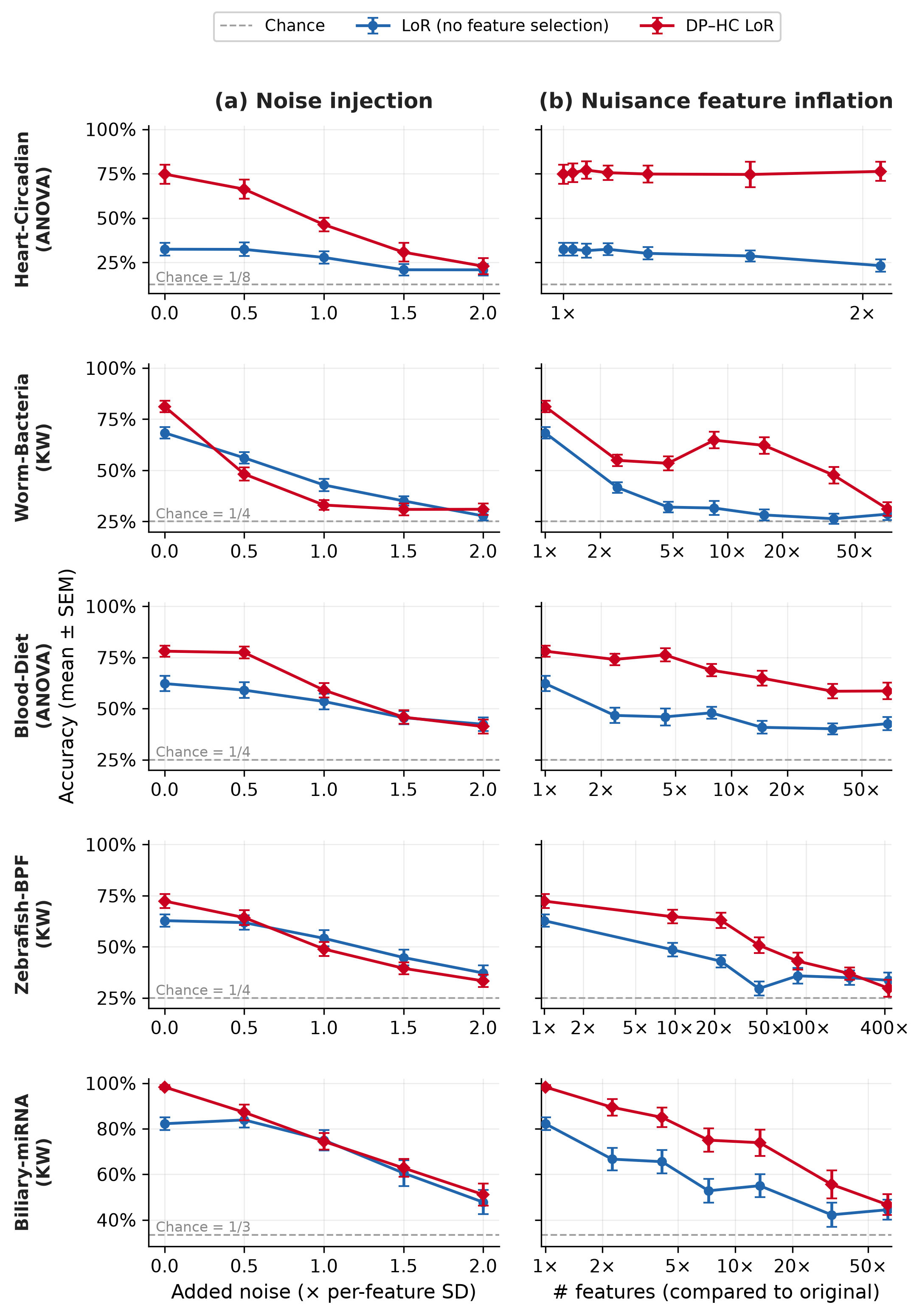}
  \caption{
    Classification with LoR pre- and post-feature screening by \DPHCT{} with added noise (left column) or nuisance features (right column). p-values are obtained either from ANOVA F (ANOVA) or Kruskal--Wallis (KW) as indicated for each panel; HC with $\gamma_0=1/2$;  
    error bars indicate $\pm$\,SEM over the pooled CV folds ($10$ folds $\times$ $3$ repeats; $7$ folds for \dsHeart{}); the uniform chance level $1/C$ is shown as a dashed horizontal line.
  }
  \label{fig:stress-real-data}
\end{figure}

We perturb each of the five datasets by one of:
\begin{itemize}[nosep, left=0pt]
  \item \emph{Noise injection} ($\sigma$ sweep): independent
    $\mathcal{N}(0,\sigma^2)$ noise is added per fold to every element of
    $X_{\mathrm{train}}$ and $X_{\mathrm{test}}$ for
    $\sigma \in \{0, 0.5, 1.0, 1.5, 2.0\}$.
  \item \emph{Nuisance feature inflation} ($p_{\mathrm{extra}}$ sweep):
    $p_{\mathrm{extra}} \in
      \{0,\,10^3,\,2.5\!\times\!10^3,\,5\!\times\!10^3,\,10^4,\,
        2.5\!\times\!10^4,\,5\!\times\!10^4\}$
    additional $\mathcal{N}(0,1)$ columns are concatenated per fold.
\end{itemize}
Both perturbations occur after the train/test split and before feature selection, so HC is never exposed to test-fold data. Both perturbations are regenerated independently for each fold, as explained below in more detail. The results are illustrated in Fig.~\ref{fig:stress-real-data}. 

\paragraph{Noise injection.}
$X_{ij} \leftarrow X_{ij} + \sigma\,\hat{s}_j\,\varepsilon_{ij}$ with
$\varepsilon_{ij}$ i.i.d.\ $\mathcal{N}(0,1)$ drawn fresh per fold, where
$\hat{s}_j$ is the standard deviation of feature $j$ on the training fold.
Grid: $\sigma \in \{0.0,\,0.5,\,1.0,\,1.5,\,2.0\}$.
The noise is scaled relative to each feature's own spread because the
panels differ in units (the median feature standard deviation ranges from $0.3$ on \dsBlood{} to $511$ on \dsWorm{}). Scales are estimated on the training fold only to prevent test information leakage.

\paragraph{Nuisance feature inflation.}
$X \leftarrow [X,\,G]$ where $G$ has i.i.d.\ $\mathcal{N}(0,1)$ entries.
Grid: $p_{\mathrm{extra}} \in \{0,\,10^3,\,2.5\!\times\!10^3,\,5\!\times\!10^3,\,
10^4,\,2.5\!\times\!10^4,\,5\!\times\!10^4\}$.
Because the grid is absolute while $p$ varies across panels, the resulting feature ratio $(p+p_{\mathrm{extra}})/p$ depends on the size of the dataset.

\section{Notation and auxiliary facts}
\label{app:overview}

We collect the notation used throughout the proofs and record several auxiliary facts about the chi-squared distribution that will be invoked repeatedly.

\subsection{Notation}
Throughout, $p \to \infty$ indexes the asymptotic regime, and all probabilistic statements are understood in this limit unless stated otherwise. We write $a_p = O(b_p)$ if $|a_p|/|b_p|$ is eventually bounded, $a_p = o(b_p)$ if $a_p/b_p \to 0$, and $a_p \asymp b_p$ if both $a_p = O(b_p)$ and $b_p = O(a_p)$. Probabilistic versions $O_\Prob$, $o_\Prob$ are defined analogously in probability. For a set $Q$, we write $a_p(q) = o(b_p(q))$ \emph{uniformly in $q \in Q$} if $\sup_{q \in Q} |a_p(q)/b_p(q)| \to 0$.

We use $C, K, c, c'$ to denote positive constants that may change from line to line but do not depend on $p$ unless explicitly indicated.

\subsection{Model and objects}
%We observe $\{(X_i, Y_i)\}_{i=1}^n$ with $X_i \in \R^p$, $Y_i \in [C]$, balanced classes $n_c = n/C$, and conditional distribution $X_i \mid Y_i = c \sim \mathcal{N}(\mu_c, \sigma^2 I_p)$. The signal set $\mathcal{J} \subseteq [p]$ has $|\mathcal{J}| = p^{1-\beta}$ with $\beta \in (1/2, 1)$. On $\mathcal{J}$, the theoretical results assume the homogeneous ANOVA non-centrality calibration~\eqref{eq:homogeneous-nc}. This is achieved exactly by the centered-normalized construction~\eqref{eq:rademacher} with calibration~\eqref{eq:tau-calibration}, and asymptotically in Regime~B by the plain mean-calibrated Rademacher construction~\eqref{eq:plain-rademacher}--\eqref{eq:plain-tau-calibration}. Outside of $\mathcal{J}$, $\mu_{1,j} = \cdots = \mu_{C,j} = 0$ w.l.o.g.

We work under the Gaussian multiclass rare/weak model of
Section~\ref{sec:model-setup}. Thus
\[
    X_i\mid Y_i=c\sim \mathcal N(\mu_c,\sigma^2I_p),
    \qquad n_c=n/C,
\]
and the set of useful features \(\mathcal J\subset[p]\) has rarity
\(\epsilon_p=p^{-\beta}\), \(\beta\in(1/2,1)\). On \(\mathcal J\), the
spherical homogeneous contrast construction
\eqref{eq:spherical-contrast}--\eqref{eq:lambda-calibration} gives
 $\lambda_j = 2r\log p$ for $j\in\mathcal J$ and $\lambda_j = 0$ for $j\notin\mathcal J$. Class-\(c\) sample means are \(\hat{\mu}_{c,j}\) as in \eqref{eq:mu_hat}, and
\[
    \hat{\mu}_{\cdot,j}
    :=
    (\hat{\mu}_{1,j},\ldots,\hat{\mu}_{C,j})^\top
\]
is the vector of those means at feature \(j\). The model means are the superscript coordinates \(\mu_c^{(j)}\) of \(\mu^{(j)}=(\mu_1^{(j)},\ldots,\mu_C^{(j)})^\top\).

%For $j \in [p]$, let $\bar X_{c,j} := n_c^{-1}\sum_{i : Y_i = c} X_{i,j}$, $\bar X_j := C^{-1}\sum_c \bar X_{c,j}$, and $P := I_C - C^{-1}\mathbf{1}\mathbf{1}^\top$. The ANOVA statistic is $T_j := \mathrm{SS}_\mathrm{bet}(j)/\sigma^2$, and its ANOVA p-value is $\pi_j := \overline F_{\chi^2_k}(T_j)$ with $k := C-1$. We let $Y_j := -2\log \pi_j = G_k(T_j)$ with $G_k(t) := -2\log \overline F_{\chi^2_k}(t)$.

%For \(j\in[p]\), let
%\[
%    \bar X_{c,j}:=n_c^{-1}\sum_{i:Y_i=c}X_{i,j},
%    \qquad
%    \bar X_j:=C^{-1}\sum_{c=1}^C\bar X_{c,j},
%    \qquad
%    P:=I_C-C^{-1}\mathbf 1\mathbf 1^\top .
%\]
For the numerator-based analysis, define
\[
    T_j:=\frac{\mathrm{SS}_{\mathrm{between}}(j)}{\sigma^2},
    \qquad
    \pi_j^\chi:=\overline F_{\chi^2_k}(T_j),
    \qquad
    k:=C-1,
\]
and
\[
    Y_j^\chi:=-2\log\pi_j^\chi
    =
    G_k(T_j),
    \qquad
    G_k(t):=-2\log\overline F_{\chi^2_k}(t).
\]
% The actual algorithm uses the ANOVA F-test p-value defined in
% \eqref{eq:F-statistic}. Section~\ref{app:anova-f-reduction} of the Supplementary Material shows that, on the moderate-deviation scale relevant for the theory, the F-test p-value and the oracle numerator p-value \(\pi_j^\chi\) have the same tail exponent.

\subsection{Chi-squared tail: quantitative expansion}
The following one-sided tail expansion records useful auxiliary estimates. The exact exponent in Theorem~\ref{thm:md} is proved separately by exponential tilting.

\begin{lemma}[Chi-squared tail expansion]
\label{lem:chi2-tail}
Let $\overline F_k(t):=\mathbb P(\chi_k^2\ge t)$ denote the survival function of a
central chi-squared random variable with $k\ge 1$ degrees of freedom, and define
\[
    G_k(t):=-2\log \bar F_k(t),
    \qquad
    c_k:=(k-2)\log 2+2\log\Gamma(k/2).
\]
Fix $\eta\in(0,1)$. Uniformly over integers $k\ge 1$ and real numbers
$t\to\infty$ satisfying
\[
    \frac{k}{t}\le 1-\eta,
\]
we have
\begin{equation}
\label{eq:chi-tail-expansion}
\overline F_k(t)
=
\frac{t^{(k-2)/2}e^{-t/2}}
{2^{(k-2)/2}\Gamma(k/2)}
\left[
1+\frac{k-2}{t}
+O_\eta\left(\frac{(k+1)^2}{t^2}\right)
\right].
\end{equation}
Consequently,
\begin{equation}
\label{eq:Gk-expansion}
G_k(t)
=
t-(k-2)\log t+c_k+r_k(t),
\end{equation}
where
\begin{equation}
\label{eq:rk-def}
r_k(t)
=
-2\log
\left[
1+\frac{k-2}{t}
+O_\eta\left(\frac{(k+1)^2}{t^2}\right)
\right].
\end{equation}
In particular, if additionally $k/t=o(1)$, then
\begin{equation}
\label{eq:rk-small-k}
r_k(t)
=
-\frac{2(k-2)}{t}
+
O\left(\frac{(k+1)^2}{t^2}\right).
\end{equation}
Moreover, in the same uniform regime,
\begin{equation}
\label{eq:Gk-prime-expansion}
G_k'(t)
=
1-\frac{k-2}{t}
+
O_\eta\left(\frac{(k+1)^2}{t^2}\right),
\end{equation}
and
\begin{equation}
\label{eq:Gk-second-expansion}
G_k''(t)
=
\frac{k-2}{t^2}
+
O_\eta\left(\frac{(k+1)^2}{t^3}\right).
\end{equation}
Finally, as $k\to\infty$,
\begin{equation}
\label{eq:ck-stirling}
c_k
=
k\log k-k-\log k+O(1).
\end{equation}
\end{lemma}

\begin{proof}
Write
\[
    s=\frac{k}{2},\qquad x=\frac{t}{2}.
\]
Then
\[
    \overline F_k(t)
    =
    \frac{\Gamma(s,x)}{\Gamma(s)},
\]
where
\[
    \Gamma(s,x)=\int_x^\infty u^{s-1}e^{-u}\,du
\]
is the upper incomplete gamma function.

We first prove the tail expansion. Integrating by parts once gives
\[
\Gamma(s,x)
=
x^{s-1}e^{-x}
+
(s-1)\int_x^\infty u^{s-2}e^{-u}\,du .
\]
Integrating by parts a second time gives
\[
\int_x^\infty u^{s-2}e^{-u}\,du
=
x^{s-2}e^{-x}
+
(s-2)\int_x^\infty u^{s-3}e^{-u}\,du .
\]
Therefore
\[
\Gamma(s,x)
=
x^{s-1}e^{-x}
\left[
1+\frac{s-1}{x}
+
\frac{(s-1)(s-2)}{x^2}
\frac{\int_x^\infty u^{s-3}e^{-u}\,du}
     {x^{s-3}e^{-x}}
\right].
\]

We now bound the remainder. Since $k/t\le 1-\eta$, we have
\[
    \frac{s}{x}=\frac{k}{t}\le 1-\eta.
\]
For $u\ge x$,
\[
u^{s-3}e^{-u}
=
x^{s-3}e^{-x}
\left(\frac ux\right)^{s-3}e^{-(u-x)}.
\]
Using the change of variables $u=x+v$,
\[
\frac{\int_x^\infty u^{s-3}e^{-u}\,du}
     {x^{s-3}e^{-x}}
=
\int_0^\infty
\left(1+\frac vx\right)^{s-3}e^{-v}\,dv .
\]
For $v\ge0$ and $s/x\le 1-\eta$,
\[
\left(1+\frac vx\right)^{s-3}
\le
\exp\left(\frac{s v}{x}\right)
\le
\exp((1-\eta)v).
\]
Hence
\[
\frac{\int_x^\infty u^{s-3}e^{-u}\,du}
     {x^{s-3}e^{-x}}
\le
\int_0^\infty e^{-\eta v}\,dv
=
\frac1\eta .
\]
Substituting this remainder bound into the expansion of \(\Gamma(s,x)\) gives
\[
\Gamma(s,x)
=
x^{s-1}e^{-x}
\left[
1+\frac{s-1}{x}
+
O_\eta\left(\frac{(s+1)^2}{x^2}\right)
\right],
\]
uniformly in the stated regime. Dividing by $\Gamma(s)$ and substituting
$s=k/2$, $x=t/2$, we obtain
\[
\overline F_k(t)
=
\frac{(t/2)^{k/2-1}e^{-t/2}}{\Gamma(k/2)}
\left[
1+\frac{k-2}{t}
+
O_\eta\left(\frac{(k+1)^2}{t^2}\right)
\right].
\]
This proves~\eqref{eq:chi-tail-expansion}.

Taking logarithms of~\eqref{eq:chi-tail-expansion},
\[
\log\overline F_k(t)
=
\frac{k-2}{2}\log t
-\frac t2
-\frac{k-2}{2}\log 2
-\log\Gamma(k/2)
+
\log\left[
1+\frac{k-2}{t}
+
O_\eta\left(\frac{(k+1)^2}{t^2}\right)
\right].
\]
Multiplying by $-2$ yields
\[
G_k(t)
=
t-(k-2)\log t
+
(k-2)\log 2
+
2\log\Gamma(k/2)
-
2\log\left[
1+\frac{k-2}{t}
+
O_\eta\left(\frac{(k+1)^2}{t^2}\right)
\right].
\]
Since
\[
c_k=(k-2)\log 2+2\log\Gamma(k/2),
\]
this proves~\eqref{eq:Gk-expansion}--\eqref{eq:rk-def}. If $k/t=o(1)$, then the logarithm may be expanded as
$\log(1+u)=u+O(u^2)$, giving~\eqref{eq:rk-small-k}.

It remains to prove the derivative expansions. We use the hazard representation
rather than differentiating the remainder in~\eqref{eq:chi-tail-expansion}. The chi-squared density is
\[
f_k(t)
=
\frac{1}{2^{k/2}\Gamma(k/2)}
t^{k/2-1}e^{-t/2}.
\]
Since $\bar F_k'(t)=-f_k(t)$,
\[
G_k'(t)
=
-2\frac{\bar F_k'(t)}{\bar F_k(t)}
=
2\frac{f_k(t)}{\bar F_k(t)}.
\]
But~\eqref{eq:chi-tail-expansion} is equivalently
\[
\bar F_k(t)
=
2f_k(t)
\left[
1+\frac{k-2}{t}
+
O_\eta\left(\frac{(k+1)^2}{t^2}\right)
\right].
\]
Therefore
\[
G_k'(t)
=
\left[
1+\frac{k-2}{t}
+
O_\eta\left(\frac{(k+1)^2}{t^2}\right)
\right]^{-1}.
\]
Expanding the reciprocal gives
\[
G_k'(t)
=
1-\frac{k-2}{t}
+
O_\eta\left(\frac{(k+1)^2}{t^2}\right),
\]
which proves~\eqref{eq:Gk-prime-expansion}.

For the second derivative, set $a=(k-2)/2$ and use the exact integral
\[
 H_k(t):=\frac{\overline F_k(t)}{2f_k(t)}
 =\frac12\int_0^\infty(1+v/t)^a e^{-v/2}\,dv.
\]
In the stated regime this integral is bounded above and below by positive
constants depending only on $\eta$, and $H_k(t)=1+O_\eta((k+1)/t)$.
Differentiation under the integral is justified by an exponentially
decaying dominating function, because $a/t\le(1-\eta)/2$.
Consequently,
\[
 H_k'(t)=-\frac{a}{2t^2}
 \int_0^\infty v(1+v/t)^{a-1}e^{-v/2}\,dv
 =-\frac{k-2}{t^2}+O_\eta\!\left(\frac{(k+1)^2}{t^3}\right).
\]
For the remainder, apply the mean value theorem to
$(1+v/t)^{a-1}-1$ and integrate the bound
$|a-1|(v/t)\exp((a-1)_+v/t)$.
Since $G_k'=H_k^{-1}$, it follows that
\[
 G_k''(t)=-H_k'(t)H_k(t)^{-2}
 =\frac{k-2}{t^2}+O_\eta\!\left(\frac{(k+1)^2}{t^3}\right),
\]
proving~\eqref{eq:Gk-second-expansion}.

Finally, Stirling's formula gives
\[
\log\Gamma(z)
=
z\log z-z-\frac12\log z+O(1),
\qquad z\to\infty.
\]
Taking $z=k/2$,
\[
2\log\Gamma(k/2)
=
k\log(k/2)-k-\log(k/2)+O(1).
\]
Thus
\[
2\log\Gamma(k/2)
=
k\log k-k\log 2-k-\log k+O(1).
\]
Adding $(k-2)\log 2=k\log 2+O(1)$ yields
\[
c_k
=
k\log k-k-\log k+O(1),
\]
as claimed.
\end{proof}

\begin{lemma}[Gaussian moderate-deviation bound]
\label{lem:gaussian-md}
Let $Z \sim \mathcal{N}(0,1)$ and $x_p \to \infty$. Then
\begin{equation*}
-\log \Prob(Z \geq x_p) \;=\; \frac{x_p^2}{2} + \log(x_p\sqrt{2\pi}) + O(x_p^{-2}).
\end{equation*}
In particular, if $x_p^2/(2\log p) \to \ell \in [0, \infty)$, then $\Prob(Z \geq x_p) = p^{-\ell + o(1)}$.
\end{lemma}

The proof is simply Mills' ratio.

\section{ANOVA F-statistic and reduction to the chi-square numerator}
\label{app:anova-f-reduction}

We derive the exact distribution of the one-way ANOVA F-statistic under the Gaussian model and explain why, on the moderate-deviation scale used in Theorem~\ref{thm:md}, it suffices to analyze the noncentral chi-square
numerator.

Let
\[
    k:=C-1,
    \qquad
    m:=n-C.
\]
For feature \(j\), define
\[
    T_j
    :=
    \frac{\mathrm{SS}_{\mathrm{between}}(j)}{\sigma^2},
    \qquad
    U_j
    :=
    \frac{\mathrm{SS}_{\mathrm{within}}(j)}{\sigma^2}.
\]
Thus
\[
    f_j
    =
    \frac{T_j/k}{U_j/m}.
\]

\begin{lemma}[Exact ANOVA distribution]
\label{lem:anova-f-exact}
Under model~\eqref{eq:gaussian-model}, for each \(j\in[p]\),
\[
    T_j\sim \chi^2_k(\lambda_j),
    \qquad
    U_j\sim \chi^2_m,
    \qquad
    T_j\perp U_j,
\]
where
\[
    \lambda_j
    =
    \frac{n_c}{\sigma^2}
    \mu^{(j)\top}P\mu^{(j)},
    \qquad
    \mu^{(j)}
    :=
    (\mu_1^{(j)},\ldots,\mu_C^{(j)})^\top.
\]
Consequently,
\[
    f_j\sim F_{k,m}(\lambda_j).
\]
For \(j\notin\mathcal J\), \(\lambda_j=0\), so \(f_j\sim F_{k,m}\) and the
F-test p-value is uniform under \(H_0^{(j)}\). For \(j\in\mathcal J\), under
the spherical homogeneous calibration~\eqref{eq:lambda-calibration},
\[
    \lambda_j=2r\log p.
\]
\end{lemma}

\begin{proof}
The vector of class sample means at feature \(j\) satisfies
\[
    \hat{\mu}_{\cdot,j}
    =
    (\hat{\mu}_{1,j},\ldots,\hat{\mu}_{C,j})^\top
    \sim
    \mathcal N
    \left(
    \mu^{(j)},
    \frac{\sigma^2}{n_c}I_C
    \right).
\]
Since \(P\) is the orthogonal projection onto \(\mathbf 1^\perp\),
\[
    T_j
    =
    \frac{n_c}{\sigma^2}
    \hat{\mu}_{\cdot,j}^\top P\hat{\mu}_{\cdot,j}
\]
is a noncentral chi-square random variable with \(k=C-1\) degrees of freedom
and non-centrality
\[
    \lambda_j
    =
    \frac{n_c}{\sigma^2}
    \mu^{(j)\top}P\mu^{(j)}.
\]
The within-class residual sum of squares is independent of the class sample
means in the Gaussian one-way ANOVA model, and
\[
    U_j
    =
    \frac{\mathrm{SS}_{\mathrm{within}}(j)}{\sigma^2}
    \sim
    \chi^2_{n-C}.
\]
Therefore \(T_j\perp U_j\), and
\[
    f_j
    =
    \frac{T_j/k}{U_j/m}
    \sim
    F_{k,m}(\lambda_j).
\]
Finally, if \(j\in\mathcal J\), then by
\eqref{eq:lambda-calibration},
\[
    \lambda_j=2r\log p.
\]
If \(j\notin\mathcal J\), then \(P\mu^{(j)}=0\), so \(\lambda_j=0\).
\end{proof}

\begin{lemma}[Reduction from F p-values to the chi-square numerator]
\label{lem:F-to-chisq-reduction}
Let $\pi_j^F$ be the ANOVA F-test p-value at the feature $j$, as defined in \eqref{eq:F-p-value} and let
\[
    \pi_j^\chi
    :=
    \Prob\{\chi_k^2\ge T_j\}
\]
be the oracle chi-square numerator p-value. Suppose
\[
    \log p=o(m),
    \qquad
    m=n-C.
\]
Assume also $k/(2\log p)\to s\ge0$, $\lambda_p/(2\log p)\to r>0$,
and use the continuous exponent $\Psi_{r,s}$ derived directly in
Section~\ref{app:md-proof} of the Supplementary Material. If, for \(q\) in a compact subset of \((0,1)\),
\[
    \Prob\bigl(\pi_j^\chi\le p^{-q}\mid j\in\mathcal J\bigr)
    =
    p^{-\Psi_p(q)+o(1)}
\]
uniformly in \(q\), then
\[
    \Prob\bigl(\pi_j^F\le p^{-q}\mid j\in\mathcal J\bigr)
    =
    p^{-\Psi_p(q)+o(1)}
\]
uniformly in \(q\).
\end{lemma}

\begin{proof}
Write
\[
    f_j=\frac{T_j/k}{R_j/m},
    \qquad R_j\sim\chi^2_m,
    \qquad R_j\perp T_j.
\]
Put $L=\log p$ and $\eta_p=(L/m)^{1/4}$, so that $\eta_p\to0$.
The chi-square concentration inequality gives
\begin{equation}
\label{eq:denominator-concentration}
    a_p := \Prob\{|R_j/m-1|>\eta_p\}
    \le 2\exp\{-c\eta_p^2m\}
    =2\exp\{-c\sqrt{mL}\}=p^{-\omega(1)}.
\end{equation}
On the event $|R_j/m-1|\le\eta_p$,
\[
 \{T_j\ge 2Lx(1+\eta_p)\}
 \subseteq \{kf_j\ge2Lx\}
 \subseteq \{T_j\ge2Lx(1-\eta_p)\}.
\]
Adding the exceptional denominator event gives the corresponding probability
bounds.  The central and noncentral numerator rates derived directly in Section~\ref{app:md-proof} of the Supplementary Material are continuous in $x$, and the moving
thresholds converge to $x$, so both bounds have the same exponent uniformly
on compact $q$-intervals bounded away from zero.  The additive error $a_p$ is
negligible compared with every fixed polynomial tail.  Applying the same
bracketing to the central quantiles transfers the numerator quantile to the
central-$F$ quantile and proves the claimed uniform expansion.
\end{proof}

\section{Moderate-deviation exponent of ANOVA p-values}
\label{app:md-proof}
For $x\ge0$, define
\begin{equation}
\label{eq:J-variational}
 J_{r,s}(x)=\sup_{0\le\theta<1}
 \left\{\theta x+s\log(1-\theta)-\frac{r\theta}{1-\theta}\right\}.
\end{equation}
This upper-tail rate equals zero for $x\le s+r$.
For $x>s+r$, put $d=\sqrt{s^2+4rx}$ and $y=2x/(s+d)>1$; then
\begin{equation}
\label{eq:J-explicit}
 J_{r,s}(x)=x+r-d-s\log y.
\end{equation}
For $s=0$, this is $(\sqrt x-\sqrt r)_+^2$.
For $s>0$, let $g_s(x)=x-s-s\log(x/s)$ on $x\ge s$,
and let $x_s(q)$ be its increasing inverse; set $x_0(q)=q$.
Write $L=\log p$. The p-value exponent is $\Psi_{r,s}(q)=J_{r,s}(x_s(q))$.

\begin{theorem}[Uniform p-value tail exponent]
\label{thm:md}
Suppose $k/(2L)\to s\ge0$, $\lambda_p/(2L)\to r>0$, and $m/L\to\infty$,
with integer $k,m\ge1$. Let $T\sim\chi_k^2(\lambda_p)$ and
$R\sim\chi_m^2$ be independent. If $\pi$ is the central-$F_{k,m}$
upper-tail p-value of $(T/k)/(R/m)$, then, for every fixed $Q<\infty$,
\begin{equation}
\label{eq:md-exponent}
 \Prob(\pi\le p^{-q})=p^{-\Psi_p(q)+o(1)},
 \qquad \Psi_p(q):=\Psi_{r,s}(q),
 \quad\text{uniformly for }q\in[0,Q].
\end{equation}
The same conclusion holds for the numerator p-value
$\pi^\chi=\Prob(\chi_k^2\ge T)$.
\end{theorem}

For later use, define the boundary exponent (the HC growth exponent under null-variance normalization)
\begin{equation}
\label{eq:HC-exponent}
    E_p(q):=\frac12-\beta+\frac q2-\Psi_p(q).
\end{equation}
\subsection{Noncentral numerator tails}
Let $W_p=T/(2L)$. For $\theta<1$,
\[
 \Lambda_p(\theta):=L^{-1}\log\mathbb E e^{L\theta W_p}
 =-\frac{k}{2L}\log(1-\theta)
   +\frac{\lambda_p}{2L}\frac{\theta}{1-\theta}
 \longrightarrow
 \Lambda(\theta):=-s\log(1-\theta)+\frac{r\theta}{1-\theta}.
\]
The first two derivatives converge at each fixed $\theta<1$.
Chernoff's bound gives
\[
 \liminf_p -L^{-1}\log\Prob(W_p\ge x)\ge J_{r,s}(x).
\]
For the matching lower bound when $x>s+r$, choose $\varepsilon>0$ and
the unique $\theta_\varepsilon\in(0,1)$ satisfying
$\Lambda'(\theta_\varepsilon)=x+\varepsilon$; existence follows from
$\Lambda'(0)=s+r$ and $\Lambda'(\theta)\uparrow\infty$.
Tilt the law by
\[
 \frac{d\Prob_{\theta_\varepsilon}}{d\Prob}
 =\exp\{L\theta_\varepsilon W_p-L\Lambda_p(\theta_\varepsilon)\}.
\]
Under this law, the mean of $W_p$ tends to $x+\varepsilon$ and its
variance is $\Lambda_p''(\theta_\varepsilon)/L\to0$.
Thus $A_\varepsilon=\{x\le W_p\le x+2\varepsilon\}$ has tilted
probability tending to one, and
\[
 \Prob(W_p\ge x)\ge
 e^{-L\theta_\varepsilon(x+2\varepsilon)+L\Lambda_p(\theta_\varepsilon)}
 \Pr_{\theta_\varepsilon}(A_\varepsilon).
\]
Taking logarithms, then $p\to\infty$ and $\varepsilon \to 0$,
proves the matching exponent. Below $s+r$, $W_p\to s+r$ in probability
because its variance is $O(L^{-1})$, so the upper-tail exponent is zero.
At $x=s+r$, bound the tail below by its value at $x+\varepsilon$ and
let $\varepsilon\to 0$. This proves the rate at every $x\ge0$.

The optimizer above the mean satisfies
$x=s/(1-\theta)+r/(1-\theta)^2$.
Writing $y=(1-\theta)^{-1}$ gives $x=sy+ry^2$, whose positive root
is $2x/(s+\sqrt{s^2+4rx})$. Substitution proves
\eqref{eq:J-explicit}. These formulas also show continuity of the rate.

\subsection{Central numerator tails and quantiles}
For $T_0\sim\chi_k^2$ and $s>0$, the preceding argument with $r=0$
gives upper-tail rate zero for $x\le s$ and $g_s(x)$ for $x>s$.
When $s=0$, for any fixed $x>0$ and $0<\theta<1$,
Chernoff's bound gives
$\liminf -L^{-1}\log\Prob(T_0\ge2Lx)\ge\theta x$.
Letting $\theta\uparrow1$ gives the lower bound $x$ on the exponent.
Since $k\ge1$, $T_0$ stochastically dominates a squared standard normal,
whose tail at $2Lx$ is $p^{-x+o(1)}$. This gives the opposite bound.
Thus the rate is $x=g_0(x)$ in this case.

For every fixed $q>0$, the central chi-square upper quantile at
probability $p^{-q}$, divided by $2L$, tends to $x_s(q)$.
Indeed, evaluate the tail at $x_s(q)-\varepsilon$ and
$x_s(q)+\varepsilon$, choosing the lower point above $s$ when $s>0$;
strict monotonicity of $g_s$ brackets the quantile. Applying the
noncentral rate to these same brackets proves the numerator p-value
exponent for fixed $q>0$.

\subsection{Transfer to exact F p-values}
Let $D=R/m$, $\eta_p=(L/m)^{1/4}\to0$, and
$a_p=\Prob(|D-1|>\eta_p)$. By \eqref{eq:denominator-concentration},
\[
 a_p\le2e^{-c\sqrt{mL}}=p^{-\omega(1)}.
\]
For every fixed $x>0$, the variable $V=T/D=kF$ satisfies
\[
 \Prob(T\ge2Lx(1+\eta_p))-a_p
 \le\Prob(V\ge2Lx)
 \le\Prob(T\ge2Lx(1-\eta_p))+a_p.
\]
The moving numerator thresholds have the same rate as $x$, by
bracketing them between fixed $x\pm\varepsilon$ and using continuity.
The additive error is negligible relative to every fixed polynomial
tail. These statements hold for both central and noncentral numerators.
Consequently, if $c_p(q)$ is the central-$F_{k,m}$ upper quantile at
$p^{-q}$, then $kc_p(q)/(2L)\to x_s(q)$ for each fixed $q>0$.
Applying the noncentral bounds at this moving quantile yields
\eqref{eq:md-exponent} pointwise for $q>0$. At $q=0$, the event is
$\{\pi\le1\}$, with probability one, so its exponent is zero.
We do not claim convergence of the scaled quantile to $x_s(0)$ at
$q=0$ when $s>0$.

\subsection{Uniformity and interpretation}
The functions $f_p(q)=-L^{-1}\log\Prob(\pi\le p^{-q})$ are
nondecreasing, and their pointwise limit $\Psi_{r,s}$ is continuous
and nondecreasing. On $[0,Q]$, choose a finite partition so that
the oscillation of the limit on each subinterval is at most
$\varepsilon$. Pointwise convergence at its endpoints and monotonicity
bound $|f_p-\Psi_{r,s}|$ by $2\varepsilon$ everywhere for all large $p$.
This proves uniformity; the numerator argument is identical.
All useful features have the same non-centrality, so feature-wise
uniformity follows as well. This completes the proof of
Theorem~\ref{thm:md}.

In particular, for $s>0$,
\[
 \Psi_{r,s}(q)=0
 \quad\Longleftrightarrow\quad
 q\le g_s(s+r)=r-s\log(1+r/s).
\]
Local Gaussian approximations can describe fluctuations near this
location, but do not determine the full tail exponent at fixed $q$.
The tail-equivalence condition in \citet[Eq.~(2.5)]{kipnis2025unification}
must be verified for a proposed shifted-square approximation;
our direct calculation instead supplies the exact rate needed here.

\subsection{Local expansion at the useful-feature mean}
For $s>0$, define
\[
 \bar\mu=r-s\log(1+r/s),\qquad
 \bar\nu^2=\left(\frac r{r+s}\right)^2\frac{s+2r}{2\bar\mu}.
\]
These parameters describe a local expansion, not the full polynomial-tail rate.
Indeed, $g_s'(s+r)=r/(s+r)>0$, so as $q\to \bar\mu$ from above,
\[
 x_s(q)-(s+r)=\frac{s+r}{r}(q-\bar\mu)+O((q-\bar\mu)^2).
\]
The cumulant function satisfies $\Lambda''(0)=s+2r$. Consequently,
\[
 J_{r,s}(s+r+h)=\frac{h^2}{2(s+2r)}+O(h^3),\qquad h\to0,
\]
and therefore
\[
 \Psi_{r,s}(q)
 =\frac{(s+r)^2}{2r^2(s+2r)}(q-\bar\mu)^2+O((q-\bar\mu)^3)
 =\frac{(\sqrt q-\sqrt{\bar\mu})^2}{\bar\nu^2}
   +O((q-\bar\mu)^3).
\]
Below $\bar\mu$, both upper-tail expressions are zero when the positive part
is used. The displayed expansion holds locally as $q\to\bar\mu$;
it does not justify replacing $\Psi_{r,s}$ at other fixed thresholds.

\section{Proof of Theorem~\ref{thm:impossibility} (Impossibility below the boundary)}
\label{app:impossibility-proof}

We prove the impossibility result under the spherical homogeneous contrast
model. Let
\[
    \mathcal{V}=\mathbf 1^\perp
    =
    \left\{
    v\in\mathbb R^C:\mathbf 1^\top v=0
    \right\},
    \qquad
    P=I_C-\frac1C\mathbf 1\mathbf 1^\top.
\]
Each feature is randomly set as useful independently with probability
\[
    \epsilon_p := \epsilon = p^{-\beta}, \qquad\beta\in(1/2,1).
\]
If \(j\notin\mathcal J\), then \(\mu^{(j)}=0\). If \(j\in\mathcal J\), then
\[
    \mu^{(j)}=a_p U_j,
    \qquad
    U_j\sim
    \operatorname{Unif}
    \left\{
    u\in\mathcal{V}:\|u\|_2=1
    \right\},
\]
independently over \(j\), where
\[
    a_p
    =
    \frac{\sigma}{\sqrt{n_c}}\sqrt{2r\log p}.
\]
Thus every useful feature has ANOVA non-centrality
\[
    \lambda_j
    =
    \frac{n_c}{\sigma^2}\mu^{(j)\top}P\mu^{(j)}
    =
    2r\log p.
\]

For feature \(j\), define the normalized ANOVA statistic
\[
    T_j
    =
    \frac{n_c}{\sigma^2}
    \hat{\mu}_{\cdot,j}^\top P\hat{\mu}_{\cdot,j},
\]
and its null-calibrated p-value
\[
    \pi_j
    =
    \Prob\{\chi^2_{C-1}\ge T_j\}.
\]
Let
\[
    Q_j=-\log_p\pi_j.
\]
By Theorem~\ref{thm:md}, under a useful feature,
\[
    \Prob_{\mathrm{sig}}(Q_j\ge q)
    =
    p^{-\Psi_p(q)+o(1)}
\]
uniformly on compact \(q\)-intervals, where
\[
    \Psi_p(q):=\Psi_{r,s}(q).
\]
Here we have used the equivalence between \(Q_j\ge q\) and
\(-2\log\pi_j\ge 2q\log p\).

Recall the boundary exponent from~\eqref{eq:HC-exponent},
\[
    E_p(q)
    =
    \frac12-\beta+\frac q2-\Psi_p(q).
\]

For the exact exponent, the below-boundary assumption is equivalently
that the positive-exponent gap is strict.  Hence there exist constants \(\eta>0\) and \(a>0\) such that
\begin{equation}
\label{eq:strict-below-boundary}
    \sup_{0\le q\le 1+\eta}E_p(q)\le -a
\end{equation}
for all sufficiently large \(p\). We prove the result under the equivalent
strict-below-boundary formulation~\eqref{eq:strict-below-boundary}.

Let $P_0^{(p)}$ and $P_1^{(p)}$ denote the null and rare/weak
alternative laws of the training data $\mathcal D_n = \{(X_i, Y_i)\}_{i=1}^n$. The ancillary residual
components have the same law under both models, so it is enough to control the
likelihood ratio of the independent feature scores.  We first prove
\[
    \operatorname{TV}(P_1^{(p)},P_0^{(p)})\to0
\]
for this training experiment.  The additional test-observation argument is
given after the truncated-moment calculation.

\subsection{Reduction to the ANOVA p-value experiment}

For each feature \(j\), define
\[
    W_j
    =
    \frac{\sqrt{n_c}}{\sigma}P\hat{\mu}_{\cdot,j}
    \in \mathcal{V}.
\]
Under the null,
\[
    W_j\sim N_{\mathcal{V}}(0,I_{\mathcal{V}}).
\]
Under a useful feature,
\[
    W_j\mid U_j
    \sim
    N_{\mathcal{V}}(\theta_p U_j,I_{\mathcal{V}}),
    \qquad
    \theta_p=\sqrt{2r\log p},
\]
where
\[
    U_j\sim
    \operatorname{Unif}
    \left\{
    u\in\mathcal{V}:\|u\|_2=1
    \right\}.
\]
The within-class residuals and the grand mean across classes have the
same distribution under the null and alternative and are independent of
$W_j$: the mean contrast lies in $\mathcal{V}$, and the Gaussian projections onto
$\mathcal{V}$ and its orthogonal complement are independent. Thus the feature-wise
likelihood ratio depends only on \(W_j\).

Under the spherical prior, the useful-feature density is
\[
    f_1(w)
    =
    \int_{\{u\in\mathcal{V}:\|u\|_2=1\}}
    \phi_{\mathcal{V}}(w-\theta_pu)\,d\sigma(u),
\]
where \(d\sigma\) is the uniform probability measure on the unit sphere in
\(\mathcal{V}\). Since \(d\sigma\) is rotationally invariant, \(f_1(w)\) depends only
on \(\|w\|_2\). The null density
\[
    f_0(w)=\phi_{\mathcal{V}}(w)
\]
also depends only on \(\|w\|_2\). Hence \(f_1(w)/f_0(w)\) depends only on
\(\|w\|_2^2\). But
\[
    \|W_j\|_2^2
    =
    \frac{n_c}{\sigma^2}
    \hat{\mu}_{\cdot,j}^\top P\hat{\mu}_{\cdot,j}
    =
    T_j.
\]
Therefore the full likelihood ratio depends on feature \(j\) only through
\(T_j\), equivalently through \(\pi_j\), equivalently through
\(Q_j=-\log_p\pi_j\). Thus we may work with the independent scores
\[
    Q_1,\ldots,Q_p.
\]

\subsection{One-feature likelihood ratios}

Under the null, \(\pi_j\sim \operatorname{Unif}(0,1)\). Hence
\[
    \Prob_0(Q_j\ge q)=p^{-q},
\]
and the null density of \(Q_j\) is
\[
    h_0(q)=(\log p)p^{-q},
    \qquad q\ge0.
\]
Let \(h_{\mathrm{sig}}\) denote the density of \(Q_j\) conditional on
\(j\in\mathcal J\), and define
\[
    \ell_p(q)
    =
    \frac{h_{\mathrm{sig}}(q)}{h_0(q)}.
\]
This is the likelihood ratio of a useful feature versus a null feature
at score \(q\).

The rare/weak one-feature likelihood ratio is
\[
    L_j(q)
    =
    1-\epsilon_p+\epsilon_p\ell_p(q).
\]
Since features are independent,
\[
    L_p
    =
    \frac{dP_1^{(p)}}{dP_0^{(p)}}
    =
    \prod_{j=1}^p L_j(Q_j).
\]

\subsection{A likelihood-ratio upper bound}

We need a pointwise upper bound on \(\ell_p(q)\). Let
\(T\sim\chi^2_k\) under the null and \(T\sim\chi^2_k(\lambda_p)\) under a
useful feature, with \(k=C-1\) and \(\lambda_p=2r\log p\). Let \(f_0\)
and \(f_1\) be their densities, and define
\[
    R_p(t)=\frac{f_1(t)}{f_0(t)}.
\]
The noncentral chi-square density has the Poisson mixture representation
\[
    f_1(t)
    =
    e^{-\lambda_p/2}
    \sum_{m=0}^{\infty}
    \frac{(\lambda_p/2)^m}{m!}
    f_{k+2m}(t).
\]
Moreover,
\[
    \frac{f_{k+2m}(t)}{f_k(t)}
    =
    \frac{\Gamma(k/2)}{2^m\Gamma(k/2+m)}t^m.
\]
Therefore
\[
    R_p(t)
    =
    e^{-\lambda_p/2}
    \sum_{m=0}^{\infty}a_m t^m,
\]
where all coefficients satisfy \(a_m\ge0\). Hence \(R_p(t)\) is
nondecreasing in \(t\).

The map \(t\mapsto Q=-\log_p S_0(t)\), where
\(S_0(t)=\Prob(\chi^2_k\ge t)\), is increasing. Therefore
\(q\mapsto \ell_p(q)\) is nondecreasing. Hence, for any \(q\),
\[
\begin{aligned}
    \Prob_{\mathrm{sig}}(Q\ge q)
    &=
    \int_q^\infty \ell_p(u)h_0(u)\,du  \\
    &\ge
    \ell_p(q)\int_q^\infty h_0(u)\,du  \\
    &=
    \ell_p(q)\Prob_0(Q\ge q).
\end{aligned}
\]
Thus
\[
    \ell_p(q)
    \le
    \frac{\Prob_{\mathrm{sig}}(Q\ge q)}
    {\Prob_0(Q\ge q)}.
\]
Using Theorem~\ref{thm:md} and the exact null tail,
\[
    \Prob_{\mathrm{sig}}(Q\ge q)=p^{-\Psi_p(q)+o(1)},
    \qquad
    \Prob_0(Q\ge q)=p^{-q},
\]
we obtain
\begin{equation}
\label{eq:lr-upper-impossibility}
    \ell_p(q)
    \le
    p^{q-\Psi_p(q)+o(1)}
\end{equation}
uniformly for \(q\in[0,1+\eta]\).

\subsection{Truncation}

Define
\[
    A_p
    =
    \left\{
    \max_{1\le j\le p}Q_j\le 1+\eta
    \right\}.
\]
We first show that \(A_p\) is typical under both \(P_0^{(p)}\) and
\(P_1^{(p)}\).

Under the null,
\[
    \Prob_0(Q_j>1+\eta)=p^{-(1+\eta)}.
\]
By the union bound,
\[
    \Prob_0(A_p^c)
    \le
    p\cdot p^{-(1+\eta)}
    =
    p^{-\eta}
    \to0.
\]
Under the alternative,
\[
\begin{aligned}
    \Prob_1(Q_j>1+\eta)
    &=
    (1-\epsilon_p)\Prob_0(Q_j>1+\eta)
    +
    \epsilon_p\Prob_{\mathrm{sig}}(Q_j>1+\eta)  \\
    &\le
    p^{-(1+\eta)}
    +
    p^{-\beta-\Psi_p(1+\eta)+o(1)}.
\end{aligned}
\]
Therefore,
\[
    \Prob_1(A_p^c)
    \le
    p^{-\eta}
    +
    p^{1-\beta-\Psi_p(1+\eta)+o(1)}.
\]
Since
\[
    E_p(1+\eta)
    =
    1-\beta+\frac{\eta}{2}-\Psi_p(1+\eta),
\]
and by~\eqref{eq:strict-below-boundary}
\(E_p(1+\eta)\le -a\), we have
\[
    1-\beta-\Psi_p(1+\eta)
    \le
    -a-\frac{\eta}{2}
    <0.
\]
Thus
\[
    \Prob_1(A_p^c)\to0.
\]
Equivalently, since \(L_p=dP_1^{(p)}/dP_0^{(p)}\),
\begin{equation}
\label{eq:first-truncated-moment-impossibility}
    \mathbb E_0[L_p\mathbf 1_{A_p}]
    =
    \Prob_1(A_p)
    \to1.
\end{equation}

\subsection{Truncated second moment}

Because
\[
    L_p=\prod_{j=1}^p L_j(Q_j),
    \qquad
    A_p=\bigcap_{j=1}^p\{Q_j\le 1+\eta\},
\]
we have, under \(P_0^{(p)}\),
\[
    \mathbb E_0[L_p^2\mathbf 1_{A_p}]
    =
    \left(
    \mathbb E_0[
    L_1(Q)^2\mathbf 1_{\{Q\le1+\eta\}}
    ]
    \right)^p.
\]
Now
\[
    L_1(q)=1-\epsilon_p+\epsilon_p\ell_p(q),
\]
so
\[
    L_1(q)^2
    =
    (1-\epsilon_p)^2
    +
    2\epsilon_p(1-\epsilon_p)\ell_p(q)
    +
    \epsilon_p^2\ell_p(q)^2.
\]
Using
\[
    \Prob_0(Q\le1+\eta)\le1
\]
and
\[
    \mathbb E_0[\ell_p(Q)\mathbf 1_{\{Q\le1+\eta\}}]
    \le
    \mathbb E_0[\ell_p(Q)]
    =
    1,
\]
we get
\[
    \mathbb E_0[
    L_1(Q)^2\mathbf 1_{\{Q\le1+\eta\}}
    ]
    \le
    1+\epsilon_p^2
    \mathbb E_0[
    \ell_p(Q)^2\mathbf 1_{\{Q\le1+\eta\}}
    ].
\]
Moreover,
\[
\begin{aligned}
    \mathbb E_0[
    \ell_p(Q)^2\mathbf 1_{\{Q\le1+\eta\}}
    ]
    &=
    \int_0^{1+\eta}\ell_p(q)^2h_0(q)\,dq  \\
    &\le
    \int_0^{1+\eta}
    p^{2q-2\Psi_p(q)+o(1)}
    (\log p)p^{-q}\,dq  \\
    &=
    (\log p)
    \int_0^{1+\eta}
    p^{q-2\Psi_p(q)+o(1)}\,dq.
\end{aligned}
\]
The \(o(1)\) term is uniform on the compact interval \([0,1+\eta]\), so
\[
    \mathbb E_0[
    \ell_p(Q)^2\mathbf 1_{\{Q\le1+\eta\}}
    ]
    \le
    p^{
    \sup_{0\le q\le1+\eta}
    \{q-2\Psi_p(q)\}
    +o(1)
    }.
\]
Since \(\epsilon_p=p^{-\beta}\),
\[
\begin{aligned}
    p\epsilon_p^2
    \mathbb E_0[
    \ell_p(Q)^2\mathbf 1_{\{Q\le1+\eta\}}
    ]
    &\le
    p^{
    1-2\beta+
    \sup_{0\le q\le1+\eta}
    \{q-2\Psi_p(q)\}
    +o(1)
    }  \\
    &=
    p^{
    2\sup_{0\le q\le1+\eta}E_p(q)
    +o(1)
    }  \\
    &\le
    p^{-2a+o(1)}
    \to0.
\end{aligned}
\]
Therefore,
\[
    \epsilon_p^2
    \mathbb E_0[
    \ell_p(Q)^2\mathbf 1_{\{Q\le1+\eta\}}
    ]
    =
    o(p^{-1}),
\]
and hence
\[
    \mathbb E_0[
    L_1(Q)^2\mathbf 1_{\{Q\le1+\eta\}}
    ]
    \le
    1+o(p^{-1}).
\]
Consequently,
\[
    \mathbb E_0[L_p^2\mathbf 1_{A_p}]
    \le
    (1+o(p^{-1}))^p
    =
    1+o(1).
\]
For the matching lower bound, Cauchy--Schwarz gives
\[
    \mathbb E_0[L_p^2\mathbf 1_{A_p}]
    \ge
    \left(
    \mathbb E_0[L_p\mathbf 1_{A_p}]
    \right)^2.
\]
By~\eqref{eq:first-truncated-moment-impossibility}, the right-hand side
tends to \(1\). Thus
\begin{equation}
\label{eq:second-truncated-moment-impossibility}
    \mathbb E_0[L_p^2\mathbf 1_{A_p}]
    \to1.
\end{equation}

\subsection{Conclusion: total variation and classification}

We now show that the truncated moment bounds imply total variation
convergence. For any measurable event \(B\),
\[
    P_1^{(p)}(B)-P_0^{(p)}(B)
    =
    \mathbb E_0[L_p\mathbf 1_B]-P_0^{(p)}(B).
\]
Decompose
\[
    \mathbb E_0[L_p\mathbf 1_B]
    =
    \mathbb E_0[L_p\mathbf 1_{A_p}\mathbf 1_B]
    +
    \mathbb E_0[L_p\mathbf 1_{A_p^c}\mathbf 1_B].
\]
Then
\[
\begin{aligned}
    |P_1^{(p)}(B)-P_0^{(p)}(B)|
    &\le
    \left|
    \mathbb E_0[(L_p\mathbf 1_{A_p}-1)\mathbf 1_B]
    \right|  \\
    &\quad+
    \mathbb E_0[L_p\mathbf 1_{A_p^c}]
    +
    P_0^{(p)}(A_p^c).
\end{aligned}
\]
The second term equals
\[
    \mathbb E_0[L_p\mathbf 1_{A_p^c}]
    =
    P_1^{(p)}(A_p^c)
    \to0,
\]
and the third term also tends to zero. For the first term,
Cauchy--Schwarz gives
\[
    \left|
    \mathbb E_0[(L_p\mathbf 1_{A_p}-1)\mathbf 1_B]
    \right|
    \le
    \left(
    \mathbb E_0[(L_p\mathbf 1_{A_p}-1)^2]
    \right)^{1/2}.
\]
But
\[
\begin{aligned}
    \mathbb E_0[(L_p\mathbf 1_{A_p}-1)^2]
    &=
    \mathbb E_0[L_p^2\mathbf 1_{A_p}]
    -
    2\mathbb E_0[L_p\mathbf 1_{A_p}]
    +
    1  \\
    &\to
    1-2+1
    =
    0,
\end{aligned}
\]
by~\eqref{eq:first-truncated-moment-impossibility} and
\eqref{eq:second-truncated-moment-impossibility}. Hence
\[
    \sup_B |P_1^{(p)}(B)-P_0^{(p)}(B)|
    \to0.
\]
Therefore
\[
    \operatorname{TV}(P_1^{(p)},P_0^{(p)})\to0.
\]

The training-score calculation above proves indistinguishability for the
training experiment.  To include the test observation, augment the experiment
by one independent observation $Z_c$ from each class, conditional on the same
latent means.  The augmented design has class size $n_c+1$ and non-centrality
\[
    \lambda_p^+=2r\log p\,(1+1/n_c).
\]
Because $n_c\to\infty$, its normalized non-centrality has the same limit $r$;
the preceding truncated-moment proof therefore gives total variation $o(1)$
for the augmented experiment as well.  Now draw $Y$ uniformly from $[C]$,
independently of the augmented observations, and evaluate
$\hat h(\mathcal D_n,Z_Y)$.  Under the alternative this has exactly the law
of the original test pair.  Under the null, $Y$ is independent of the
observed data, so the correctness probability is $1/C$.  Applying total
variation to the correctness event gives, uniformly over classifiers,
\[
    \Prob_1\{\hat h(\mathcal D_n,Z_Y)=Y\}
    \le \frac1C+o(1).
\]
A constant classifier attains $1/C$, completing the proof of
Theorem~\ref{thm:impossibility}.

% ================================================

\section{HC growth and HCT adaptivity}
\label{app:detection-proof}

We analyze the HCT rule of Algorithm~\ref{alg:dphct}.
For a threshold $t\in[0,1]$, write
\[
 \mathrm{TP}(t)=\sum_{j\in\mathcal J}\mathbf1\{\pi_j\le t\},\qquad
 \mathrm{FP}(t)=\sum_{j\notin\mathcal J}\mathbf1\{\pi_j\le t\},\qquad
 N(t)=\mathrm{TP}(t)+\mathrm{FP}(t).
\]
On the admissible set $\mathcal A_p=\{t:1\le N(t)\le\lfloor\gamma_0p\rfloor\}$,
define
\begin{equation}
\label{eq:HC-rank-threshold}
 H_p(t)=\frac{N(t)-pt}{\sqrt{N(t)(1-N(t)/p)}}.
\end{equation}
For pointwise probability statements set $H_p(t)=0$ outside $\mathcal A_p$.
The exact F p-values are continuous, so ties have probability zero and
$H_p(\pi_{(j)})=\HC_{p,j}$. Between consecutive p-values the count and
denominator are constant, while the numerator decreases with $t$.
Thus the algorithm maximizes~\eqref{eq:HC-rank-threshold} over $\mathcal A_p$.
Write $\hat t=\pi_{(j^\star)}$ for its chosen threshold, with deterministic
tie-breaking if several scores attain the maximum.

For $q\in(0,1)$ define
\[
 A_p(q)=1-\beta-\Psi_p(q),\qquad B(q)=1-q,
 \qquad E_p(q)=A_p(q)-B(q)/2,
\]
and the HC growth exponent for the empirical denominator in~\eqref{eq:HC-rank-threshold}
\begin{equation}
\label{eq:rank-HC-exponent}
 \widetilde E_p(q)=A_p(q)-\tfrac12\max\{A_p(q),B(q)\}
 =\min\{E_p(q),A_p(q)/2\}.
\end{equation}
Here $A_p(q)$ and $B(q)$ are the exponents of the expected useful and
null selection counts. The denominator in~\eqref{eq:HC-rank-threshold}
uses the observed fraction $N(t)/p$, rather than the null probability $t$.
Where $E_p(q)>0$, its exponent is
$\tfrac12\max\{A_p(q),B(q)\}$; subtracting this from the numerator
exponent $A_p(q)$ gives $\widetilde E_p(q)$.
Since $B(q)>0$, $\widetilde E_p(q)>0$ if and only if $E_p(q)>0$.
Thus HC grows as a positive power of $p$ for the same parameter values
that define the boundary in~\eqref{eq:xi-boundary}, although its growth
rate and maximizing threshold can differ from those for the null denominator.
The algorithm scans the ordered observed p-values; the deterministic grids
below are used only to prove uniform bounds.

\begin{theorem}[HC growth and HCT adaptivity]
\label{thm:detection}
Assume the spherical homogeneous model, feature-wise independence,
the calibrations of Theorem~\ref{thm:md}, $\beta\in(1/2,1)$, and fixed
$\gamma_0\in(0,1)$. Then:
\begin{enumerate}[leftmargin=*,label=\textup{(\alph*)}]
\item For every fixed $q\in(0,1)$ with $E_p(q)>0$,
\begin{equation}
\label{eq:HC-growth}
 H_p(p^{-q})=p^{\widetilde E_p(q)+o_\Prob(1)}.
\end{equation}
For arbitrary fixed $q\in(0,1)$, its positive part satisfies
\begin{equation}
\label{eq:HC-growth-upper}
 [H_p(p^{-q})]_+=O_\Prob\bigl(p^{\max\{\widetilde E_p(q),0\}+o(1)}\bigr).
\end{equation}
At a nonadmissible threshold set this positive part to zero; any fixed
$q\in(0,1)$ is admissible with probability tending to one.
\item If $\sup_{0<q<1}E_p(q)>0$, there is a fixed $d>0$ such that
\begin{equation}
\label{eq:rank-count-certificate}
 \Prob\left\{\frac{\mathrm{TP}(\hat t)^2}{N(\hat t)}\ge p^d\right\}
 \longrightarrow1.
\end{equation}
In particular, after decreasing $d$ if necessary,
$\mathrm{TP}(\hat t)\ge p^d$ and
$\mathrm{TP}(\hat t)^2/(\mathrm{FP}(\hat t)+1)\ge p^d$
with probability tending to one.
\end{enumerate}
\end{theorem}

\subsection{Count lemmas}
\label{app:count-lemmas}
In the following two lemmas, $\mathcal Q_p\subset(0,1)$ is any deterministic grid of polynomial cardinality. It is used for analysis, not to define HCT.
Write $\mathrm{TP}_q=\mathrm{TP}(p^{-q})$ and
$\mathrm{FP}_q=\mathrm{FP}(p^{-q})$ for exponent thresholds.

\begin{lemma}[False positives]
\label{lem:FP}
Assume $\{\pi_j:j\in\mathcal J^c\}$ are i.i.d. $\mathrm{Uniform}[0,1]$
and the Bernoulli signal-set model of Section~\ref{sec:model}. Then, for every compact $Q\Subset(0,1)$ and every polynomial grid $\mathcal Q_p\subset Q$ with $|\mathcal Q_p|\le p^a$,
\[
    \mathrm{FP}_q=p^{1-q}(1+o(1))
\]
uniformly over $q\in\mathcal Q_p$, with probability $1-o(1)$. Moreover,
\[
    \mathrm{FP}_q-p^{1-q}=O_\Prob(p^{(1-q)/2})
\]
for each fixed $q\in Q$.
\end{lemma}

\begin{proof}
Unconditionally, $\mathrm{FP}_q$ is binomial with parameters
$p$ and $(1-\epsilon_p)p^{-q}$. Its mean differs from $p^{1-q}$ by
$p^{1-\beta-q}$, which is $o(p^{(1-q)/2})$ since $\beta>1/2$.
Bernstein's inequality gives relative concentration uniformly on the stated
grid because its smallest mean is polynomial in $p$. The variance bound
and the negligible mean difference give the fixed-$q$ statement.
\end{proof}

\begin{lemma}[True positives]
\label{lem:TP}
Assume the conditions of Theorem~\ref{thm:md}, feature-wise independence, and condition on a signal set of size $|\mathcal J|=p^{1-\beta}(1+o(1))$, an event of probability $1-o(1)$. Then, uniformly over the deterministic grid,
\[
    \E[\mathrm{TP}_q]=p^{1-\beta-\Psi_p(q)+o(1)}.
\]
Furthermore, whenever $1-\beta-\Psi_p(q)>0$ is bounded away from zero on a subset of the grid, the corresponding counts concentrate relatively uniformly:
\[
    \mathrm{TP}_q=p^{1-\beta-\Psi_p(q)+o_\Prob(1)}.
\]
For all grid points, the upper bound
\[
    \mathrm{TP}_q=O_\Prob\bigl(p^{\max\{1-\beta-\Psi_p(q),0\}+o(1)}\bigr)
\]
holds uniformly over $q\in\mathcal Q_p$.
\end{lemma}

\begin{proof}
Condition on the signal set $\mathcal J$. For $q\in \mathcal Q_p$, write
\[
    \mathrm{TP}_q
    =
    \sum_{j\in\mathcal J} X_j(q),
    \qquad
    X_j(q):=\mathbf 1\{\pi_j\le p^{-q}\}.
\]
By feature-wise independence, for each fixed $q$ the variables
$\{X_j(q):j\in\mathcal J\}$ are independent Bernoulli variables. By
Theorem~\ref{thm:md},
\[
    \Prob(X_j(q)=1)
    =
    \Prob(\pi_j\le p^{-q}\mid j\in\mathcal J)
    =
    p^{-\Psi_p(q)+o(1)}
\]
uniformly over $q\in \mathcal Q_p$. Hence, on the event $|\mathcal J|=p^{1-\beta}(1+o(1))$,
\[
    m_p(q):=\E[\mathrm{TP}_q\mid \mathcal J]
    =
    |\mathcal J|p^{-\Psi_p(q)+o(1)}
    =
    p^{1-\beta-\Psi_p(q)+o(1)}
\]
uniformly over $q\in\mathcal Q_p$. This proves the expectation claim.

We next prove the relative concentration statement. Let
$\mathcal A_p\subseteq \mathcal Q_p$ be a subset on which
\[
    1-\beta-\Psi_p(q)\ge \eta
\]
for some fixed $\eta>0$ and all large $p$. Then, uniformly over
$q\in\mathcal A_p$,
\[
    m_p(q)\ge p^{\eta/2}
\]
for all sufficiently large $p$. Bernstein's inequality for sums of
independent Bernoulli variables gives, for any fixed $\varepsilon>0$,
\[
    \Prob\left(
        |\mathrm{TP}_q-m_p(q)|\ge \varepsilon m_p(q)
        \,\middle|\, \mathcal J
    \right)
    \le
    2\exp\{-c_\varepsilon m_p(q)\}
\]
for a constant $c_\varepsilon>0$. Therefore
\[
    \Prob\left(
        \exists q\in\mathcal A_p:
        |\mathrm{TP}_q-m_p(q)|\ge \varepsilon m_p(q)
        \,\middle|\, \mathcal J
    \right)
    \le
    2|\mathcal Q_p|\exp\{-c_\varepsilon p^{\eta/2}\}.
\]
Since $|\mathcal Q_p|\le p^a$, the right-hand side is $o(1)$. Hence
\[
    \sup_{q\in\mathcal A_p}
    \left|
        \frac{\mathrm{TP}_q}{m_p(q)}-1
    \right|
    =
    o_\Prob(1).
\]
Combining this with
$m_p(q)=p^{1-\beta-\Psi_p(q)+o(1)}$ yields
\[
    \mathrm{TP}_q
    =
    p^{1-\beta-\Psi_p(q)+o_\Prob(1)}
\]
uniformly over $q\in\mathcal A_p$. This proves the relative
concentration statement.

It remains to prove the uniform upper bound over all grid points. Again by
Bernstein's inequality, for every $t>0$,
\[
    \Prob\left(
        \mathrm{TP}_q-m_p(q)\ge t
        \,\middle|\, \mathcal J
    \right)
    \le
    \exp\left(
        -\frac{t^2}{2m_p(q)+2t/3}
    \right).
\]
Take
\[
    t=K\{\sqrt{m_p(q)\log p}+\log p\}.
\]
For $K$ sufficiently large,
\[
    \Prob\left(
        \mathrm{TP}_q
        \ge
        m_p(q)+K\{\sqrt{m_p(q)\log p}+\log p\}
        \,\middle|\, \mathcal J
    \right)
    \le
    p^{-a-2}.
\]
A union bound over $|\mathcal Q_p|\le p^a$ therefore gives, with
probability $1-o(1)$,
\[
    \mathrm{TP}_q
    \le
    m_p(q)+K\sqrt{m_p(q)\log p}+K\log p
\]
simultaneously for all $q\in\mathcal Q_p$.

Since
\[
    m_p(q)=p^{1-\beta-\Psi_p(q)+o(1)},
\]
the preceding display implies
\[
    \mathrm{TP}_q
    =
    O_\Prob\left(
        p^{\max\{1-\beta-\Psi_p(q),0\}+o(1)}
    \right)
\]
uniformly over $q\in\mathcal Q_p$. Indeed, when
$1-\beta-\Psi_p(q)>0$, the mean term dominates on the polynomial scale,
while when $1-\beta-\Psi_p(q)\le 0$, all three terms are at most
$p^{o(1)}$. This proves the lemma.
\end{proof}

\begin{lemma}[Auxiliary grid and uniform counts]
\label{lem:uniform-threshold-counts}
Conditional on $\mathcal J$, put $m_J=|\mathcal J|$,
$a_p(t)=\Prob(\pi_j\le t\mid j\in\mathcal J)$ and $m_T(t)=m_Ja_p(t)$.
There is a deterministic grid $\mathcal T_p\subset[0,1]$ of cardinality
$O(p^4)$ such that, with probability $1-o(1)$, each interval between
successive grid points contains at most one observed p-value.
Furthermore, with probability $1-o(1)$, simultaneously for every $t\in[0,1]$,
\begin{align}
\label{eq:fp-additive-full}
 |\mathrm{FP}(t)-(p-m_J)t|
 &\le K\{\sqrt{pt\log p}+\log p\},\\
\label{eq:tp-additive-full}
 |\mathrm{TP}(t)-m_T(t)|
 &\le K\{\sqrt{m_T(t)\log p}+\log p\}.
\end{align}
For each fixed $d>0$, these bounds imply
$m_T(t)/\mathrm{TP}(t)=1+o(1)$ uniformly over thresholds with
$\mathrm{TP}(t)\ge p^d$.
\end{lemma}

\begin{proof}
The useful-feature F p-value distribution $a_p$ is continuous. The function
$G_p(t)=(t+a_p(t))/2$ is therefore continuous and strictly increasing
from zero to one. Set $M_p=\lceil p^4\rceil$ and
\[
 u_\ell=G_p^{-1}(\ell/M_p),\quad 0\le\ell\le M_p,
 \qquad \mathcal T_p=\{u_0,\ldots,u_{M_p}\}.
\]
Each cell $(u_{\ell-1},u_\ell]$ has probability at most $2/M_p$ under
either a null or a useful p-value distribution. Conditional on $\mathcal J$,
two different features are independent, so the probability that their
p-values share a cell is at most $2/M_p$. A union bound gives
\[
 \Prob(\text{some cell contains at least two p-values}\mid\mathcal J)
 \le p(p-1)/M_p=o(1).
\]
At each deterministic grid point, apply binomial Bernstein bounds to
$\mathrm{FP}$ and $\mathrm{TP}$ with deviation parameter $K_0\log p$.
Taking $K_0$ sufficiently large makes a union bound over $O(p^4)$ points
valid. This proves the two displayed bounds on $\mathcal T_p$.
For an arbitrary threshold, move to its lower adjacent grid point.
On the no-collision event each count changes by at most one; its mean
changes by at most $2p/M_p$. Both changes are absorbed by the
$K\log p$ terms, proving uniformity on $[0,1]$.
Finally~\eqref{eq:tp-additive-full} implies
$\mathrm{TP}(t)\le2m_T(t)+K'\log p$. If $\mathrm{TP}(t)\ge p^d$,
then $m_T(t)$ is polynomially large, and dividing the same bound by
$m_T(t)$ proves relative concentration uniformly in this region.
The grid depends on model distributions but is used only in the proof;
the algorithm does not construct it.
\end{proof}

\subsection{Pointwise HC growth}
\label{app:growth-proof}

Fix $q\in(0,1)$, put $t=p^{-q}$, and condition on the high-probability
event $m_J=p^{1-\beta}(1+o(1))$. The null count has mean
$p^{B(q)}(1+o(1))$ and fluctuation $O_\Prob(p^{B(q)/2})$.
If $E_p(q)>0$, then $A_p(q)>B(q)/2>0$, so the useful count concentrates:
\[
 \mathrm{TP}(t)=p^{A_p(q)+o_\Prob(1)}.
\]
The identity
\[
 N(t)-pt=\mathrm{TP}(t)
       +\{\mathrm{FP}(t)-(p-m_J)t\}-m_Jt
\]
shows that the numerator has exponent $A_p(q)$. Indeed, the null
fluctuation has smaller exponent, and $\Psi_p(q)<q$ makes $m_Jt$
negligible. Moreover,
\[
 N(t)=p^{\max\{A_p(q),B(q)\}+o_\Prob(1)},\qquad N(t)/p\to0.
\]
Substituting in~\eqref{eq:HC-rank-threshold} proves~\eqref{eq:HC-growth}.
If $E_p(q)\le0$, then $A_p(q)\le B(q)/2<B(q)$; the null count
dominates $N(t)$, while the positive numerator is at most
$p^{B(q)/2+o_\Prob(1)}$ by the count bounds. The positive part of HC
is thus at most subpolynomial, proving~\eqref{eq:HC-growth-upper}.
Since $B(q)>0$ and $A_p(q)\le1-\beta<1$, each such fixed threshold
is admissible with probability tending to one.

\subsection{Adaptivity of the HCT threshold}
\label{app:adaptivity-proof}

Choose a fixed $q_0\in(0,1)$ with $E_p(q_0)>0$, which exists above
the boundary by continuity, including when the maximum over $[0,1]$
is achieved at an endpoint. Part (a) gives some $a>0$ such that
$H_p(p^{-q_0})\ge p^a$ with probability $1-o(1)$.
Let $j_0=N(p^{-q_0})$. With high probability this is an admissible rank,
and $\pi_{(j_0)}\le p^{-q_0}$. The denominator depends only on the count,
so
\[
 H_p(\hat t)\ge H_p(\pi_{(j_0)})\ge H_p(p^{-q_0})\ge p^a.
\]
On the uniform event in Lemma~\ref{lem:uniform-threshold-counts}, consider
any admissible $t$ with $H_p(t)>0$. Then $pt<N(t)$, and dropping the
nonpositive missing-null term $-m_Jt$ gives
\begin{align*}
 H_p(t)
 &\le\frac{\mathrm{TP}(t)+K\{\sqrt{pt\log p}+\log p\}}
 {\sqrt{N(t)(1-N(t)/p)}}\\
 &\le\frac{\mathrm{TP}(t)}{\sqrt{(1-\gamma_0)N(t)}}
       +K_{\gamma_0}\log p.
\end{align*}
Applying this inequality at $\hat t$ yields
$\mathrm{TP}(\hat t)/\sqrt{N(\hat t)}\ge c p^a$ with high probability.
Squaring and reducing the exponent proves~\eqref{eq:rank-count-certificate}.
Finally $\mathrm{TP}\le N$ and $\mathrm{FP}+1\le2N$ imply the stated
count consequences. This argument covers the entire rank scan, including
thresholds below $1/p$, without asserting that the selected exponent
belongs to $(0,1)$. This proves Theorem~\ref{thm:detection}. \qed

\section{Boundary optimization}
\label{app:pivot}
\begin{proposition}[Explicit maximizer and boundary properties]
\label{cor:pivot}
Write $\Delta(\beta,r,s)=\max_{0\le q\le1}E_p(q)$. For $s\ge0$, let $g_0(x)=x$, $x_0(1)=1$, and define
\[
 x_1=x_s(1),\qquad
 x_\dagger=2r+\sqrt{s^2+4r^2},\qquad
 x_*=\min\{x_1,x_\dagger\}.
\]
Then
\begin{equation}
\label{eq:explicit-optimizer}
 \Delta(\beta,r,s)=\frac12-\beta+\frac12g_s(x_*)-J_{r,s}(x_*).
\end{equation}
For every $\beta\in(1/2,1)$ and fixed $s\ge0$, $\Delta$ crosses
zero at a unique positive $r=\xi^\star(\beta,s)$, with strict signs
on either side. The boundary is strictly increasing in $s$ and
\[
 \xi^\star(\beta,0)=\rho^\star(\beta),\quad
 \lim_{s\to0}\xi^\star(\beta,s)=\rho^\star(\beta),\quad
 \xi^\star(\beta,s)\sim\sqrt{2s\rho^\star(\beta)}
 \quad(s\to\infty).
\]
\end{proposition}
\begin{proof}
Substitute $q=g_s(x)$ so that $s\le x\le x_1$.
Below $s+r$, $J=0$ and $g_s$ is increasing. Above $s+r$,
the envelope theorem in \eqref{eq:J-variational}, or direct
differentiation of~\eqref{eq:J-explicit}, gives
\[
 \frac{d}{dx}\left(\tfrac12g_s(x)-J_{r,s}(x)\right)
 =\frac{\sqrt{s^2+4rx}-x}{2x}.
\]
This derivative changes sign once, at $x=x_\dagger>s+r$.
The same formula holds for $s=0$ above $x=r$, where $x_\dagger=4r$.
This proves \eqref{eq:explicit-optimizer}.

For fixed $s$, as $r\to0$ the useful rate tends to the null
rate and $\Delta\to1/2-\beta<0$. Once $s+r\ge x_1$,
$\Delta=1-\beta>0$. Continuity follows from the explicit formulas.
At any $r$ where $\Delta\le0$, the optimizer lies strictly above
$s+r$ (otherwise $x_1\le s+r$ and $\Delta=1-\beta$).
In that region, at fixed $x$, $\partial_rJ=-(y-1)<0$.
Increasing $r$ thus strictly increases the optimized value until it
becomes positive; if the same $x$ enters the zero-rate region, its
rate has still strictly decreased. Monotonicity thereafter prevents
a second crossing. This proves the existence, uniqueness and signs.

At $s=0$, if $r\le1/4$, the maximum is
$\Delta=1/2-\beta+r$. If $1/4<r<1$, it is
$\Delta=1-\beta-(1-\sqrt r)^2$; for $r\ge1$ it is $1-\beta$.
Solving for its zero gives precisely the fixed-class boundary $\rho^\star(\beta)$ stated in Section~\ref{sec:theory-boundary} of the main paper.

For strict dependence on $s$, fix $r,q$ in the positive-rate region
and write $z=x_s(q)/s$ and $y=2x_s(q)/(s+\sqrt{s^2+4rx_s(q)})$.
Then $z>y>1$. Implicit differentiation of $g_s(x_s(q))=q$ gives
\[
 \partial_s x_s(q)=A(z),\qquad
 A(u)=\frac{\log u}{1-1/u},\qquad
 A'(u)=\frac{u-1-\log u}{(u-1)^2}>0.
\]
The envelope theorem now yields
\[
 \partial_s\Psi_{r,s}(q)
 =(1-1/y)\{A(z)-A(y)\}>0.
\]
In the zero-rate region the rate is zero; its onset
$g_s(s+r)=r-s\log(1+r/s)$ decreases with $s$.
Thus $\Psi$ is nondecreasing in $s$ everywhere and strictly increases
once positive. At a boundary point every maximizing $q$ has positive
rate, by the optimizer argument above. If $s_2>s_1$, evaluating at
$r=\xi^\star(\beta,s_1)$ gives $E_{s_2}(q)\le E_{s_1}(q)\le0$,
with strict inequality at every zero of $E_{s_1}$.
Compactness implies $\max_qE_{s_2}(q)<0$, and hence
$\xi^\star(\beta,s_2)>\xi^\star(\beta,s_1)$.
The same conclusion with $s_1=0$ follows by continuity at zero and
strict increase on positive $s$ (apply it first between $s_2/2$ and $s_2$).

As $s\to0$, $x_s(q)\to q$ uniformly on $[0,1]$:
$x_s(q)-q=s+s\log(x_s(q)/s)$ lies between $s$ and its value at
$q=1$, which tends to zero. The explicit formula for $J$ then gives
uniform convergence of $\Psi_{r,s}$ to $(\sqrt q-\sqrt r)_+^2$
for every fixed $r>0$. Strict signs on each side of the classical
boundary bracket the roots and prove their convergence.

Finally let $s\to\infty$ and set $r=a\sqrt{2s}$ for fixed $a>0$.
Taylor expansion of $g_s$ yields, uniformly for $q\in[0,1]$,
\[
 x_s(q)=s+\sqrt{2sq}+O(1).
\]
For $x=x_s(q)>s+r$, the optimizer $\theta$ in
\eqref{eq:J-variational} is $O(s^{-1/2})$ uniformly. Expanding at zero,
\[
 \theta x-\Lambda(\theta)
 =\theta(x-s-r)-\tfrac12(s+2r)\theta^2
   +O((s+r)\theta^3).
\]
Optimization, with the zero-rate case included, gives
\[
 \Psi_{a\sqrt{2s},s}(q)
 =\frac{(x_s(q)-s-r)_+^2}{2(s+2r)}+o(1)
 =(\sqrt q-a)_+^2+o(1)
\]
uniformly on $[0,1]$. The limiting boundary functional has its unique zero
at $a^2=\rho^\star(\beta)$, so bracketing again proves
the claimed large-$s$ equivalence. This is a limit of the fixed-$s$
boundary, not a theorem for arbitrary joint sequences $s=s_p\to\infty$.
\end{proof}

\section{Proof of Theorem~\ref{thm:classification} (Misclassification rate of DP-HCT)}
\label{app:classification-proof}

We prove the classification theorem under the spherical homogeneous contrast
model. 
The proof analyzes the inner-product margin after HCT feature selection.
Fix a true class \(c^\star\in[C]\) and a competing class \(c\ne c^\star\).
For a useful feature $j$, the population contribution to the
pairwise margin is
\begin{equation}
\label{eq:spherical-pairwise-pop-margin-appF}
    \mu_{c^\star}^{(j)}
    \bigl(\mu_{c^\star}^{(j)}-\mu_c^{(j)}\bigr)
    =
    \tau^2
    v_{c^\star}^{(j)}
    \bigl(v_{c^\star}^{(j)}-v_c^{(j)}\bigr).
\end{equation}
The spherical symmetry of \(v^{(j)}\) gives positive average separation for
every fixed class pair. Indeed,
\begin{equation}
\label{eq:spherical-pairwise-mean-appF}
    \mathbb E
    \left[
    v_{c^\star}^{(j)}
    \bigl(v_{c^\star}^{(j)}-v_c^{(j)}\bigr)
    \right]
    =
    \frac{C}{C-1}.
\end{equation}
Therefore, once HCT selects polynomially many true coordinates, the aggregate
spherical contrast over the selected features produces a positive
margin of order \(\tau^2 \mathrm{TP}\), where
\[
    \mathrm{TP}:=|\hat{\mathcal J}\cap\mathcal J|.
\]
The rest of the proof shows that this positive population margin dominates
both the training-noise terms introduced by estimating the centroids on the
selected features and the variance contributed by selected null features.

\subsection{Proof roadmap}
\label{app:classification-roadmap}

Fix the true test label \(c^\star\) and a competitor \(c\ne c^\star\).
The proof has seven steps.
\begin{enumerate}[label=(I.\arabic*),leftmargin=*]
\item \textbf{Pairwise reduction.}
Misclassification is contained in the union of the pairwise margin failures
\(\{\Delta_c\le0\}\).

\item \textbf{Conditional Gaussianity.}
Given the training data, latent means, and true test label, each margin \(\Delta_c\) is Gaussian
with conditional mean \(m_c(\mathcal D)\) and variance \(v_c(\mathcal D)\).

\item \textbf{HCT count consequences.}
HCT adaptivity makes $\mathrm{TP}^2/(\mathrm{TP}+\mathrm{FP})$
polynomially large. Consequently, for some fixed $d>0$,
\[
 \mathrm{TP}\ge p^d,\qquad
 \frac{\mathrm{TP}^2}{\mathrm{FP}+1}\ge p^d
\]
with probability $1-o(1)$. We do not need, and generally do not have,
$\mathrm{FP}/\mathrm{TP}\to0$.

\item \textbf{Overshoot control.}
Selected null and selected signal ANOVA statistics are \(O(\log p)\)
uniformly. This keeps empirical selected contrasts on the rare-and-weak
scale.

\item \textbf{Spherical pairwise diversity.}
For every fixed pair \(c^\star\ne c\), spherical symmetry implies positive
average pairwise separation:
\[
    \mathbb E
    \left[
    v_{c^\star}^{(j)}
    \bigl(v_{c^\star}^{(j)}-v_c^{(j)}\bigr)
    \right]
    =
    \frac{C}{C-1}.
\]
Since HCT selects polynomially many true positives, these pairwise
separations concentrate over \(\hat{\mathcal J}\cap\mathcal J\), giving an
aggregate population margin of order \(\tau^2\mathrm{TP}\) uniformly over
class pairs.

\item \textbf{No-sample-splitting stability.}
Although the same data are used for screening and centroid estimation, the
selection-induced signal-noise term is bounded below by
\(-o_\Prob(\tau^2\mathrm{TP})\). The reason is that ANOVA selection is
radial in the projected class-mean vector and does not create an aggregate
negative drift large enough to cancel the spherical signal margin.

\item \textbf{SNR divergence.}
The conditional mean is at least order \(\tau^2\mathrm{TP}\), while the
conditional variance is bounded by a true-signal term plus a selected-null
term. The first is dominated by polynomial growth of \(\mathrm{TP}\); the
second is dominated by \(\mathrm{TP}^2/(\mathrm{FP}+1)\to\infty\).
\end{enumerate}

\subsection{Reduction to pairwise margins}
\label{app:cls-reduction}

Recall the score classifier~\eqref{eq:ncm}, $\hat h(x)=\arg\max_{a\in[C]}V_a(x)$ with
\[
    V_a(x):=\langle \hat\mu_a[\hat{\mathcal J}],x[\hat{\mathcal J}]\rangle.
\]
For a test point $X_{n+1}$ with true label $c^\star$, define the pairwise margin against competitor $c\ne c^\star$ by
\begin{equation}
\label{eq:Delta-c-def-new}
    \Delta_c
    :=
    V_{c^\star}(X_{n+1})-V_c(X_{n+1})
    =
    \sum_{j\in\hat{\mathcal J}}
    \bigl(\hat{\mu}_{c^\star,j}-\hat{\mu}_{c,j}\bigr)X_{n+1,j}.
\end{equation}
Then
\begin{equation}
\label{eq:union-bound-new}
    \Prob\{\hat h(X_{n+1})\ne c^\star\}
    \le
    \sum_{c\ne c^\star}\Prob\{\Delta_c\le0\}.
\end{equation}
We will prove a super-polynomial conditional error bound on one event that holds simultaneously for all class pairs with probability $1-o(1)$. The complement of that event is counted once when removing the conditioning.

\subsection{Conditional Gaussianity of pairwise margins}
\label{app:cls-gaussian}

For this section, let $\mathcal D$ denote the sigma-field generated by
the training data, the latent signal set, all latent class means, and the
fixed true test label $c^\star$. This enlarged conditioning is only a proof
device; the classifier observes the training data and test feature vector.
Define
\[
    a_j(\mathcal D)
    :=
    \hat{\mu}_{c^\star,j}-\hat{\mu}_{c,j}.
\]

\begin{lemma}[Conditional Gaussianity]
\label{lem:cond-gauss}
Conditional on $\mathcal D$, for each $c\ne c^\star$,
\begin{equation}
\label{eq:cond-gauss-new}
    \Delta_c\mid\mathcal D
    \sim
    \mathcal N\bigl(m_c(\mathcal D),v_c(\mathcal D)\bigr),
\end{equation}
where
\begin{align}
\label{eq:mc-def-new}
    m_c(\mathcal D)
    &=
    \sum_{j\in\hat{\mathcal J}}a_j(\mathcal D)\mu_{c^\star}^{(j)},\\
\label{eq:vc-def-new}
    v_c(\mathcal D)
    &=
    \sigma^2\sum_{j\in\hat{\mathcal J}}a_j(\mathcal D)^2.
\end{align}
\end{lemma}

\begin{proof}
Given $\mathcal D$, the coefficients $a_j(\mathcal D)$ are fixed. Write $X_{n+1,j}=\mu_{c^\star}^{(j)}+\varepsilon_j^\star$, where the test noises $\varepsilon_j^\star$ are independent $\mathcal N(0,\sigma^2)$ and independent of $\mathcal D$. Substitution in~\eqref{eq:Delta-c-def-new} gives a deterministic mean plus a centered Gaussian sum with variance $\sigma^2\sum_j a_j^2$.
\end{proof}

By Mills' bound, on the event $m_c(\mathcal D)>0$,
\begin{equation}
\label{eq:mills-applied-new}
    \Prob(\Delta_c\le0\mid\mathcal D)
    \le
    \exp\left\{-\frac{m_c(\mathcal D)^2}{2v_c(\mathcal D)}\right\}.
\end{equation}
Thus the task is to prove $m_c(\mathcal D)^2/v_c(\mathcal D)\to\infty$ with high probability.

\subsection{Count consequences of HCT adaptivity}
\label{app:cls-count}

Let $\hat t=\pi_{(j^\star)}$ be the HCT threshold of
Algorithm~\ref{alg:dphct}, and write
\[
 \mathrm{TP}=|\hat{\mathcal J}\cap\mathcal J|,\qquad
 \mathrm{FP}=|\hat{\mathcal J}\cap\mathcal J^c|,\qquad
 N=\mathrm{TP}+\mathrm{FP}.
\]
\begin{lemma}[Selected counts]
\label{lem:count-control}
If $\sup_{0<q<1}E_p(q)>0$, there is a fixed $d>0$ such that, with
probability $1-o(1)$,
\begin{align}
\label{eq:TP-diverges-new}
 \mathrm{TP}&\ge p^d,\\
\label{eq:FP-size-new}
 \frac{\mathrm{TP}^2}{N}&\ge p^d,\\
\label{eq:TP2-over-FP-new}
 \frac{\mathrm{TP}^2}{\mathrm{FP}+1}&\ge p^d.
\end{align}
\end{lemma}
\begin{proof}
Theorem~\ref{thm:detection}(b) proves the middle inequality with some
positive exponent. Since $\mathrm{TP}\le N$,
$\mathrm{TP}\ge\mathrm{TP}^2/N$. Also $\mathrm{FP}+1\le N+1\le2N$.
These two inequalities prove the remaining conclusions after slightly
decreasing the exponent to absorb constants. No restriction on a
random selected exponent $-\log\hat t/\log p$ is required.
\end{proof}

\subsection{Overshoot control}
\label{app:cls-overshoot}

The next lemmas say that selected empirical ANOVA contrasts remain on the $\log p$ scale. This is not a weakness assumption; it is exactly what the rare-and-weak calibration $\lambda_j\asymp\log p$ predicts.

\begin{lemma}[Null overshoot]
\label{lem:null-overshoot}
For any $B>0$, there exists $M_0<\infty$ such that
\[
    \Prob\left(\max_{j\in\mathcal J^c}T_j>M_0\log p\right)
    \le p^{-B}.
\]
Consequently, with probability $1-o(1)$,
\[
    \max_{j\in\hat{\mathcal J}\cap\mathcal J^c}T_j\le M_0\log p.
\]
\end{lemma}

\begin{proof}
For $j\in\mathcal J^c$, $T_j\sim\chi_k^2$. The standard Chernoff bound gives, for $x>k$,
\[
    \Prob(\chi_k^2\ge x)
    \le
    \exp\left\{-\frac12\bigl[x-k-k\log(x/k)\bigr]\right\}.
\]
Since $k=O(\log p)$ in the regimes considered, taking $x=M_0\log p$ with $M_0$ large enough gives probability at most $p^{-(B+2)}$. A union bound over at most $p$ null coordinates proves the claim.
\end{proof}

\begin{lemma}[Signal overshoot]
\label{lem:signal-overshoot}
For any $B>0$, there exists $M_1<\infty$ such that
\[
    \Prob\left(\max_{j\in\mathcal J}T_j>M_1\log p\right)
    \le p^{-B}.
\]
\end{lemma}

\begin{proof}
For \(j\in\mathcal J\), \(T_j\sim\chi_k^2(\lambda_j)\) with
\(k=O(\log p)\), and under the spherical homogeneous calibration
\eqref{eq:spherical-contrast}--\eqref{eq:lambda-calibration},
\[
    \lambda_j=2r\log p
\]
exactly. A noncentral chi-squared Chernoff bound states that, for $u>0$,
\[
    \Prob\{\chi_k^2(\lambda)\ge k+\lambda+2\sqrt{(k+2\lambda)u}+2u\}
    \le e^{-u}.
\]
Taking $u=(B+2)\log p$ shows that $T_j\le M_1\log p$ with probability at least $1-p^{-(B+2)}$, uniformly over $j\in\mathcal J$, for $M_1$ large enough. Since $|\mathcal J|\le p$, a union bound proves the result.
\end{proof}

\begin{corollary}[Selected empirical contrast bound]
\label{cor:selected-contrast-bound}
With probability $1-o(1)$, for every selected feature $j\in\hat{\mathcal J}$ and every pair $a,b\in[C]$,
\begin{equation}
\label{eq:selected-contrast-bound-new}
    \bigl(\hat{\mu}_{a,j}-\hat{\mu}_{b,j}\bigr)^2
    \le
    \frac{4\sigma^2}{n_c}M\log p,
\end{equation}
where $M:=\max\{M_0,M_1\}$.
\end{corollary}

\begin{proof}
The deterministic contrast bound
\[
    (\hat{\mu}_{a,j}-\hat{\mu}_{b,j})^2
    \le
    \frac{4\sigma^2}{n_c}T_j
\]
follows from Cauchy--Schwarz: $\hat{\mu}_{a,j}-\hat{\mu}_{b,j}=(e_a-e_b)^\top P\hat{\mu}_{\cdot,j}$ and $\|e_a-e_b\|^2=2$, while $T_j=(n_c/\sigma^2)\|P\hat{\mu}_{\cdot,j}\|^2$. Combining this deterministic inequality with Lemmas~\ref{lem:null-overshoot} and~\ref{lem:signal-overshoot} gives~\eqref{eq:selected-contrast-bound-new}.
\end{proof}

\subsection{Spherical pairwise diversity}
\label{app:cls-diversity}

We next record the pairwise diversity properties of the spherical contrast
directions. These properties are used to lower-bound the population part of
the pairwise inner-product margin and to control the signal part of its
conditional variance.

For \(c^\star\ne c\), define
\[
    B_j^{(c^\star,c)}
    :=
    v_{c^\star}^{(j)}
    \bigl(v_{c^\star}^{(j)}-v_c^{(j)}\bigr),
    \qquad
    D_j^{(c^\star,c)}
    :=
    \bigl(v_{c^\star}^{(j)}-v_c^{(j)}\bigr)^2 .
\]
The variables \(B_j^{(c^\star,c)}\) control the population pairwise margin,
whereas \(D_j^{(c^\star,c)}\) controls the signal part of the conditional
variance.

\begin{lemma}[Spherical pairwise diversity]
\label{lem:diversity}
Assume the spherical homogeneous contrast model
\eqref{eq:spherical-contrast}--\eqref{eq:lambda-calibration}.
Let
\[
    \mathrm{TP}:=|\hat{\mathcal J}\cap\mathcal J|.
\]
On the event \(\mathrm{TP}\ge p^d\) for some fixed \(d>0\), for every fixed pair
\(c^\star\ne c\),
\[
    \sum_{j\in\hat{\mathcal J}\cap\mathcal J}
    v_{c^\star}^{(j)}
    \bigl(v_{c^\star}^{(j)}-v_c^{(j)}\bigr)
    =
    \frac{C}{C-1}\,\mathrm{TP}
    +
    o_\Prob(\mathrm{TP}),
\]
and
\[
    \sum_{j\in\hat{\mathcal J}\cap\mathcal J}
    \bigl(v_{c^\star}^{(j)}-v_c^{(j)}\bigr)^2
    =
    \frac{2C}{C-1}\,\mathrm{TP}
    +
    o_\Prob(\mathrm{TP}).
\]
In particular, with probability \(1-o(1)\), uniformly over all
\(c^\star\ne c\),
\[
    \sum_{j\in\hat{\mathcal J}\cap\mathcal J}
    v_{c^\star}^{(j)}
    \bigl(v_{c^\star}^{(j)}-v_c^{(j)}\bigr)
    \ge
    \frac12\,\mathrm{TP},
\]
and
\[
    \sum_{j\in\hat{\mathcal J}\cap\mathcal J}
    \bigl(v_{c^\star}^{(j)}-v_c^{(j)}\bigr)^2
    \le
    4\,\mathrm{TP}.
\]
Moreover, for every fixed \(c^\star\),
\[
    \sum_{j\in\hat{\mathcal J}\cap\mathcal J}
    \bigl(v_{c^\star}^{(j)}\bigr)^2
    =
    \mathrm{TP}
    +
    o_\Prob(\mathrm{TP}),
\]
and hence, with probability \(1-o(1)\),
\[
    \sum_{j\in\hat{\mathcal J}\cap\mathcal J}
    \bigl(v_{c^\star}^{(j)}\bigr)^2
    \le
    2\,\mathrm{TP}
\]
uniformly over \(c^\star\in[C]\).
\end{lemma}

\begin{proof}
Let
\[
    U\sim
    \operatorname{Unif}
    \left\{
    u\in\mathbf 1^\perp:\|u\|_2=1
    \right\},
    \qquad
    v=\sqrt C\,U.
\]
By rotational symmetry on the \((C-1)\)-dimensional subspace
\(\mathbf 1^\perp\),
\[
    \mathbb E[UU^\top]
    =
    \frac{1}{C-1}P.
\]
Therefore,
\[
    \mathbb E[vv^\top]
    =
    \frac{C}{C-1}P.
\]
Since \(P_{aa}=1-1/C\) and \(P_{ab}=-1/C\) for \(a\ne b\), we get
\[
    \mathbb E[v_a^2]
    =
    \frac{C}{C-1}\left(1-\frac1C\right)
    =
    1,
\]
and, for \(a\ne b\),
\[
    \mathbb E[v_av_b]
    =
    \frac{C}{C-1}\left(-\frac1C\right)
    =
    -\frac{1}{C-1}.
\]
Thus, for \(c^\star\ne c\),
\[
    \mathbb E
    \left[
    v_{c^\star}
    \bigl(v_{c^\star}-v_c\bigr)
    \right]
    =
    \mathbb E[v_{c^\star}^2]
    -
    \mathbb E[v_{c^\star}v_c]
    =
    1+\frac{1}{C-1}
    =
    \frac{C}{C-1},
\]
and
\[
    \mathbb E
    \left[
    \bigl(v_{c^\star}-v_c\bigr)^2
    \right]
    =
    \mathbb E[v_{c^\star}^2]
    +
    \mathbb E[v_c^2]
    -
    2\mathbb E[v_{c^\star}v_c]
    =
    2+\frac{2}{C-1}
    =
    \frac{2C}{C-1}.
\]

It remains to justify that these averages still concentrate after HCT
selection. Conditional on \(j\in\mathcal J\), the statistic used for
selection is the ANOVA statistic. Under the spherical homogeneous calibration,
its non-centrality is
\[
    \lambda_j=2r\log p
\]
for every direction \(v^{(j)}\). Consequently, the entire conditional distribution of a useful
feature's p-value depends on its spherical direction only through
\(\|v^{(j)}\|_2^2=C\), which is fixed. By independence across features,
the vector of useful directions is independent of the full vector of
p-values, conditional on the signal set. This also applies to the adaptive
HCT selection, which is a measurable function of that p-value vector. Hence, conditional on the selected
true-positive set \(\hat{\mathcal J}\cap\mathcal J\) and on its size
\(\mathrm{TP}\), the directions
\[
    \{v^{(j)}:j\in\hat{\mathcal J}\cap\mathcal J\}
\]
remain iid draws from the spherical law.

For each fixed pair \(c^\star\ne c\), the variables
\[
    B_j^{(c^\star,c)}
    =
    v_{c^\star}^{(j)}
    \bigl(v_{c^\star}^{(j)}-v_c^{(j)}\bigr)
\]
are iid with finite moments of all orders and mean \(C/(C-1)\). Therefore,
conditional on \(\mathrm{TP}\ge p^d\),
\[
    \frac1{\mathrm{TP}}
    \sum_{j\in\hat{\mathcal J}\cap\mathcal J}
    B_j^{(c^\star,c)}
    =
    \frac{C}{C-1}
    +
    o_\Prob(1).
\]
Similarly,
\[
    \frac1{\mathrm{TP}}
    \sum_{j\in\hat{\mathcal J}\cap\mathcal J}
    D_j^{(c^\star,c)}
    =
    \frac{2C}{C-1}
    +
    o_\Prob(1),
\]
and
\[
    \frac1{\mathrm{TP}}
    \sum_{j\in\hat{\mathcal J}\cap\mathcal J}
    \bigl(v_{c^\star}^{(j)}\bigr)^2
    =
    1+o_\Prob(1).
\]

The number of class pairs is at most \(C^2\). In the regimes considered in
Theorem~\ref{thm:classification}, \(C=O(\log p)\), while
Lemma~\ref{lem:count-control} gives that \(\mathrm{TP}\) is polynomially
large with probability \(1-o(1)\). Each displayed summand is bounded in absolute value by $4C$.
Hoeffding's inequality bounds deviations of their averages by
$2\exp\{-c_\varepsilon\mathrm{TP}/C^2\}$ for every fixed
$\varepsilon>0$. A union bound over $O(C^2)$ class pairs upgrades
the fixed-pair convergence to uniform convergence over all
$c^\star\ne c$. The displayed
high-probability bounds follow immediately.
\end{proof}

\subsection{One-sided stability of selected signal noise}
\label{app:cls-stability}

The delicate point in the proof is that $\hat{\mathcal J}$ is selected
using the same training noise that appears in the empirical centroids.
We first record a symmetrization identity that controls this dependence,
and then use it to prove the one-sided stability bound.

\begin{lemma}[Symmetrization]\label{lem:symmetrization} 
Let $Z_j\mid\mu^{(j)}\sim\mathcal N(0,(\sigma^2/n_c)I_C)$, let
$P=I_C-C^{-1}\mathbf 1\mathbf 1^\top$ denote the orthogonal projection
onto $\mathbf 1^\perp$, and let
\[
    A_j(q)
    :=
    \bigl\{\,\|P\mu^{(j)}+PZ_j\|^2\ge t_q\,\bigr\}
\]
for a deterministic threshold $t_q>0$. Then there exists a scalar
$b_{j,q}\ge 0$ such that
\begin{equation}\label{eq:symmetrization}
    \E\bigl[\,PZ_j\mathbf 1_{A_j(q)}\mid\mu^{(j)}\,\bigr]
    =
    b_{j,q}\,P\mu^{(j)}.
\end{equation}
\end{lemma}

\begin{proof}
All expectations are conditional on $\mu^{(j)}$; we suppress this from
the notation. Write $W:=PZ_j$ and $m:=P\mu^{(j)}$, both vectors in the
subspace $\mathcal{V}:=\mathbf 1^\perp$. Then $W$ is a centered Gaussian on $\mathcal{V}$
with covariance $(\sigma^2/n_c)P$, and $A_j(q)=\{\|m+W\|^2\ge t_q\}$
depends on $W$ only through $\|m+W\|^2$.

\medskip
\noindent\emph{Step 1: the conditional mean is parallel to $m$.}
If $m=0$, then $A_j(q)$ depends only on $\|W\|^2$, so
$\E[W\mathbf 1_{A_j(q)}]=0$ by symmetry of $W$, and the claim holds
with $b_{j,q}=0$. Assume $m\ne 0$.

Let $R$ be any orthogonal transformation of $\mathcal{V}$ that fixes $m$, i.e.,
$Rm=m$. Then $\|m+RW\|^2=\|R(m+W)\|^2=\|m+W\|^2$, so $A_j(q)$ is
$R$-invariant. Since the law of $W$ on $\mathcal{V}$ is also $R$-invariant
(rotational invariance of the Gaussian),
\[
    \E[W\mathbf 1_{A_j(q)}]
    =\E[RW\mathbf 1_{A_j(q)}]
    =R\,\E[W\mathbf 1_{A_j(q)}].
\]
Hence $v:=\E[W\mathbf 1_{A_j(q)}]$ is fixed by every orthogonal $R$ on
$\mathcal{V}$ with $Rm=m$. The stabilizer of $m$ acts as the full orthogonal
group on $m^\perp\cap\mathcal{V}$, so any vector fixed by all such $R$ must lie
in $\mathrm{span}(m)$. Therefore
\[
    \E[W\mathbf 1_{A_j(q)}]=b_{j,q}\,m
    \qquad\text{for some scalar } b_{j,q}\in\R.
\]

\medskip
\noindent\emph{Step 2: the scalar is nonnegative.}
Decompose $W$ along $m$ and its orthogonal complement in $\mathcal{V}$:
\[
    W=X\hat m+W_\perp,
    \qquad
    \hat m:=m/\|m\|,
    \quad X:=\langle W,\hat m\rangle,
    \quad W_\perp\perp\hat m.
\]
Under the Gaussian law, $X\sim\mathcal N(0,\sigma_*^2)$ with
$\sigma_*^2:=\sigma^2/n_c$, and $X$ is independent of $W_\perp$.
Taking the inner product of Step~1 with $\hat m$ gives
\[
    b_{j,q}\,\|m\|=\E[X\mathbf 1_{A_j(q)}],
\]
so it suffices to show $\E[X\mathbf 1_{A_j(q)}]\ge 0$.

Conditional on $W_\perp$, the event $A_j(q)$ becomes
\[
    \bigl\{(\|m\|+X)^2+\|W_\perp\|^2\ge t_q\bigr\}.
\]
We consider two cases.

\emph{Case (i): $\|W_\perp\|^2\ge t_q$.} The event holds for all $X$,
so the conditional contribution is $\E[X]=0$.

\emph{Case (ii): $\|W_\perp\|^2<t_q$.} Setting
$\rho:=\sqrt{t_q-\|W_\perp\|^2}>0$, the event becomes
$\{|\|m\|+X|\ge\rho\}$, i.e.,
\[
    \{X\le a\}\cup\{X\ge b\},
    \qquad a:=-\|m\|-\rho,\quad b:=-\|m\|+\rho.
\]
Note that $a+b=-2\|m\|\le 0$, so $|a|\ge|b|$. Using the Gaussian
truncation identities
$\E[X\mathbf 1_{\{X\le a\}}]=-\sigma_*^2\phi_{\sigma_*}(a)$
and
$\E[X\mathbf 1_{\{X\ge b\}}]=\sigma_*^2\phi_{\sigma_*}(b)$,
where $\phi_{\sigma_*}$ denotes the $\mathcal N(0,\sigma_*^2)$
density, the conditional contribution is
\[
    \sigma_*^2\bigl(\phi_{\sigma_*}(b)-\phi_{\sigma_*}(a)\bigr)\ge 0,
\]
because $|b|\le|a|$ and $\phi_{\sigma_*}$ is decreasing in $|\cdot|$.

Both cases yield a nonnegative contribution, so integrating over
$W_\perp$ gives $\E[X\mathbf 1_{A_j(q)}]\ge 0$, hence $b_{j,q}\ge 0$.
\end{proof}

\begin{remark}
Geometrically, conditioning on the radial event
$\{\|m+W\|^2\ge t_q\}$ biases $W$ along $+m$ rather than $-m$:
reaching the threshold on the $+m$ side requires a smaller deviation
than on the $-m$ side, while rotational symmetry kills all components
of the conditional mean orthogonal to $m$.
\end{remark} 

For exact F p-values the threshold is random, but independent of the
class-mean noise. Conditional on the residual sum of squares $U_j$,
the event $\{\pi_j\le p^{-q}\}$ is
\[
 \{\|P\mu^{(j)}+PZ_j\|^2
      \ge(\sigma^2/n_c)\,k f_{k,m}^{-1}(1-p^{-q})\,U_j/m\}.
\]
Here $f_{k,m}^{-1}$ denotes the central F quantile function.
Apply Lemma~\ref{lem:symmetrization} at this threshold and average over $U_j$.
Its conclusion remains valid with a nonnegative scalar depending only
on the common signal norm, $q$, and the model parameters.

\begin{lemma}[One-sided selected-signal stability]
\label{lem:one-sided-stability}
For $a\ne b$ and $t\in[0,1]$, let
\[
 R_{a,b}(t)=\sum_{j\in\mathcal J}
 \mu_a^{(j)}(\bar\varepsilon_{a,j}-\bar\varepsilon_{b,j})
 \mathbf1\{\pi_j\le t\},\qquad
 \bar\varepsilon_{a,j}=\hat\mu_{a,j}-\mu_a^{(j)}.
\]
For every fixed $d>0$, uniformly over all class pairs and all thresholds
with $\mathrm{TP}(t)\ge p^d$,
\[
 R_{a,b}(t)\ge-o_\Prob(\tau^2\mathrm{TP}(t)).
\]
In particular, writing $R_c(t)=R_{c^\star,c}(t)$,
\[
 R_c(\hat t)\ge-o_\Prob(\tau^2\mathrm{TP})
\]
simultaneously for all ordered class pairs.
\end{lemma}

\begin{proof}
Condition on $\mathcal J$; expectations below average over both spherical
directions and training observations, including residuals. For a fixed
threshold $t$ put
\[
 W_j^{a,b}=\mu_a^{(j)}(\bar\varepsilon_{a,j}-\bar\varepsilon_{b,j}),
 \qquad a_p(t)=\Prob(\pi_j\le t\mid j\in\mathcal J),
 \qquad m_T(t)=|\mathcal J|a_p(t).
\]
The extended symmetrization identity for exact F selection gives
\[
 \E[W_j^{a,b}\mathbf1\{\pi_j\le t\}\mid\mu^{(j)}]
 =b_t\mu_a^{(j)}(\mu_a^{(j)}-\mu_b^{(j)}),\qquad b_t\ge0.
\]
The scalar is common to all directions because their norms are equal.
Averaging over the spherical direction therefore yields
\begin{equation}
\label{eq:selected-noise-nonnegative-mean}
 \E[W_j^{a,b}\mathbf1\{\pi_j\le t\}]
 =b_t\tau^2\frac{C}{C-1}\ge0.
\end{equation}
There is no additional selection indicator outside this expectation;
nonnegativity is asserted after averaging over the direction.

\paragraph{Truncation and selected second moments.}
Put $L=\log p$, $\sigma_*^2=\sigma^2/n_c$, and $\xi_j=PZ_j/\sigma_*$.
Then $\|\xi_j\|^2\sim\chi^2_{C-1}$ and
$\sigma_*^2=C\tau^2/(2rL)$. Choose a fixed $M$ large enough that
$B_j=\{\|\xi_j\|^2\le ML\}$ satisfies $\Prob(B_j^c)\le p^{-12}$.
Since $C=O(L)$, the chi-square tail bound permits this choice uniformly
over features. On $B_j$, for every class pair,
\[
 |W_j^{a,b}|\le\tau\sqrt C\,\sqrt2\sigma_*\sqrt{ML}
 \le K_0\tau^2C.
\]
For $\widetilde H_j(t)=W_j^{a,b}\mathbf1_{B_j}\mathbf1\{\pi_j\le t\}$,
\begin{equation}
\label{eq:selected-signal-noise-second-moment}
 \E\widetilde H_j(t)^2\le K_0^2\tau^4 C^2a_p(t).
\end{equation}
Also $\E(W_j^{a,b})^2\le2\tau^2C\sigma_*^2$.
Cauchy--Schwarz therefore bounds the lost first moment, uniformly in $t$,
by $K\tau^2 C p^{-6}$ (for all sufficiently large $p$).
After summing over at most $p$ features this is $o(\tau^2)$, and
$\Prob(\bigcup_{j\in\mathcal J}B_j^c)=o(1)$.

\paragraph{Uniform concentration over deterministic thresholds.}
Use the auxiliary quantile grid $\mathcal T_p$ of
Lemma~\ref{lem:uniform-threshold-counts}. The truncated summands are
independent across features conditional on $\mathcal J$.
Bernstein's inequality, the selected second-moment bound, and a union
bound over $O(p^4)$ grid points and $O(C^2)$ class pairs give, with
probability $1-o(1)$,
\[
 \sum_{j\in\mathcal J}(\widetilde H_j(t)-\E\widetilde H_j(t))
 \ge-K_1\tau^2C\{\sqrt{m_T(t)L}+L\}
\]
simultaneously on the grid. The nonnegative untruncated mean in
\eqref{eq:selected-noise-nonnegative-mean}, the lost-mean bound,
and the event that all $B_j$ hold imply
\[
 R_{a,b}(t)\ge-K_1\tau^2C\{\sqrt{m_T(t)L}+L\}-o(\tau^2)
\]
at every grid point, simultaneously for all pairs.

\paragraph{Extension to all thresholds and adaptive selection.}
On the no-collision event of Lemma~\ref{lem:uniform-threshold-counts},
moving any threshold to its lower adjacent grid point changes the sum
by at most one truncated mark, whose absolute value is at most
$K_0\tau^2C$. This is absorbed by the $\tau^2CL$ remainder.
Since $m_T$ is nondecreasing, the same lower bound, with an enlarged
constant, holds simultaneously for every $t\in[0,1]$.
On thresholds with $\mathrm{TP}(t)\ge p^d$, that lemma also gives
$m_T(t)=\mathrm{TP}(t)(1+o(1))$ uniformly. Dividing by
$\tau^2\mathrm{TP}(t)$ bounds the negative relative error by
\[
 O\!\left(C\sqrt{\frac{L}{\mathrm{TP}(t)}}
       +\frac{CL}{\mathrm{TP}(t)}\right)+o(1)=o(1),
\]
because $C=O(L)$ and $\mathrm{TP}(t)\ge p^d$.
Lemma~\ref{lem:count-control} puts the selected threshold in this region
with probability $1-o(1)$. Evaluation of the simultaneous bound at
$\hat t$ proves the result without conditioning on the random threshold
as though it were fixed or independent of the training noise.
\end{proof}

\subsection{SNR divergence and conclusion}
\label{app:cls-snr}

We now combine the preceding ingredients. Fix the true class \(c^\star\) and
a competitor \(c\ne c^\star\). Recall from Lemma~\ref{lem:cond-gauss}
that, conditional on the enlarged sigma-field \(\mathcal D\), the pairwise margin
\(\Delta_c\) is Gaussian with mean
\[
    m_c(\mathcal D)
    =
    \sum_{j\in\hat{\mathcal J}}
    \widehat d_{c^\star c}(j)\mu_{c^\star}^{(j)},
    \qquad
    \widehat d_{c^\star c}(j)
    :=
    \hat{\mu}_{c^\star,j}-\hat{\mu}_{c,j},
\]
and variance
\[
    v_c(\mathcal D)
    =
    \sigma^2
    \sum_{j\in\hat{\mathcal J}}
    \widehat d_{c^\star c}(j)^2.
\]
We show that
\[
    \frac{m_c(\mathcal D)^2}{v_c(\mathcal D)}
    \to\infty
\]
in probability, uniformly over \(c\ne c^\star\).

First decompose the conditional mean into its population part and its
training-noise part:
\[
\begin{aligned}
    m_c(\mathcal D)
    &=
    \sum_{j\in\hat{\mathcal J}}
    \mu_{c^\star}^{(j)}
    \bigl(\mu_{c^\star}^{(j)}-\mu_c^{(j)}\bigr)
    \\
    &\qquad+
    \sum_{j\in\hat{\mathcal J}}
    \mu_{c^\star}^{(j)}
    \bigl(\bar\varepsilon_{c^\star,j}-\bar\varepsilon_{c,j}\bigr).
\end{aligned}
\]
Null coordinates have \(\mu_{c^\star}^{(j)}=0\), so only selected signals
contribute. Therefore
\[
\begin{aligned}
    m_c(\mathcal D)
    &=
    \sum_{j\in\hat{\mathcal J}\cap\mathcal J}
    \mu_{c^\star}^{(j)}
    \bigl(\mu_{c^\star}^{(j)}-\mu_c^{(j)}\bigr)
    +
    R_c(\hat t).
\end{aligned}
\]
Using \(\mu^{(j)}=\tau v^{(j)}\), the population term is
\[
    \sum_{j\in\hat{\mathcal J}\cap\mathcal J}
    \mu_{c^\star}^{(j)}
    \bigl(\mu_{c^\star}^{(j)}-\mu_c^{(j)}\bigr)
    =
    \tau^2
    \sum_{j\in\hat{\mathcal J}\cap\mathcal J}
    v_{c^\star}^{(j)}
    \bigl(v_{c^\star}^{(j)}-v_c^{(j)}\bigr).
\]
By Lemma~\ref{lem:diversity}, with probability \(1-o(1)\), uniformly over
all \(c^\star\ne c\),
\[
    \sum_{j\in\hat{\mathcal J}\cap\mathcal J}
    v_{c^\star}^{(j)}
    \bigl(v_{c^\star}^{(j)}-v_c^{(j)}\bigr)
    \ge
    \frac12\,\mathrm{TP}.
\]
By Lemma~\ref{lem:one-sided-stability},
\[
    R_c(\hat t)
    \ge
    -o_\Prob(\tau^2\mathrm{TP}).
\]
Consequently,
\begin{equation}
\label{eq:mean-lower-spherical-appF}
    m_c(\mathcal D)
    \ge
    \left(\frac12-o_\Prob(1)\right)\tau^2\mathrm{TP}
    \gtrsim
    \tau^2\mathrm{TP},
\end{equation}
uniformly over \(c\ne c^\star\).

We next bound the conditional variance. Write
\[
    \widehat d_{c^\star c}(j)
    =
    \bigl(\mu_{c^\star}^{(j)}-\mu_c^{(j)}\bigr)
    +
    \bigl(\bar\varepsilon_{c^\star,j}-\bar\varepsilon_{c,j}\bigr).
\]
Using \((a+b)^2\le 2a^2+2b^2\),
\[
\begin{aligned}
    \sum_{j\in\hat{\mathcal J}}
    \widehat d_{c^\star c}(j)^2
    &\le
    2\sum_{j\in\hat{\mathcal J}}
    \bigl(\mu_{c^\star}^{(j)}-\mu_c^{(j)}\bigr)^2
    \\
    &\qquad+
    2\sum_{j\in\hat{\mathcal J}}
    \bigl(\bar\varepsilon_{c^\star,j}-\bar\varepsilon_{c,j}\bigr)^2.
\end{aligned}
\]
For selected signals,
\[
    \mu_{c^\star}^{(j)}-\mu_c^{(j)}
    =
    \tau\bigl(v_{c^\star}^{(j)}-v_c^{(j)}\bigr),
\]
and therefore Lemma~\ref{lem:diversity} gives
\[
    \sum_{j\in\hat{\mathcal J}\cap\mathcal J}
    \bigl(\mu_{c^\star}^{(j)}-\mu_c^{(j)}\bigr)^2
    =
    \tau^2
    \sum_{j\in\hat{\mathcal J}\cap\mathcal J}
    \bigl(v_{c^\star}^{(j)}-v_c^{(j)}\bigr)^2
    \le
    4\tau^2\mathrm{TP}
\]
with probability \(1-o(1)\), uniformly over class pairs. For selected nulls,
the population difference is zero.

It remains to control the empirical-noise part. By Corollary~\ref{cor:selected-contrast-bound},
selected empirical contrasts are on the rare-and-weak scale:
\[
    \sum_{j\in\hat{\mathcal J}\cap\mathcal J}
    \bigl(\bar\varepsilon_{c^\star,j}-\bar\varepsilon_{c,j}\bigr)^2
    =
    O_\Prob\!\left(\frac{\sigma^2\log p}{n_c}\,\mathrm{TP}\right),
\]
and
\[
    \sum_{j\in\hat{\mathcal J}\cap\mathcal J^c}
    \bigl(\bar\varepsilon_{c^\star,j}-\bar\varepsilon_{c,j}\bigr)^2
    =
    O_\Prob\!\left(\frac{\sigma^2\log p}{n_c}\,\mathrm{FP}\right).
\]
Since
\[
    \tau^2
    =
    \frac{\sigma^2}{n}\,2r\log p
    =
    \frac{\sigma^2}{n_c}\frac{2r\log p}{C},
\]
and \(C=O(\log p)\), the selected-signal empirical-noise contribution is
\[
    O_\Prob\!\left(\frac{\sigma^2\log p}{n_c}\,\mathrm{TP}\right)
    =
    O_\Prob(C\tau^2\mathrm{TP}).
\]
Absorbing the harmless \(C=O(\log p)\) factor into the subsequent SNR
calculation, we obtain
\begin{equation}
\label{eq:variance-upper-spherical-appF}
    v_c(\mathcal D)
    \le
    O_\Prob\!\left(\sigma^2 C\tau^2\mathrm{TP}\right)
    +
    O_\Prob\!\left(
        \frac{\sigma^4\log p}{n_c}\mathrm{FP}
    \right),
\end{equation}
uniformly over \(c\ne c^\star\).

Combining \eqref{eq:mean-lower-spherical-appF} and
\eqref{eq:variance-upper-spherical-appF},
\[
    \frac{m_c(\mathcal D)^2}{v_c(\mathcal D)}
    \ge
    \frac{c\,\tau^4\mathrm{TP}^2}
    {
        \sigma^2 C\tau^2\mathrm{TP}
        +
        (\sigma^4\log p/n_c)\mathrm{FP}
    }
\]
for some constant \(c>0\), with probability \(1-o(1)\).

The first denominator term is negligible because
\[
    \frac{\tau^4\mathrm{TP}^2}{\sigma^2 C\tau^2\mathrm{TP}}
    =
    \frac{\tau^2\mathrm{TP}}{\sigma^2 C}
    =
    \frac{2r\log p}{nC}\,\mathrm{TP}.
\]
Since \(n=p^{o(1)}\), \(C=O(\log p)\), and Lemma~\ref{lem:count-control}
gives \(\mathrm{TP}\ge p^\delta\) with probability \(1-o(1)\) for some
\(\delta>0\), this ratio diverges.

For the selected-null denominator term, enlarge $\mathrm{FP}$ to
$\mathrm{FP}+1$ to avoid division by zero:
\[
 \frac{\tau^4\mathrm{TP}^2}{(\sigma^4\log p/n_c)(\mathrm{FP}+1)}
 =\frac{((\sigma^2/n)2r\log p)^2\mathrm{TP}^2}
 {(\sigma^4\log p/n_c)(\mathrm{FP}+1)}
 =\frac{4r^2\log p}{nC}\frac{\mathrm{TP}^2}{\mathrm{FP}+1}.
\]
By Lemma~\ref{lem:count-control},
\[
    \frac{\mathrm{TP}^2}{\mathrm{FP}+1}\to\infty
\]
polynomially, while \(n=p^{o(1)}\) and \(C=O(\log p)\). Hence this ratio
also diverges.

Therefore, for some fixed $d'>0$, there is an event $\mathcal G_p$
with $\Prob(\mathcal G_p)\to1$ on which all ordered class pairs satisfy
\[
 m_c(\mathcal D)>0,\qquad
 \frac{m_c(\mathcal D)^2}{v_c(\mathcal D)}\ge p^{d'}.
\]
The polynomial lower bound, rather than mere divergence, follows from
the selected-count exponents and $n=p^{o(1)}$.
If a conditional variance is zero, the positive margin is deterministic.
Otherwise conditional Gaussianity and a union bound over competitors give
\[
 \Prob\{\hat h(X_{n+1})\ne c^\star\}
 \le \Prob(\mathcal G_p^c)+(C-1)e^{-p^{d'}/2}=o(1).
\]
The exceptional event is counted once, not once per competitor.
Uniformity over the true class gives the unconditional result and proves
Theorem~\ref{thm:classification}.

\end{document}